\pdfoutput=1
\documentclass[10pt,twocolumn,letterpaper]{article}

\usepackage[pagenumbers]{cvpr}

\usepackage{multirow}
\usepackage{colortbl}
\usepackage{tikz}
\usepackage{pgfplots}
\pgfplotsset{compat=1.18}

\definecolor{cvprblue}{rgb}{0.21,0.49,0.74}
\definecolor{tabhead}{RGB}{221,233,247}\definecolor{tabgroup}{RGB}{241,246,252}\definecolor{tabours}{RGB}{235,232,248} 
\definecolor{tabourslight}{RGB}{245,243,251}%
\newcommand{\tabhead}[2]{\multicolumn{#1}{>{\columncolor{tabhead}[0pt][0pt]}c}{#2}}
\usepackage[pagebackref,breaklinks,colorlinks,allcolors=cvprblue]{hyperref}

\usepackage{microtype}
\renewcommand{\paragraph}[1]{\vspace{.35em}\noindent\textbf{#1}}
\newcommand{\ci}[1]{\textsuperscript{$\pm$#1}}

\newcommand{\dataset}{FindEgo-1M\xspace}
\newcommand{\ours}{EgoRay\xspace}
\newcommand{\vggto}{VGGT-$\Omega$\xspace}
\newcommand{\who}{wearer identification\xspace}
\newcommand{\where}{wearer segmentation\xspace}
\newcommand{\exo}{mutual localization\xspace}
\newcommand{\Who}{Wearer identification\xspace}
\newcommand{\Where}{Wearer segmentation\xspace}
\newcommand{\Exo}{Mutual camera localization\xspace}
\newcommand{\ka}{$\kappa$-Acc\xspace}
\newcommand{\giou}{gIoU@20\xspace}
\newcommand{\vex}{\mathbf{e}_\mathrm{exo}}
\newcommand{\veg}{\mathbf{e}_\mathrm{ego}}
\newcommand{\hvex}{\hat{\mathbf{e}}_\mathrm{exo}}
\newcommand{\hveg}{\hat{\mathbf{e}}_\mathrm{ego}}
\newcommand{\hc}{\hat{\mathbf{c}}}
\newcommand{\aex}{\mathrm{AE}_\mathrm{exo}}
\newcommand{\aeg}{\mathrm{AE}_\mathrm{ego}}
\newcommand{\cmark}{\ensuremath{\checkmark}}
\newcommand{\no}{--}

\title{Who Is Ego? Scaling Cross-View Camera Wearer \\ Identification and Segmentation beyond Humans}

\author{
Yifan Jin$^{1,*}$\hspace{0.4em} Zheng Zhang$^{2,*}$\hspace{0.4em} Tianwen Qian$^{3}$\hspace{0.4em} Dong Wang$^{4}$\hspace{0.4em} Bo Zhao$^{5}$\hspace{0.4em} Yanwei Fu$^{2,6}$\hspace{0.4em} Yuqian Fu$^{7,\dagger}$\\[2pt]
\normalsize $^{1}$Central South University \quad $^{2}$Fudan University \quad $^{3}$East China Normal University\\
\normalsize $^{4}$Shanghai AI Laboratory \quad $^{5}$Shanghai Jiao Tong University \quad $^{6}$Shanghai Innovation Institute \quad $^{7}$KAUST\\
{\tt\small yifanjin1124@gmail.com} \quad {\tt\small yuqian.fu.ai@gmail.com}
}

\begin{document}
\twocolumn[{%
\renewcommand\twocolumn[1][]{#1}%
\maketitle
\vspace{-3.2em}
\begin{center}
\includegraphics[width=\textwidth]{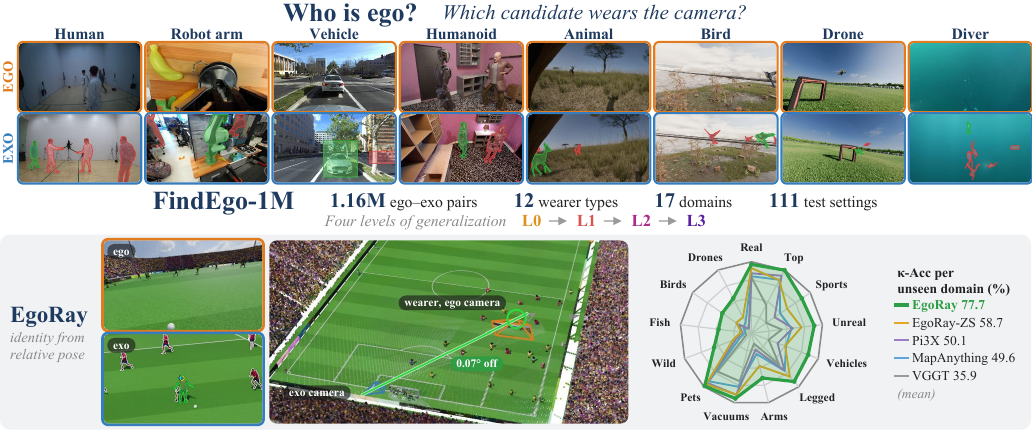}
\captionof{figure}{
\textbf{Illustration of our FindEgo-1M dataset and EgoRay method.}
\textbf{Top:} ego--exo pairs across diverse camera wearers, with the wearer highlighted in green and other candidates in red. FindEgo-1M spans 1.16M pairs, 12 wearer types, 17 domains, and 111 test settings at four generalization levels.
\textbf{Bottom:} EgoRay identifies the wearer through relative camera geometry, with a direction error of $0.07^\circ$ in this example.
\textbf{Right:} \ka on each unseen domain of \cref{tab:who}, with the mean in the legend.
} 
\label{fig:teaser}
\end{center}
\vspace{-.4em}
}]
{\renewcommand{\thefootnote}{\fnsymbol{footnote}}\footnotetext[1]{Equal contribution.}\footnotetext[2]{Corresponding author.}}

\begin{abstract}
A first-person (ego) image reveals what a camera sees, but not who or what carries it. Given a synchronized third-person (exo) image, camera wearer identification asks \emph{who is ego}: which candidate carries the ego camera? We extend this task beyond humans to robots, vehicles, animals, and drones, and broaden its spatial scope (\emph{where is ego}) from selecting a provided candidate mask to wearer segmentation with absence prediction and mutual camera localization across ego--exo views. We introduce \textbf{\dataset}, comprising 1.16M real and simulated ego--exo pairs across 12 wearer types, 17 domains, and 111 test settings. Our approach exploits a geometric relation: the ego camera's projection into the exo image, i.e., the epipole, lies on or near its wearer. We introduce \textbf{\ours-ZS}, the first method to turn a frozen feed-forward 3D reconstruction model, \vggto, into a training-free solution for wearer identification and segmentation. Its geometric readout ranks candidate masks or segmentation proposals by angular offset and distance to the ego camera, while the relative pose provides mutual camera directions. We then develop the full \textbf{\ours} model through pose-supervised training on \dataset and label-free test-time adaptation constrained by candidate masks. 
Without task-specific training, EgoRay-ZS achieves 78.0\% identification accuracy on TF2023's original labels \cite{zhao2024pen}, versus 52.9\% reported for the dataset-trained PEN \cite{zhao2024pen}.
On \dataset, EgoRay achieves state-of-the-art chance-corrected accuracy of 87.1 and 77.7 on seen and unseen settings. On matched queries, EgoRay also outperforms GPT-6 Astra by 31.6 and 21.3 points on seen and unseen settings, respectively, while running 18$\times$ faster.
Data, code, and models have been released at \href{https://travor278.github.io/EgoRay/}{EgoRay}.

\end{abstract}

\section{Introduction}
\label{sec:intro}
Given synchronized first-person (ego) and third-person (exo) images, camera wearer identification asks \emph{who is ego}: which person in the exo view carries the ego camera~\cite{fan2017identifying,zhao2024pen}? This task connects a person's first-person footage to their presence in an external view, supporting applications such as body-camera analysis in broadcast or surveillance footage. When identifiers, GPS, and camera calibration are unavailable, this association must be inferred visually.

Benchmarks such as TF2023 and TF2025~\cite{zhao2024pen,zhao2026sequence} have supported progress on this task. However, existing research remains limited in three respects.
\textit{(1) Narrow wearer scope:} studies focus on humans, overlooking the need to associate onboard and external observations in other settings. 
\textit{(2) Limited data scale and diversity:} existing datasets constrain both training and systematic evaluation of generalization to new environments and wearer types. 
\textit{(3) Limited task scope:} existing evaluations mostly identify the wearer among provided candidates or select a detected mask for segmentation~\cite{zhao2024pen}, leaving segmentation without given candidates, absence prediction, and mutual camera localization underexplored.

We therefore ask whether \emph{who is ego} can scale beyond human-centric identification, and address this question along three dimensions.
First, we broaden the wearer scope to robots, vehicles, animals, and drones, recognizing cross-view camera wearer identification as a shared problem across camera-bearing agents. These settings introduce distinct challenges: robots of the same model can look nearly identical, while birds and drones may occupy only a few pixels in the exo view.
Second, we scale training and evaluation through \textbf{\dataset}, combining public recordings, such as TF2023~\cite{zhao2024pen} and Ego-Exo4D~\cite{grauman2024egoexo4d}, with new simulations to provide 1.16M ego--exo pairs across 12 wearer types, 17 domains, and 111 test settings. Real recordings capture real-world variability, while simulation broadens coverage and supplies exact geometry. 
Evaluation spans four levels of generalization relative to training: new recordings at seen locations (L0), new locations within seen domains (L1), unseen data sources or scene types with seen wearer types (L2), and unseen wearer types (L3). The top of \cref{fig:teaser} illustrates \dataset.
Finally, we move beyond candidate-mask selection to wearer segmentation without given candidates, requiring methods to delineate the wearer or report its absence. We further study mutual camera localization, estimating each camera's direction from the other.

Existing methods associate ego and exo views through appearance or motion matching~\cite{fan2017identifying,ardeshir2016ego2top,zhao2024pen,zhao2026sequence}. Appearance is ambiguous among similar-looking candidates, while motion requires video and distinct trajectories. We instead exploit a geometric relation: \emph{the wearer is where the camera is}. The ego camera's projection into the exo image, i.e., the epipole, lies on or near its wearer~\cite{chen2019epipole,chen2021belief}. The bottom of \cref{fig:teaser} illustrates how relative camera geometry reveals wearer identity. We are the first to use a feed-forward 3D reconstruction model, \vggto~\cite{vggtomega2026}, for this task.  Our \textbf{EgoRay-ZS} combines the angular offset to each candidate mask with camera distance to distinguish candidates along similar rays. The same readout ranks SAM~3 proposals~\cite{carion2025sam3} for segmentation and predicts absence through angular offset, while relative pose provides mutual camera directions.
The full \textbf{EgoRay} model improves pose generalization through pose-supervised training on \dataset and test-time adaptation, which constrains the projected ego camera to lie on or near a candidate mask, providing a geometric learning signal without wearer labels.

Without task-specific training, EgoRay-ZS identifies 78.0\% of wearers on TF2023's original labels~\cite{zhao2024pen}, compared with 52.9\% reported for the dataset-trained PEN~\cite{zhao2024pen} (\cref{sec:exp_prior}). On \dataset, \ours achieves state-of-the-art identification accuracy at all four generalization levels. For the most challenging unseen wearer types (L3), it reaches 71.1 chance-corrected accuracy, versus 42.1 for the strongest competing 3D model. It further outperforms strong multimodal LLMs, including Claude Opus and GPT-6 Astra, by 16.5 and 21.3 percentage points, respectively, on matched unseen identification queries (\cref{tab:llm}).

Our contributions are fourfold.
\textit{(1) \textbf{Task:}} we extend wearer identification beyond humans and to segmentation without given candidates, absence prediction, and mutual camera localization.
\textit{(2) \textbf{Data}:} we introduce \dataset, with 1.16M real and simulated ego--exo pairs for large-scale training and systematic generalization evaluation.
\textit{(3) \textbf{Method:}} we explore a geometric perspective on camera wearer understanding and present two models: zero-shot \ours-ZS and trained \ours.
\textit{(4) \textbf{Results:}} we achieve state-of-the-art results on \dataset and TF2023, outperforming prior wearer identification methods, competing 3D models, and strong multimodal LLMs.

\section{Related Work}
\label{sec:related}

\paragraph{Camera wearer identification.}
Early work matches appearance and motion across views~\cite{fan2017identifying}, assigns egocentric viewers to humans in top views by graph matching~\cite{ardeshir2016ego2top,ardeshir2018egocentric}, jointly segments and identifies humans~\cite{xu2018joint}, or predicts where the wearer is~\cite{wen2021seeing}. Ego-surfing correlates head motion across videos~\cite{yonetani2015egosurfing}.
PEN~\cite{zhao2024pen} fuses person and environment cues to identify and segment human wearers on TF2023, MAF~\cite{zhao2026sequence} extends it to sequences, EgoExoBench~\cite{he2025egoexobench} poses the question to multimodal LLMs, and EgoPrivacy~\cite{egoprivacy2025} retrieves the wearer.
All of them address humans, and none predicts a relative pose.
Closest to our approach, Han~\etal~\cite{han2019mha,han2020cvmhat,han2022dmha} and BEVSee~\cite{bevsee2024} locate cameras by aligning the layout of a crowd of detected people across views, while we use a 3D prior on one pair.

\paragraph{Ego--exo understanding.}
Paired first- and third-person datasets~\cite{sigurdsson2018actor,huang2024egoexolearn,grauman2024egoexo4d} support a growing set of cross-view tasks.
Representations learned in one view transfer to the other for action recognition~\cite{li2021egoexo,luo2025rosetta}, and generative models translate exo videos into the ego view~\cite{luo2024putmyself,liu2024exo2egov,xu2025egoexogen, mahdi2025exo2egosyn}.
Object correspondence segments in one view the object queried in the other~\cite{fu2025objectrelator, murlabadia2025omama, pan2026v}, also with VGGT features~\cite{vggts2026}.
SEERBench~\cite{seerbench2026} asks multimodal LLMs spatial questions across views, and Ego-Exo4D-HM~\cite{egoexo4dhm2026} assumes known head poses.
Though these tasks share the challenge of cross-view understanding, none identifies the ego camera wearer.

\paragraph{Feed-forward 3D reconstruction.}
DUSt3R~\cite{wang2024dust3r}, MASt3R~\cite{leroy2024mast3r}, and Reloc3r~\cite{dong2025reloc3r} regress pointmaps or relative pose from image pairs, CUT3R~\cite{wang2025cut3r} does so over a stream through a recurrent state, and VGGT~\cite{wang2025vggt}, $\pi^3$ with its release Pi3X~\cite{wang2025pi3}, MapAnything~\cite{keetha2025mapanything}, \vggto~\cite{vggtomega2026}, and Depth Anything~3~\cite{lin2025da3} predict cameras and geometry of many views in one pass. We show that their pose also answers a recognition question.
Epipoles~\cite{hartley2004multiple} have been used for calibration and mutual localization~\cite{bushnevskiy2016multicamera,dhiman2013mutual,franchi2013mutual,stegagno2016ground,wang2022anonymous} and linked to wearer location~\cite{fan2017identifying,chen2019epipole,chen2021belief}, but prior work has not developed an epipole-based method for identifying wearers among candidates.

\paragraph{Test-time adaptation.}
Test-time training updates a model on each test input through a self-supervised auxiliary task~\cite{fu2021meta, fu2023styleadv, fu2024cross, sun2020ttt}, fully test-time adaptation minimizes the entropy of predictions~\cite{wang2021tent}, and source-free adaptation trains on its own pseudo-labels~\cite{liang2020shot}.
For 3D reconstruction, Test3R~\cite{yuan2025test3r} enforces consistency between the pointmaps of image triplets, and TTT3R~\cite{chen2026ttt3r} treats a recurrent model as an online learner.
Our objective needs neither an auxiliary task nor extra views. The task itself constrains every single pair, which makes our objective a geometric form of learning from candidate label sets~\cite{cour2011partial} and from bags of instances~\cite{dietterich1997mil}.

\section{\dataset: Tasks, Data, and Evaluation}
\label{sec:benchmark}

\subsection{Task Formulation}
\label{sec:task}
The \emph{wearer} is the body that carries the ego camera, be it a human, robot, vehicle, animal, or drone.
The input is one ego image $I_\mathrm{ego}$ and one exo image $I_\mathrm{exo}$ taken at the same moment, with no wearer identity, pose, or intrinsics. The exo camera is \emph{fixed} or \emph{peer} (worn by another body).
Let $(R,\mathbf{t})$ map exo to ego camera coordinates, so that the ego center in exo coordinates is $\mathbf{c}=-R^\top\mathbf{t}$ and the exo center in ego coordinates is $\mathbf{t}$. The two epipoles, as unit directions, are
\begin{equation}
\vex=\mathbf{c}/\|\mathbf{c}\|,\qquad \veg=\mathbf{t}/\|\mathbf{t}\|,\qquad \veg=-R\,\vex .
\label{eq:epipoles}
\end{equation}
We define three tasks on this input.

\paragraph{\Who (Track A).}
Given $I_\mathrm{ego}$, $I_\mathrm{exo}$, and the masks of all instances in $I_\mathrm{exo}$, a method selects the wearer. The candidates include humans, animals, and machines, including those without cameras. Crucially, under Track A, the wearer is always among the provided mask candidates.

\paragraph{\Where (Track B).}
Given $I_\mathrm{ego}$ and $I_\mathrm{exo}$, Track B provides wearer category words, such as ``robot, person'', instead of predefined candidate masks. A method must output the wearer's mask or report it \emph{absent}. 
The category words are shared within each set, specifying wearer types, not identities.

\paragraph{\Exo (Track C).}
Given $I_\mathrm{ego}$ and $I_\mathrm{exo}$, this track requires models to predict the directions $\vex$ and $\veg$ of each camera as viewed from the other.

\begin{table*}[t]
\centering
\caption{\textbf{Composition of \dataset.} Test omits two sets outside the averages (58,148 queries). Simulated sets have pinhole (v1) and real-lens (v2) versions. $^\dagger$Partly simulated. $^\S$Locked: used for no design decision.}
\label{tab:data}
\footnotesize
\setlength{\tabcolsep}{3pt}
\resizebox{\textwidth}{!}{%
\begin{tabular}{llrrcl@{\hspace{10pt}}llrrcl}
\toprule
\tabhead{6}{\textbf{Public data}} & \tabhead{6}{\textbf{Our simulations}} \\
\cmidrule(lr{8pt}){1-6}\cmidrule(lr){7-12}
Set & Wearers & Train & Test & Level & Tracks & Domain & Scenes & Train & Test & Level & Tracks \\
\midrule
TF2023~\cite{zhao2024pen} & humans & 46,296 & 21,689 & L1 & A B & Indoor humans & 269 rooms & 292,623 & 22,697 & L1 & A B C \\
Harmony4D~\cite{khirodkar2024harmony4d} & humans & 47,000 & 24,724 & L0 & A C & Nature & 8 sites & 245,772 & 26,508 & L1 & A B C \\
CoMind~\cite{comind2026} & humans & 40,000 & 87,214 & L0 & A B C & City & 100 layouts & 145,669 & 19,330 & L1 & A B C \\
Ego-Exo4D~\cite{grauman2024egoexo4d} & humans & 23,256 & 8,862 & L0/L1 & A B C & Divers, ROVs, fish & 36 scenes & 32,284 & 17,897 & L1/L3 & A B C \\
EgoHumans~\cite{khirodkar2023egohumans} & humans & \no & 51,169 & L2 & A B C & Thermal humans & 36 rooms & 3,391 & 2,484 & L1 & A B C \\
CORE4D-Real~\cite{liu2025core4d} & humans & \no & 36,479 & L2 & A B C & Sports (4 games) & 96 scenes & \no & 48,910 & L2 & A B C \\
Ego4D-TF~\cite{grauman2022ego4d,zhao2026sequence} & humans, peer & \no & 26,685 & L2 & A B & Unreal humans & 36 maps & \no & 40,494 & L2 & A B C \\
CVMHAT, CSRD$^\dagger$~\cite{han2020cvmhat,bevsee2024} & humans, top view & \no & 7,487 & L2 & A & Humanoids, quadrupeds & 78 sites & \no & 40,603 & L3 & A B C \\
V2X family$^\dagger$~\cite{xiang2024v2xreal,yu2023v2xseq,li2022v2xsim,xu2022opv2v,opv2vair,airv2x2025,urbaning2025} & vehicles & \no & 7,258 & L3 & A C & Robot arms & 25 cells & \no & 4,376 & L3 & A B C \\
DROID~\cite{khazatsky2024droid} & robot arm & \no & 586 & L3 & A B C & Robot vacuums & 59 rooms & \no & 9,410 & L3 & A B C \\
ALOHA$^\S$~\cite{zhao2023aloha} & robot arms & \no & 540 & L3 & A & Pets, wild animals & 128 scenes & \no & 38,050 & L3 & A B C \\
TUMTraf-V2X$^\S$~\cite{zimmer2024tumtrafv2x} & vehicles & \no & 294 & L3 & A & Birds, drones & 40 scenes & \no & 33,302 & L3 & A B C \\
MILUV$^\S$~\cite{shalaby2026miluv} & drones & \no & 998 & L3 & A & & & & & & \\
\bottomrule
\end{tabular}}
\end{table*}

\subsection{Dataset Construction}
\label{sec:data}
We scale the data in two complementary ways. Public recordings bring real images, and simulation brings the wearer types and conditions that recordings lack. One pipeline turns both into queries with the same labels and protocol (\cref{sec:supp_data}).

\paragraph{Real recordings.}
Ego--exo recordings of humans provide real exo views and scenes, and top-view, vehicle, robot-arm, and drone sets add other viewpoints, wearers, and environments (\cref{tab:data}).
Most of these recordings lack wearer masks. We therefore use SAM~3~\cite{carion2025sam3} to track humans and associate each track with a camera using the projected camera center. When camera poses are unavailable, we ask a multimodal LLM to identify the wearer at several moments and assign the track by majority vote. Human annotators then audit the resulting labels (\cref{sec:supp_audit}).

\paragraph{Simulation.}
Renders contain humanoid and quadruped robots, robot arms, robot vacuums, pets, and wild animals, which often look near-identical, and birds, drones, and divers, whose ego images share little with the exo image. They also show humans of all ages in many more scenes, including crowded ones.
We render with Blender~\cite{blender}, Unreal Engine with UnrealZoo environments~\cite{zhong2025unrealzoo}, and Isaac Sim~\cite{isaacsim}, which supply exact instance masks, depth, and poses. Since real ego cameras are mostly fisheye, every simulated set also comes with lens profiles fitted to actual devices (\cref{sec:supp_data}).
The realism analysis of the simulation is provided in \cref{sec:supp_realism}.

\paragraph{Queries and quality.}
A query pairs an ego and an exo image of one moment. Every instance in view is a candidate, and a wearer out of view makes the query absent.
We inspect every set visually. In a blind audit of the seven real human sets, the automatic label is wrong for 0.3\% of the uniformly drawn queries on which both annotators agree. We also checked the TF2023 annotations by hand and corrected 797 wearer labels of its test set, 795 of them in one scene where two identities are swapped (\cref{sec:supp_data}).

\subsection{Dataset Statistics}
\label{sec:stats}
\dataset holds 1.16M labeled pairs with the wearer visible, 569K real and 589K simulated, of 12 wearer types in 17 domains (\cref{tab:data}). Its training and test lists use 0.95M of these pairs. Training draws its samples from 876K queries, and the 636K test queries form 111 settings at four levels of generalization (\cref{sec:eval}). 
The query count exceeds the pair count because each simulated pair yields one query per lens version (pinhole and real-device lenses), and queries without a visible wearer are added (\cref{sec:supp_data}).

\subsection{Evaluation Protocol}
\label{sec:eval}
\paragraph{Splits and levels of generalization.}
Splits are by scene. L0 is a new recording at a location seen in training, and L1 a new location in a seen domain. L2 is a seen wearer type from a new source or scene type (\cref{tab:data}). L3 is a wearer type absent from training (humanoid and quadruped robots, robot arms, robot vacuums, road vehicles, pets, wild animals, fish, birds, and drones).
We call L0--L1 \emph{seen} and L2--L3 \emph{unseen}.
A \emph{setting} is a set with one exo type, lens version, and level of generalization.

\paragraph{Metrics for Track A.}
As the number $N$ of candidates varies tenfold across settings, we report accuracy (Acc) and the chance-corrected $\kappa\text{-Acc}=(\text{Acc}-c)/(1-c)$, with $c$ the mean of $1/N$ over the queries of a setting.

\paragraph{Metrics for Track B.}
Absent wearers count in the main score, since a score over present queries alone rewards methods that never judge presence.
Like GRES~\cite{liu2023gres}, our score counts absent targets, and it weights absent queries at 20\% in every setting, $\text{\giou}=0.8\,\text{mIoU-P}+0.2\,\text{N-acc}$, where mIoU-P is the mean IoU on present queries and N-acc the accuracy on absent ones. We also report Pr@0.5, the share of present queries above 0.5 IoU, and VA, the mean accuracy of the presence verdict on present and on absent queries. \giou and VA are averaged over the settings with absent queries. The track averages the settings with manual or rendered masks, whose unseen ones are all simulated, and reports the sets with automatic masks separately (\cref{sec:supp_where}).

\paragraph{Metrics for Track C.}
Directions keep their sign, so a reversed direction counts as wrong.
We report the medians of the angular errors $\aex$ and $\aeg$ and of the lateral error in body widths, $\text{BLE}=b\sin(\min(\aex,90^\circ))/w$ for baseline $b$ and wearer width $w$, and the share MA@$\epsilon$ of pairs with $\max(\aex,\aeg)<\epsilon$.

\paragraph{Metrics for overall assessment.}
Each metric is averaged over the settings of a domain and then over domains, separately for seen and unseen settings, with L3, the part of unseen with new wearer types, also reported alone. Of the 111 settings, 97 enter these macro averages (\cref{sec:supp_protocol}), and 95\% intervals resample test scenes, paired between methods unless noted.
The main results use all 636,194 test queries, the \emph{Full tier}, and other experiments on \dataset the \emph{Gold tier}, 8.4\% sampled by query-key hash (\cref{sec:supp_gold}).

\section{Proposed EgoRay Method}
\label{sec:method}

\begin{figure*}[t]
\centering
\includegraphics[width=\textwidth]{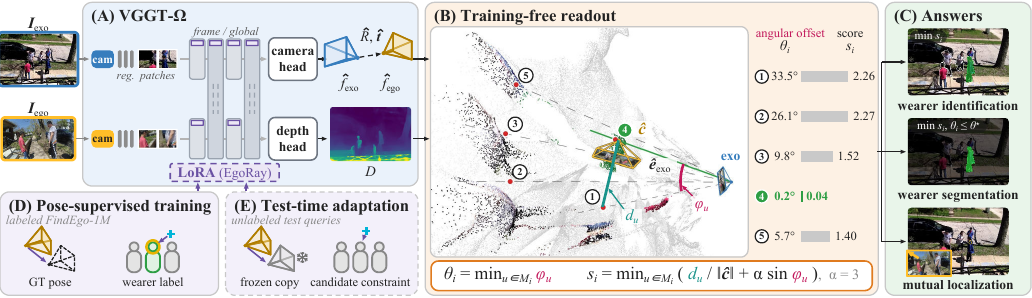}
\caption{\textbf{Overview of \ours.} In \ours-ZS, \vggto predicts the relative pose and the exo depth (A), a readout without trained parameters gives each candidate an angular offset $\theta_i$ and a score $s_i$ (B), and the lowest score identifies and segments the wearer while the pose locates both cameras (C). \ours further trains LoRA adapters of (A) by pose-supervised training (D) and label-free test-time adaptation (E).}
\label{fig:pipeline}
\end{figure*}

\paragraph{Overview.}
\Cref{fig:pipeline} shows how \ours answers a query.
(A) The exo and ego images pass through \vggto, which predicts the relative pose of the two cameras and the depth of the exo image.
(B) The pose places the ego camera in the exo view, and a readout without trained parameters gives each candidate wearer an \emph{angular offset} $\theta_i$, the angle between the candidate's direction and the ego-camera direction, and a \emph{camera distance}, the distance from the candidate to the ego camera (\cref{sec:readout}).
(C) The candidate with the lowest score $s_i$, which combines the two, is the wearer. Ranking segmentation proposals in the same way segments it, and the two epipoles locate each camera in the other's image.
With \vggto as released, (A)--(C) form the zero-shot model, \ours-ZS. The trained model, \ours, further trains LoRA adapters of (A) through the same quantities, by (D) pose-supervised training on \dataset and (E) adaptation on unlabeled test queries (\cref{sec:trained}).

\paragraph{Preliminary: \vggto.}
\vggto~\cite{vggtomega2026} is a feed-forward 3D reconstruction model. It encodes each image into patch tokens, adds a camera token and register tokens, and passes the tokens of all images through blocks that alternate frame attention, within each image, with global or register-only attention, across images. From each camera token, a camera head regresses the image's intrinsics and its extrinsics in the frame of the first image, and a dense head predicts depth.

\subsection{Zero-shot Model (\ours-ZS)}
\label{sec:readout}
\ours-ZS feeds $(I_\mathrm{exo},I_\mathrm{ego})$ to \vggto at their natural aspect ratio and takes the ego center $\hc$ in exo coordinates, the epipoles $\hvex,\hveg$ (\cref{eq:epipoles}), and the exo depth $D$. Each candidate is a mask $M_i$ of $I_\mathrm{exo}$. In \who, the candidates are the instance masks of the benchmark, and in \where, SAM~3~\cite{carion2025sam3} proposes them from the category word, so that task needs no given masks.

\begin{figure}[t]
\centering
\includegraphics[width=\columnwidth]{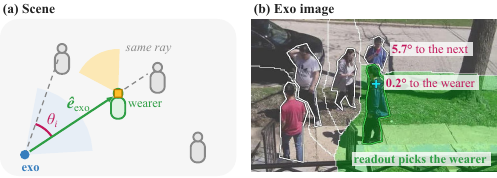}
\caption{\textbf{Geometry of the readout.} (a) Candidates on one ray need depth. (b) Any epipole in the green region picks the wearer by angular offset, and the cross marks the predicted one.}
\label{fig:geom}
\end{figure}

\paragraph{Angular offset.}
A natural readout lifts each candidate with the predicted depth and picks the one nearest the predicted camera. Predicted depth, however, is less accurate at long range and on small masks (\cref{sec:supp_read}), whereas the direction to the camera does not depend on how far it is.
We first measure candidate mask $M_i$ by its \emph{angular offset}, the angle between the ego-camera direction and its nearest mask pixel,
\begin{equation}
\theta_i=\min_{u\in M_i}\phi_u,\qquad \phi_u=\angle\big(\hat K_\mathrm{exo}^{-1}\tilde{u},\ \hvex\big),
\label{eq:offset}
\end{equation}
where $\hat K_\mathrm{exo}$ is the predicted exo intrinsic matrix and $\tilde u$ is pixel $u$ in homogeneous coordinates.
It comes with a simple guarantee. Let $\gamma_\mathrm{own}$ and $\gamma_\mathrm{other}$ be the angles from the true epipole to the wearer's mask and to the nearest other mask, with the rays of $\hat K_\mathrm{exo}$. By the triangle inequality, the candidate with the smallest angular offset is correct whenever $\aex<\frac{1}{2}(\gamma_\mathrm{other}-\gamma_\mathrm{own})$ (\cref{fig:geom}b).

\paragraph{Camera distance.}
The angular offset already identifies the wearer when the candidates lie in different directions from the exo camera. In crowded scenes, however, several candidates can lie on the viewing ray toward it (\cref{fig:geom}a), and the angular offset alone cannot tell them apart, even with the true epipole (\cref{sec:supp_pose}).
We therefore lift each pixel $u$ of mask $M_i$ with the median $z_i$ of $D$ inside the mask and take its distance to the predicted camera, $d_u=\|z_i\hat K_\mathrm{exo}^{-1}\tilde{u}-\hc\|$.
Dividing it by the baseline $\|\hc\|$ makes it comparable across scenes of any scale.

\paragraph{Wearer identification via ranking.}
The score of a candidate combines both quantities pixel by pixel,
\begin{equation}
s_i=\min_{u\in M_i}\big(d_u/\|\hc\|+\alpha\sin\phi_u\big),
\label{eq:mix}
\end{equation}
where $\sin\phi_u$, with $\phi_u$ clipped at $90^\circ$, is the distance from the camera to the ray through $u$ in units of the baseline, and $\alpha$ weighs it. Taking the minimum over the pixels of a mask, rather than scoring a fixed point such as its center, lets the camera sit anywhere on its wearer, from the head of a human to the gripper of a robot arm (\cref{sec:supp_read}). The candidate with the lowest score is the wearer.
Since \vggto treats the first image as the reference, its prediction depends on the input order. We run both orders, $(I_\mathrm{exo},I_\mathrm{ego})$ and $(I_\mathrm{ego},I_\mathrm{exo})$, and average the angular offset and the score of each candidate, and the epipoles, over the two (\cref{sec:supp_impl}).

\paragraph{Segmentation and presence.}
\label{sec:where}
Given the category word, SAM~3~\cite{carion2025sam3} proposes instances of that category in the exo image, and the proposal with the lowest score $s_i$ becomes the wearer's mask.
The answer is \emph{absent} in two cases: when even this proposal has an angular offset above a threshold $\theta^\star$, chosen on the development list, and when the predicted ego camera lies behind the exo camera, where no visible wearer can be. \Cref{sec:supp_seg} compares this design with prompting SAM~3 at the epipole.

\paragraph{Mutual camera localization.}
The same pose gives both epipoles, $\hvex$ and $\hveg$, which locate each camera in the other's image.
Unlike the ranking, which uses only where the ego camera is, \mbox{$\hveg=-\hat R\,\hvex$} also depends on the predicted rotation (\cref{eq:epipoles}).

\subsection{Trained Model (\ours)}
\label{sec:trained}
\ours keeps the \ours-ZS readout and improves pose via pose-supervised training and test-time adaptation.

\paragraph{Pose-supervised training.}
\label{sec:train}
To fit the pose prediction of \vggto to ego--exo pairs of camera wearers, which differ from its own training data (\cref{sec:supp_overlap,sec:supp_readgroup}), we propose pose-supervised training on the training split of \dataset. It trains LoRA~\cite{hu2022lora} adapters of \vggto with
\begin{multline}
\mathcal{L}_\text{sup}=\ell(\hvex,\vex)+\ell(\hveg,\veg)+\angle(\hat R,R)\\
+\sum_{v\in\{\mathrm{exo},\mathrm{ego}\}}\Big|\log\frac{\hat f_v}{f_v}\Big|+\mathcal{L}_\text{wearer},
\label{eq:loss}
\end{multline}
where $\ell$ is the chord between unit vectors, $\angle$ the geodesic rotation error, and $f$ the focal length.
$\mathcal{L}_\text{wearer}$ sums the cross-entropies of the wearer label under the logits $-\theta_i/\tau_a$ and $-\bar s_i/\tau_m$ at learned temperatures, where $\bar s_i$ is the score of \cref{eq:mix} at the centroid of the mask's upper third with $\alpha=1$. Training thus acts on the quantities the readout uses.
The focal term closes a shortcut. The wearer loss alone could sharpen its logits by shrinking the predicted focal length, which spreads the rays apart, instead of improving the pose.

\paragraph{Test-time adaptation.}
\label{sec:tta}
To further improve the pose on novel test data, we adapt the network on the unlabeled queries of each test set. The readout cannot do this alone, since it only ranks the candidates under the predicted pose.
In \who, the wearer of a query is among its candidates by construction, so the ego camera must project onto or near one of their masks.
We turn this into a transductive loss,
\begin{equation}
\mathcal{L}_\text{cand}=\mathbb{E}_q\Big[-\log\sum_i e^{-\theta_i/\tau_a}-\log\sum_i e^{-\bar s_i/\tau_m}\Big],
\label{eq:cand}
\end{equation}
averaged over both input orders.
The loss needs no wearer label, since it is low whenever the predicted ego camera lies on any of the candidates.
The temperatures come from pose-supervised training.
Moving each epipole onto its nearest candidate would mostly confirm the current picks. The gain comes instead from the adapter that all queries of the set share, which also corrects queries whose epipole lay on a wrong candidate (\cref{sec:supp_tta}).

The constraint concerns only where the ego camera is, not how it is oriented, and enlarging the exo focal length eases it by drawing the rays together.
A frozen copy of the starting model, marked by a superscript 0, anchors rotation and focal lengths,
\begin{multline}
\mathcal{L}_\text{anchor}=\|\hat R-\hat R^0\|_F+\sum_{v\in\{\mathrm{exo},\mathrm{ego}\}}\Big|\log\frac{\hat f_v}{\hat f_v^0}\Big|\\
+\mathbf{1}\big[\min_i\theta^0_i<1^\circ\big]\,\ell(\hvex,\hvex^0)/\tau_a,
\label{eq:anchor}
\end{multline}
where $\mathbf{1}[\cdot]$ is the indicator, so the last term keeps an epipole in place when it already lies within $1^\circ$ of a candidate.

\subsection{Training and Inference}
\label{sec:train_infer}
\ours-ZS requires no training.
\ours trains the LoRA once on \dataset with $\mathcal{L}_\text{sup}$ (\cref{eq:loss}) and adapts it on each test set's unlabeled Gold queries with $\mathcal{L}_\text{cand}+\mathcal{L}_\text{anchor}$ (\cref{eq:cand,eq:anchor}). All sets share the hyperparameters. Identification and mutual localization share one copy, adapted on the candidate masks of identification. Localization itself reads only the pose. Segmentation adapts a separate copy on SAM~3 proposals (\cref{sec:supp_where}).
The two models share the same inference. The readout of \cref{sec:readout} identifies the wearer and locates both cameras from the predicted pose, and segments the wearer by ranking SAM~3 proposals.

\section{Experiments}
\label{sec:exp}

\begin{table*}[t]
\centering
\caption{\textbf{\Who} (\ka, Full tier). Other 3D models use our score with the exo image first ($^\ast$pose only, angular offset alone). Macro: domain means over seen (L0--L1) and unseen (L2--L3) settings. Last two rows: paired differences.}
\label{tab:who}
\scriptsize
\setlength{\tabcolsep}{1.9pt}
\resizebox{\textwidth}{!}{%
\begin{tabular}{l*{6}{c}*{4}{c}*{9}{c}cc}
\toprule
 & \tabhead{6}{Seen (L0--L1)} & \tabhead{4}{Unseen source (L2)} & \tabhead{9}{Unseen wearer (L3)} & \tabhead{2}{Macro} \\
\cmidrule(lr){2-7}\cmidrule(lr){8-11}\cmidrule(lr){12-20}\cmidrule(lr){21-22}
Method & Real & Indoor & Nature & City & Divers & Therm. & Real & Top & Sports & Unreal & Veh. & Legged & Arm & Vac. & Pet & Wild & Fish & Bird & Drone & Seen & Unseen \\
\midrule
Largest mask & 34.8 & 1.1 & 0.2 & 1.8 & 0.7 & 0.4 & 2.9 & 7.0 & 0.4 & $-$4.1 & 7.9 & 0.2 & 0.4 & 0.0 & 0.3 & 0.2 & $-$3.2 & $-$0.1 & 0.2 & 6.5 & 0.9 \\
\midrule
\rowcolor{tabgroup}\multicolumn{22}{l}{\emph{Other models, read with our readout}} \\
DUSt3R~\cite{wang2024dust3r} & 59.8 & 57.0 & 14.4 & 23.5 & 0.9 & 3.3 & 54.1 & 27.7 & 18.9 & 27.4 & 20.7 & 47.6 & 2.7 & 49.6 & 61.8 & 2.0 & 0.8 & 2.8 & 3.5 & 26.5 & 24.6 \\
MASt3R~\cite{leroy2024mast3r} & 75.1 & 69.1 & 41.4 & 40.1 & 2.7 & 9.6 & 72.5 & 14.7 & 35.0 & 52.4 & 34.2 & 60.6 & 4.7 & 60.9 & 78.0 & 8.1 & 2.8 & 4.7 & 5.4 & 39.6 & 33.4 \\
CUT3R~\cite{wang2025cut3r} & 72.7 & 66.4 & 16.7 & 23.5 & 2.1 & 5.8 & 78.4 & $-$1.1 & 15.6 & 30.1 & 12.7 & 48.3 & 6.4 & 50.2 & 58.5 & 6.0 & 2.5 & 3.1 & 5.9 & 31.2 & 24.4 \\
VGGT~\cite{wang2025vggt} & 84.2 & 69.5 & 17.2 & 37.3 & 5.6 & 16.4 & 78.3 & 44.5 & 25.3 & 42.8 & 25.7 & 58.7 & 15.0 & 69.1 & 71.6 & 7.9 & 8.6 & 10.5 & 8.2 & 38.4 & 35.9 \\
Pi3X~\cite{wang2025pi3} & 86.0 & 78.8 & 52.2 & 53.3 & 4.8 & 18.8 & 83.9 & 90.9 & 39.4 & 58.4 & 39.2 & 69.7 & 23.7 & 88.7 & 92.8 & 23.0 & 13.1 & 21.5 & 7.3 & 49.0 & 50.1 \\
MapAnything~\cite{keetha2025mapanything} & 85.3 & 80.0 & 48.5 & 59.4 & 6.0 & 28.8 & 79.7 & 84.5 & 48.9 & 56.4 & 51.6 & 70.3 & 29.9 & 77.1 & 89.8 & 16.9 & 9.1 & 13.7 & 17.6 & 51.3 & 49.6 \\
Depth Anything~3~\cite{lin2025da3} & 77.8 & 85.1 & 58.9 & 55.4 & 7.3 & 19.8 & 85.0 & 75.7 & 46.4 & 58.6 & 25.3 & 71.3 & 29.0 & 76.7 & 89.2 & 20.9 & 12.7 & 15.3 & 19.4 & 50.7 & 48.1 \\
Reloc3r~\cite{dong2025reloc3r}$^\ast$ & 65.7 & 66.7 & 27.2 & 31.7 & 2.9 & 6.1 & 76.5 & 30.7 & 21.8 & 34.9 & 17.9 & 51.9 & 9.4 & 63.9 & 67.1 & 7.1 & 7.1 & 18.2 & 19.0 & 33.4 & 32.7 \\
RoMa~\cite{edstedt2024roma} + E$^\ast$ & 58.9 & 44.3 & 23.4 & 18.6 & 4.0 & 21.9 & 40.8 & 21.5 & 14.1 & 22.9 & 15.5 & 32.8 & 12.6 & 38.3 & 49.8 & 8.0 & 9.0 & 33.0 & 16.8 & 28.5 & 24.2 \\
MASt3R matches + E$^\ast$ & 72.7 & 69.1 & 46.5 & 39.1 & 4.7 & 16.5 & 77.3 & 5.7 & 35.8 & 50.4 & 33.4 & 57.4 & 6.2 & 66.8 & 76.6 & 12.9 & 6.2 & 34.4 & 19.8 & 41.4 & 37.1 \\
\vggto, dense tokens + PnP & 53.3 & 29.6 & 11.4 & 12.9 & 2.3 & 5.5 & 45.8 & 30.7 & 12.9 & 10.4 & 10.6 & 21.3 & 4.2 & 19.4 & 24.2 & 2.5 & 6.8 & 7.9 & 5.1 & 19.2 & 15.5 \\
\midrule
\rowcolor{tabourslight}
\ours-ZS & 91.9 & 90.0 & 72.8 & 68.4 & 10.7 & 29.0 & 91.1 & 83.9 & 60.6 & 70.5 & 50.5 & 81.2 & 47.1 & 88.4 & 95.5 & 28.3 & 14.7 & 25.8 & 25.4 & 60.5 & 58.7 \\
\rowcolor{tabours}
\ours$^\ddagger$ & \textbf{98.2} & \textbf{97.3} & \textbf{93.6} & \textbf{93.1} & \textbf{50.2} & \textbf{90.1} & \textbf{96.4} & \textbf{99.7} & \textbf{85.7} & \textbf{88.6} & \textbf{84.9} & \textbf{90.3} & \textbf{64.6} & \textbf{94.1} & \textbf{98.7} & \textbf{55.9} & \textbf{46.4} & \textbf{49.5} & \textbf{55.0} & \textbf{87.1} & \textbf{77.7} \\
\midrule
\ours-ZS $-$ best other & +6.0 & +4.9 & +13.8 & +9.0 & +3.4 & +0.3 & +6.1 & $-$7.0 & +11.6 & +11.9 & $-$1.1 & +9.9 & +17.1 & $-$0.3 & +2.7 & +5.2 & +1.6 & $-$8.6 & +5.6 & +9.1 & +8.6 \\
\ours $-$ \ours-ZS & +6.3 & +7.3 & +20.9 & +24.7 & +39.5 & +61.0 & +5.3 & +15.8 & +25.1 & +18.1 & +34.4 & +9.0 & +17.6 & +5.7 & +3.2 & +27.6 & +31.7 & +23.8 & +29.6 & +26.6 & +19.0 \\
\bottomrule
\end{tabular}}
\end{table*}

\begin{table}[t]
\centering
\caption{\textbf{\Where} (Full tier). All methods rank the same SAM~3 proposals. Oracle: the proposal with the highest IoU, a ceiling for any ranking of them.}
\label{tab:where}
\footnotesize
\setlength{\tabcolsep}{2.2pt}
\resizebox{\columnwidth}{!}{%
\begin{tabular}{lcccccccc}
\toprule
 & \tabhead{4}{Seen} & \tabhead{4}{Unseen} \\
\cmidrule(lr){2-5}\cmidrule(lr){6-9}
Method & \giou & mIoU-P & Pr@0.5 & VA & \giou & mIoU-P & Pr@0.5 & VA \\
\midrule
Always absent & 20.0 & 0.0 & 0.0 & 50.0 & 20.0 & 0.0 & 0.0 & 50.0 \\
\midrule
Pi3X~\cite{wang2025pi3} & 49.2 & 43.7 & 49.9 & 74.5 & 42.3 & 34.8 & 38.7 & 68.1 \\
MapAnything~\cite{keetha2025mapanything} & 48.4 & 43.6 & 49.9 & 73.4 & 41.1 & 34.9 & 38.8 & 68.7 \\
Depth Anything~3~\cite{lin2025da3} & 51.6 & 45.0 & 51.3 & 77.3 & 42.2 & 32.5 & 36.3 & 70.1 \\
\midrule
\rowcolor{tabourslight}
\ours-ZS & 56.3 & 49.6 & 56.9 & 79.7 & 46.5 & 37.6 & 42.0 & 73.0 \\
\rowcolor{tabours}
\ours$^\ddagger$ & \textbf{67.8} & \textbf{70.7} & \textbf{80.6} & \textbf{83.9} & \textbf{55.2} & \textbf{53.1} & \textbf{58.6} & \textbf{75.8} \\
\midrule
\ours-ZS $-$ best other & +4.6 & +4.6 & +5.6 & +2.3 & +4.2 & +2.7 & +3.2 & +2.9 \\
\ours $-$ \ours-ZS & +11.6 & +21.1 & +23.7 & +4.3 & +8.6 & +15.6 & +16.6 & +2.9 \\
\midrule
Oracle proposal & \no & 79.8 & \no & \no & \no & 75.6 & \no & \no \\
\bottomrule
\end{tabular}}
\end{table}

\paragraph{Datasets.}
The main results (\mbox{\cref{tab:who,tab:where,tab:exo}}) use the Full tier of \mbox{\dataset} and the other experiments its Gold tier (\mbox{\cref{sec:eval}}).
For comparison with prior work, we use TF2023 with its original and corrected labels under the protocol of PEN~\cite{zhao2024pen}, on its whole official test list, and the 30-frame protocol of MAF~\cite{zhao2026sequence}.

\paragraph{Implementation details.}
Pose-supervised training updates rank-64 LoRA adapters of \vggto (1.65\% of its parameters) on the training queries of \dataset (\cref{sec:supp_impl}). Test-time adaptation runs 200 steps per test set and keeps two thirds of its gain with only 25 unlabeled queries per set (\cref{sec:supp_stream}). The readout uses $\alpha=3$. Most hyperparameters were chosen on a development list of training scenes or on held-out halves of scenes, and \cref{sec:supp_design} lists every design decision with the data it saw.

\paragraph{Compared methods.}
We compare with three groups (\cref{tab:who}). The first is a rule that ignores the ego image and picks the largest mask. Other 3D models are read with our score, using their own intrinsics and the exo image first. \vggto is also read through PnP on its matched dense tokens. Pose-only methods (a relative-pose regressor and essential-matrix matching with \vggto intrinsics) use angular offset alone.
Scores are in \% except angles and BLE, and $^\ddagger$ marks transductive results.

\subsection{Main Results}
\label{sec:exp_main}

\paragraph{\Who.}
Without the ego image, accuracy stays near chance on average (\cref{tab:who}).
With the same score, \ours-ZS leads the strongest other model by 9.1 and 8.6 points on the seen and unseen averages, and pose-only methods trail further. 
Pose-supervised training on \dataset and adaptation add 26.6 and 19.0 points to \ours-ZS on seen and unseen settings, and 20.3 on L3 alone (\cref{sec:supp_levels}). \Cref{sec:exp_ablation} separates the two steps.
On three real sets that no design decision used, roadside views of a test vehicle~\cite{zimmer2024tumtrafv2x}, a bimanual robot~\cite{zhao2023aloha}, and quadcopters filming each other~\cite{shalaby2026miluv}, \ours reaches 81.1, 93.6, and 89.5 Acc, against 57.1, 88.5, and 81.2 for the strongest baseline of each set (\cref{sec:supp_locked}).

\paragraph{Comparison with Multimodal LLMs.}
We evaluate multimodal LLMs on wearer identification using an ego image and an exo image with boxed, numbered candidates, asking each model to select the wearer's number (\cref{sec:supp_llm}). On matched queries, as in \cref{tab:llm}, \ours outperforms the strongest evaluated LLM, Claude Opus, by 20.9 and 16.5 percentage points on seen and unseen settings, respectively, and GPT-6 Astra by 31.6 and 21.3 points. Open-weight models perform near chance on unseen settings. Including amortized adaptation, \ours takes 0.4\,s per item, versus 3.0--22.4\,s for the LLMs.

\begin{table}[t]
\centering
\caption{\textbf{Multimodal LLMs} (\ka on the Gold identification items each model answered, \ours on the same items). Time: median seconds per item; for \ours, inference plus adaptation spread over the queries of each set.}
\label{tab:llm}
\scriptsize
\setlength{\tabcolsep}{3pt}
\resizebox{\ifdim\width>\columnwidth\columnwidth\else\width\fi}{!}{%
\begin{tabular}{lrccc>{\columncolor{tabours}}c>{\columncolor{tabours}}c>{\columncolor{tabours}}c}
\toprule
 & & \tabhead{3}{Model} & \tabhead{3}{\ours$^\ddagger$} \\
\cmidrule(lr){3-5}\cmidrule(lr){6-8}
Model & Items & Seen & Unseen & Time & Seen & Unseen & Time \\
\midrule
\rowcolor{tabgroup}\multicolumn{8}{l}{\emph{Proprietary}} \\
Claude Opus & 2,061 & 71.5 & 62.1 & 13.4 & 92.4 & 78.6 & 0.4 \\
GPT-6.1 Sol & 14,802 & 50.5 & 49.1 & 22.4 & 86.5 & 75.9 & 0.4 \\
GPT-6 Astra & 14,181 & 55.4 & 54.5 & 7.2 & 87.0 & 75.8 & 0.4 \\
\midrule
\rowcolor{tabgroup}\multicolumn{8}{l}{\emph{Open-weight}} \\
Qwen3-VL-32B & 5,445 & 11.5 & 3.9 & 3.0 & 87.6 & 73.9 & 0.4 \\
InternVL3.5-38B & 5,445 & 9.6 & $-$2.6 & 9.7 & 87.6 & 73.9 & 0.4 \\
\bottomrule
\end{tabular}}
\end{table}

\paragraph{\Where.}
Ranking SAM~3 proposals with our readout gives a \giou of 56.3 and 46.5 on seen and unseen settings without training, ahead of the other 3D models on every metric, and 67.8 and 55.2 with pose-supervised training and adaptation (\cref{tab:where}).

\begin{table}[t]
\centering
\caption{\textbf{\Exo} (Full tier). Last two rows: differences, paired for MA and plain for the medians.}
\label{tab:exo}
\footnotesize
\setlength{\tabcolsep}{1.6pt}
\resizebox{\columnwidth}{!}{%
\begin{tabular}{lcccccccc}
\toprule
 & \tabhead{4}{Seen} & \tabhead{4}{Unseen} \\
\cmidrule(lr){2-5}\cmidrule(lr){6-9}
Method & MA@5$\uparrow$ & MA@15$\uparrow$ & $\aex\downarrow$ & BLE$\downarrow$ & MA@5$\uparrow$ & MA@15$\uparrow$ & $\aex\downarrow$ & BLE$\downarrow$ \\
\midrule
Mean training direction & \no & \no & 28.0$^\circ$ & \no & \no & \no & 35.3$^\circ$ & \no \\
MASt3R & 15.9 & 36.9 & 45.7$^\circ$ & 5.2 & 14.2 & 33.8 & 47.6$^\circ$ & 25.3 \\
VGGT & 12.0 & 35.4 & 34.7$^\circ$ & 4.4 & 13.2 & 33.7 & 34.6$^\circ$ & 19.4 \\
Pi3X & 14.1 & 40.0 & 29.9$^\circ$ & 3.1 & 22.0 & 42.0 & 26.2$^\circ$ & 18.7 \\
MapAnything & 21.0 & 50.2 & 16.4$^\circ$ & 2.2 & 19.7 & 45.5 & 19.5$^\circ$ & 14.5 \\
Depth Anything 3 & 29.2 & 49.9 & 20.0$^\circ$ & 2.3 & 23.4 & 43.9 & 22.0$^\circ$ & 16.3 \\
Reloc3r & 14.1 & 38.7 & 27.4$^\circ$ & 3.9 & 12.2 & 33.8 & 32.3$^\circ$ & 26.2 \\
\midrule
\rowcolor{tabourslight}
\ours-ZS & 28.5 & 57.9 & 27.3$^\circ$ & 2.4 & 28.1 & 53.7 & 20.0$^\circ$ & 10.6 \\
\rowcolor{tabours}
\ours$^\ddagger$ & \textbf{71.8} & \textbf{83.7} & \textbf{4.1}$^\circ$ & \textbf{0.6} & \textbf{50.6} & \textbf{72.9} & \textbf{7.2}$^\circ$ & \textbf{4.2} \\
\midrule
\ours-ZS $-$ best other & $-$0.7 & +7.6 & +10.9$^\circ$ & +0.2 & +4.7 & +8.2 & +0.5$^\circ$ & $-$3.9 \\
\ours $-$ \ours-ZS & +43.3 & +25.9 & $-$23.2$^\circ$ & $-$1.8 & +22.5 & +19.3 & $-$12.8$^\circ$ & $-$6.4 \\
\bottomrule
\end{tabular}}
\end{table}

\paragraph{\Exo.}
Without training, no model reaches an MA@5 above 29.2 (\cref{tab:exo}).
On the Gold tier, pose-supervised training raises MA@5 to 74.5 and 50.7 on seen and unseen settings, and adaptation trades 1.8 of these points on seen settings for 2.5 on unseen ones (\cref{sec:supp_tta}).

\subsection{Comparison on Prior Benchmarks}
\label{sec:exp_prior}

\begin{table}[t]
\centering
\caption{\textbf{TF2023 under PEN's protocol}, with Mask R-CNN~\cite{he2017maskrcnn} pools above 0.9 or 0.5 confidence. Orig.: the original labels, on which PEN reports (its training code and weights are not public). Other columns: the corrected labels. Ceiling: always picking the pool mask with the highest IoU with the wearer, the most any method choosing from the pool can reach, since the pool misses some wearers.}
\label{tab:pen}
\scriptsize
\setlength{\tabcolsep}{2.5pt}
\resizebox{\ifdim\width>\columnwidth\columnwidth\else\width\fi}{!}{%
\begin{tabular}{lccccc}
\toprule
 & \tabhead{3}{0.9 pool} & \tabhead{2}{0.5 pool} \\
\cmidrule(lr){2-4}\cmidrule(lr){5-6}
Method & Acc, orig. & Acc & mIoU & Acc & mIoU \\
\midrule
PEN~\cite{zhao2024pen}, reported & 52.9 & \no & \no & \no & \no \\
Depth Anything~3~\cite{lin2025da3} & 76.9 & 80.6 & 69.3 & 86.0 & 73.8 \\
\rowcolor{tabourslight}
\ours-ZS & 78.0 & 81.8 & 70.3 & 87.8 & 75.3 \\
\rowcolor{tabours}
\ours$^\ddagger$ & \textbf{81.8} & \textbf{85.6} & \textbf{73.5} & \textbf{91.6} & \textbf{78.5} \\
\midrule
Ceiling & 86.0 & 86.0 & 73.8 & 92.8 & 79.3 \\
\bottomrule
\end{tabular}}
\end{table}

We also evaluate on TF2023~\cite{zhao2024pen} under the protocols of PEN and of MAF~\cite{zhao2026sequence}. The main text reports TF2023 on the corrected labels, except where it compares with numbers published on the original ones, and \cref{sec:supp_pen} gives both.

Under PEN's protocol on TF2023 (\cref{tab:pen}), \ours-ZS identifies 78.0\% of wearers on the original labels without training, against 52.9\% reported for dataset-trained PEN.
Ten of the twelve baselines not built on \vggto also exceed PEN's reported 52.9\% accuracy, three of them with a pose alone (\cref{sec:supp_pen}).
On the corrected labels, the best of them, Depth Anything~3, reaches 80.6\%. Paired over the 12 test scenes, differences from Depth Anything~3 are +1.1 [$-$1.2, 3.3] points for \ours-ZS and +5.0 [2.8, 7.9] for \ours. 
The margin over Depth Anything~3 is smaller than on \dataset because TF2023 leaves little room. PEN's 0.9 pool caps accuracy at 86.0\%, and \ours reaches 85.6\%.

Approximating MAF's 30-frame protocol by summing median-normalized candidate scores across frames, \ours-ZS reaches 95.0\% accuracy on the original labels without training, against 90.3\% reported for MAF (\cref{sec:supp_video}).

\subsection{Ablation Studies}
\label{sec:exp_ablation}

\begin{table}[t]
\centering
\caption{\textbf{Components of \ours} (Gold tier). $\Delta$: paired change of unseen \ka. The last row also trains on other cameras of the same moment (\cref{sec:supp_joint}).}
\label{tab:components}
\footnotesize
\setlength{\tabcolsep}{2.2pt}
\resizebox{\columnwidth}{!}{%
\begin{tabular}{lcccc}
\toprule
 & \tabhead{2}{\ka} & & \tabhead{1}{MA@5} \\
\cmidrule(lr){2-3}\cmidrule(lr){5-5}
Component & Seen & Unseen & $\Delta$ unseen & Unseen \\
\midrule
\vggto camera head, angular offset & 54.6 & 56.7 & \no & 28.8 \\
+ camera distance (mask depth) & 60.9 & 58.1 & +1.4 [0.8, 2.1] & 28.8 \\
\rowcolor{tabourslight}
+ two input orders (\ours-ZS) & 60.7 & 58.6 & +0.5 [0.1, 0.8] & 28.9 \\
+ pose-supervised training, pose labels only & 85.3 & 71.1 & +12.4 [11.6, 13.3] & 55.3 \\
+ wearer loss (pose-supervised training) & 87.8 & 72.6 & +1.6 [0.9, 2.3] & 50.7 \\
\rowcolor{tabours}
+ adaptation (\ours)$^\ddagger$ & 87.4 & 78.2 & +5.5 [4.9, 6.2] & 53.2 \\
+ other cameras$^\ddagger$ & 90.3 & 79.3 & +1.1 [0.4, 1.7] & 55.8 \\
\bottomrule
\end{tabular}}
\end{table}

\paragraph{Components.}
\Cref{tab:components} adds one component at a time. On unseen settings, pose labels contribute most (+12.4), then adaptation (+5.5).
Because adaptation is transductive, we also choose its hyperparameters on half of the scenes and score the other half. On the 18 Gold sets of the families that informed the design, it still adds 7.2 [6.3, 8.1] unseen \ka over pose-supervised training, and 6.4 [5.5, 7.4] when it never sees the scored scenes (\cref{sec:supp_clean}).

\subsection{More Analysis}
\label{sec:exp_more}

\paragraph{Visualization.}
\Cref{sec:supp_qual} contrasts \ours with the strongest other zero-shot 3D model and shows correct answers and failures in every domain.

\paragraph{Error analysis.}
With the true epipole, the nearest mask is correct for 94.7--97.4\% of the queries on birds, drones, vehicles, robot arms, and underwater wearers, so most errors come from the pose (\cref{sec:supp_pose}). Dense herds of wild animals are the exception (79.0\%).

\paragraph{Training data.}
Simulated data alone come within 0.8 \ka of real and simulated data together on the macro averages, while real data alone stay close to zero-shot (shorter recipe, \cref{sec:supp_ftdata}). DINOv3~\cite{simeoni2025dinov3} trained end to end on our queries reaches \ours-ZS on seen settings (62.7 against 60.5 \ka on the Full tier) but only 17.0 on unseen ones (\cref{sec:supp_train}).

\section{Conclusion}
We broadened camera wearer understanding beyond humans, extending identification among given candidates to segmentation without given candidates, with absence prediction, and to mutual camera localization. FindEgo-1M enables large-scale learning and systematic evaluation across diverse wearer types, environments, and levels of generalization. Our \ours framework unifies these tasks through relative camera geometry, combining a zero-shot geometric readout with pose-supervised training and label-free test-time adaptation. It gives the best overall results on \dataset and TF2023, with substantial gains over prior wearer identification methods, competing 3D models, and multimodal LLMs. 
By expanding who can be the wearer, what must be predicted, and the data available for learning, this work lays a foundation for scaling ego--exo understanding across diverse camera-bearing agents. Further discussion of the limitations of our method, adaptation protocol, and data coverage is provided in the supplementary material (\cref{sec:supp_limits}).

\clearpage

{
    \small
    \bibliographystyle{ieeenat_fullname}
    \bibliography{draft}
}

\clearpage
\setcounter{page}{1}\renewcommand{\thepage}{S\arabic{page}}
\maketitlesupplementary

\appendix
\renewcommand{\topfraction}{0.9}\renewcommand{\bottomfraction}{0.8}\renewcommand{\textfraction}{0.05}\renewcommand{\floatpagefraction}{0.75}
\renewcommand{\dbltopfraction}{0.9}\renewcommand{\dblfloatpagefraction}{0.75}
\setcounter{topnumber}{4}\setcounter{bottomnumber}{3}\setcounter{totalnumber}{8}\setcounter{dbltopnumber}{3}
\raggedbottom

\makeatletter
\let\supp@addcontentsline\addcontentsline
\renewcommand{\addcontentsline}[3]{\def\supp@tmp{#1}\def\supp@toc{toc}\ifx\supp@tmp\supp@toc\supp@addcontentsline{stoc}{#2}{#3}\else\supp@addcontentsline{#1}{#2}{#3}\fi}
\makeatother
\section*{Contents}
{\makeatletter\setcounter{tocdepth}{2}\renewcommand{\@pnumwidth}{2.2em}\renewcommand{\@tocrmarg}{3.2em}%
\def\numberline#1{#1\hskip 1em}\@starttoc{stoc}}
\vspace{0.8em}

\newcommand{\qtile}[2]{\begin{tikzpicture}\node[inner sep=0pt] (t) {\includegraphics[width=0.326\textwidth]{#1}};\if\relax\detokenize{#2}\relax\else\node[anchor=north west, inner sep=1.6pt, fill=black, fill opacity=0.62, text opacity=1, text=white, font=\scriptsize\sffamily] at (t.north west) {#2};\fi\end{tikzpicture}}

Below, we provide additional details on \dataset and report supplementary
experiments and analyses of \ours. \Cref{sec:supp_data} describes the construction
and validation of \dataset, \cref{sec:supp_protocol} defines the evaluation
protocol and metrics, and \cref{sec:supp_analysis} analyzes the readout.
\Cref{sec:supp_adapt,sec:supp_impl} describe test-time adaptation, implementation,
and model selection. \Cref{sec:supp_tables,sec:supp_baselines,sec:supp_ablation}
present detailed results, comparisons with prior benchmarks and multimodal
LLMs, and experiments on training data and alternative models.
\Cref{sec:supp_qual} presents qualitative results, followed by a discussion
of limitations (\Cref{sec:supp_limits}) and broader impact (\Cref{sec:supp_impact}).

Unless otherwise specified, results are reported as percentages on the Gold
tier. Superscript values ($\pm$) indicate the half-widths of 95\% confidence
intervals computed at the scene level. For differences between methods, the
intervals are computed using paired results from the same scenes.
The symbol $^\ddagger$ marks transductive results. All results labeled \ours
use unlabeled test-time adaptation on test queries (on \dataset, those of each set's Gold tier).
The TF2023 subset of \dataset uses the corrected labels.

\section{\dataset Construction and Validation}
\label{sec:supp_data}

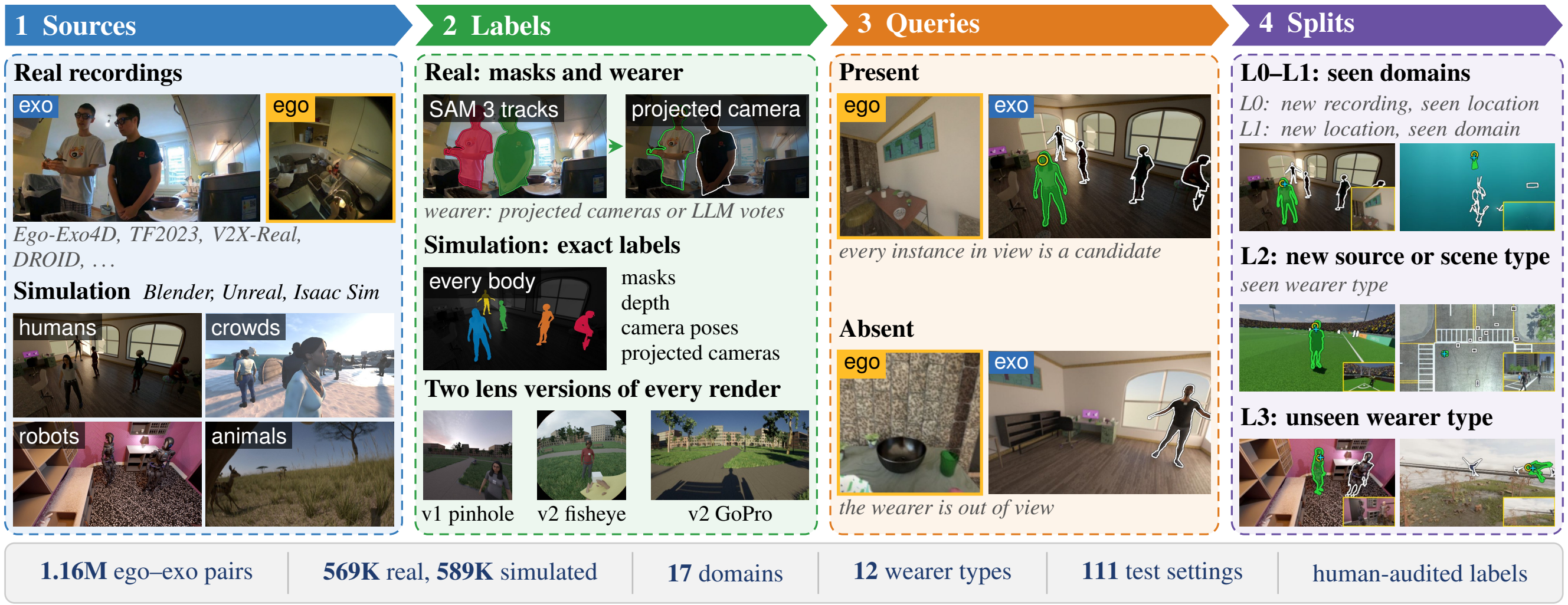
\begin{figure*}[!htb]
\definecolor{dcCAM}{HTML}{2F6DB5}
\definecolor{dcNOTE}{HTML}{555555}
\definecolor{dcSTAT}{HTML}{1F3B6E}
\definecolor{dcEGO}{HTML}{FFBE28}
\definecolor{dcLVIOLET}{HTML}{ECE7F5}
\definecolor{dcVIOLET}{HTML}{6A51A3}
\definecolor{dcLORANGE}{HTML}{FCEBD9}
\definecolor{dcORANGE}{HTML}{E07B1F}
\definecolor{dcLGREEN}{HTML}{DFF2E2}
\definecolor{dcGREEN}{HTML}{2E9E44}
\definecolor{dcLBLUE}{HTML}{DCE9F6}
\definecolor{dcBLUE}{HTML}{367DBD}
\centering
\begin{tikzpicture}[x=1cm, y=-1cm, dchead/.style={anchor=north west, inner sep=0pt, font=\footnotesize\bfseries}, dcnote/.style={anchor=north west, inner sep=0pt, font=\scriptsize\itshape, text=dcNOTE}, dcchip/.style={anchor=north west, inner sep=1.6pt, fill=black, fill opacity=0.62, text opacity=1, text=white, font=\scriptsize\sffamily}]
\fill[dcBLUE] (0.000,0.000) -- (4.370,0.000) -- (4.570,0.230) -- (4.370,0.460) -- (0.000,0.460) -- cycle;
\node[anchor=west, inner sep=0pt, font=\small\bfseries, text=white] at (0.100,0.230) {1\enspace Sources};
\draw[dcBLUE, densely dashed, line width=0.6pt, fill=dcLBLUE, fill opacity=0.55, rounded corners=4pt] (0.000,0.560) rectangle (4.450,5.940);
\fill[dcGREEN] (4.600,0.000) -- (9.020,0.000) -- (9.220,0.230) -- (9.020,0.460) -- (4.600,0.460) -- (4.800,0.230) -- cycle;
\node[anchor=west, inner sep=0pt, font=\small\bfseries, text=white] at (4.900,0.230) {2\enspace Labels};
\draw[dcGREEN, densely dashed, line width=0.6pt, fill=dcLGREEN, fill opacity=0.55, rounded corners=4pt] (4.600,0.560) rectangle (9.100,5.940);
\fill[dcORANGE] (9.250,0.000) -- (13.520,0.000) -- (13.720,0.230) -- (13.520,0.460) -- (9.250,0.460) -- (9.450,0.230) -- cycle;
\node[anchor=west, inner sep=0pt, font=\small\bfseries, text=white] at (9.550,0.230) {3\enspace Queries};
\draw[dcORANGE, densely dashed, line width=0.6pt, fill=dcLORANGE, fill opacity=0.55, rounded corners=4pt] (9.250,0.560) rectangle (13.600,5.940);
\fill[dcVIOLET] (13.750,0.000) -- (17.260,0.000) -- (17.460,0.230) -- (17.260,0.460) -- (13.750,0.460) -- (13.950,0.230) -- cycle;
\node[anchor=west, inner sep=0pt, font=\small\bfseries, text=white] at (14.050,0.230) {4\enspace Splits};
\draw[dcVIOLET, densely dashed, line width=0.6pt, fill=dcLVIOLET, fill opacity=0.55, rounded corners=4pt] (13.750,0.560) rectangle (17.460,5.940);
\node[dchead] at (0.090,0.650) {Real recordings};
\begin{scope}\clip (0.090,1.010) rectangle ++(2.770,1.420);\node[inner sep=0pt] at (1.475,1.720) {\includegraphics[width=2.770cm]{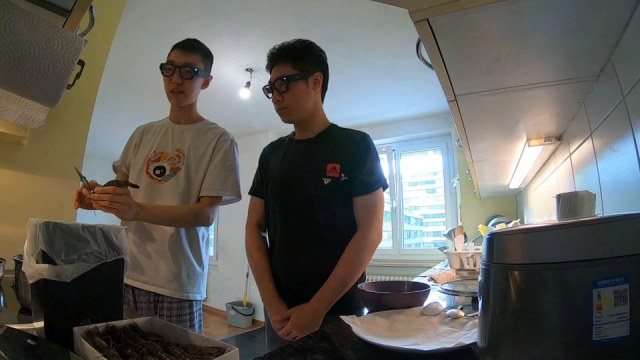}};\end{scope}
\begin{scope}\clip (2.940,1.010) rectangle ++(1.420,1.420);\node[inner sep=0pt] at (3.650,1.720) {\includegraphics[width=1.420cm]{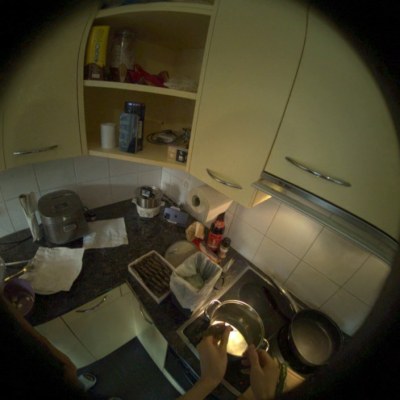}};\end{scope}
\draw[dcEGO, line width=1pt] (2.940,1.010) rectangle ++(1.420,1.420);
\node[dcchip, fill=dcCAM, fill opacity=1] at (0.090,1.010) {exo};
\node[dcchip, fill=dcEGO, fill opacity=1, text=black] at (2.940,1.010) {ego};
\node[dcnote, text width=4.270cm, align=left] at (0.090,2.480) {Ego-Exo4D, TF2023, V2X-Real,\\DROID, \ldots};
\node[dchead] at (0.090,3.100) {Simulation\enspace\normalfont\scriptsize\itshape Blender, Unreal, Isaac Sim};
\begin{scope}\clip (0.090,3.460) rectangle ++(2.110,1.170);\node[inner sep=0pt] at (1.145,4.045) {\includegraphics[width=2.110cm]{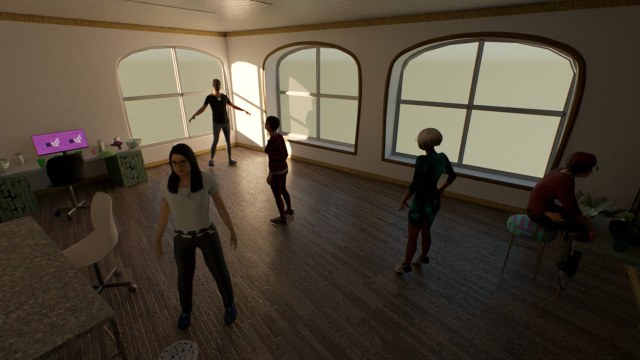}};\end{scope}
\node[dcchip] at (0.090,3.460) {humans};
\begin{scope}\clip (2.250,3.460) rectangle ++(2.110,1.170);\node[inner sep=0pt] at (3.305,4.045) {\includegraphics[width=2.110cm]{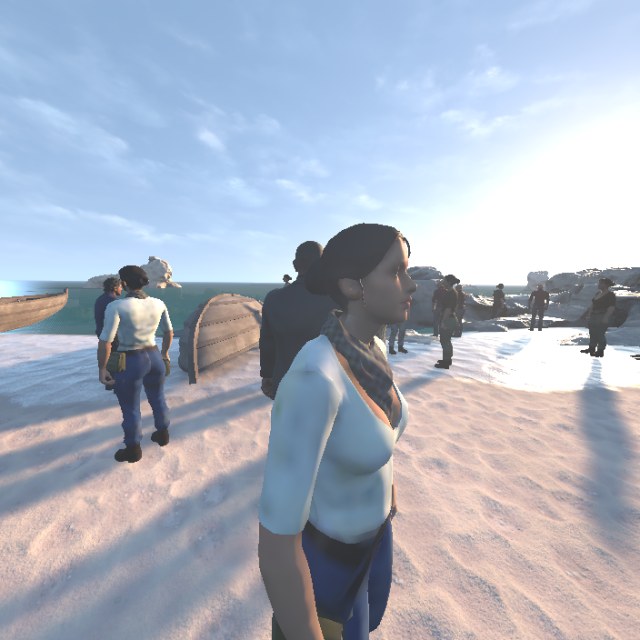}};\end{scope}
\node[dcchip] at (2.250,3.460) {crowds};
\begin{scope}\clip (0.090,4.680) rectangle ++(2.110,1.170);\node[inner sep=0pt] at (1.145,5.265) {\includegraphics[width=2.110cm]{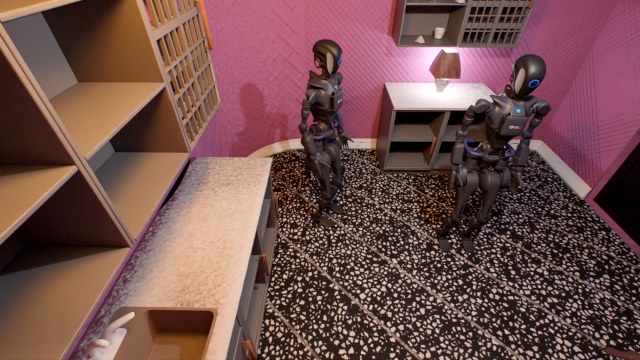}};\end{scope}
\node[dcchip] at (0.090,4.680) {robots};
\begin{scope}\clip (2.250,4.680) rectangle ++(2.110,1.170);\node[inner sep=0pt] at (3.305,5.265) {\includegraphics[width=2.110cm]{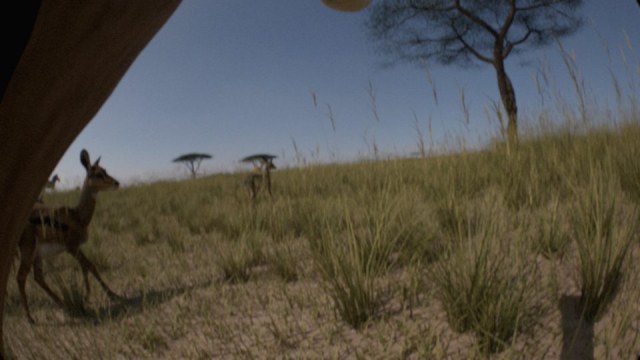}};\end{scope}
\node[dcchip] at (2.250,4.680) {animals};
\node[dchead] at (4.690,0.650) {Real: masks and wearer};
\begin{scope}\clip (4.690,1.010) rectangle ++(2.050,1.153);\node[inner sep=0pt] at (5.715,1.587) {\includegraphics[width=2.050cm]{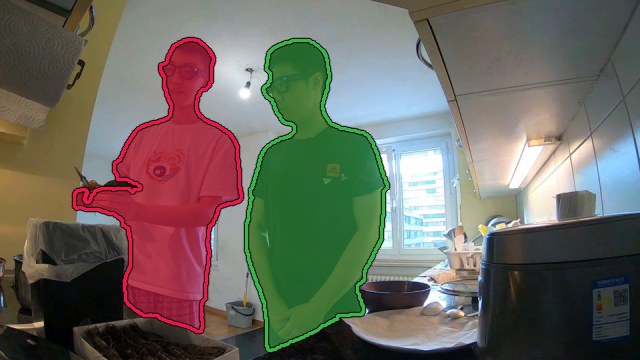}};\end{scope}
\begin{scope}\clip (6.960,1.010) rectangle ++(2.050,1.153);\node[inner sep=0pt] at (7.985,1.587) {\includegraphics[width=2.050cm]{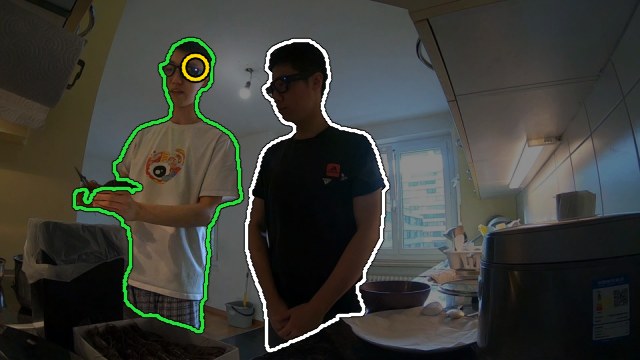}};\end{scope}
\draw[-stealth, dcGREEN, line width=1.1pt] (6.770,1.587) -- (6.930,1.587);
\node[dcchip] at (4.690,1.010) {SAM 3 tracks};
\node[dcchip] at (6.960,1.010) {projected camera};
\node[dcnote] at (4.690,2.213) {wearer: projected cameras or LLM votes};
\node[dchead] at (4.690,2.583) {Simulation: exact labels};
\begin{scope}\clip (4.690,2.943) rectangle ++(2.050,1.153);\node[inner sep=0pt] at (5.715,3.520) {\includegraphics[width=2.050cm]{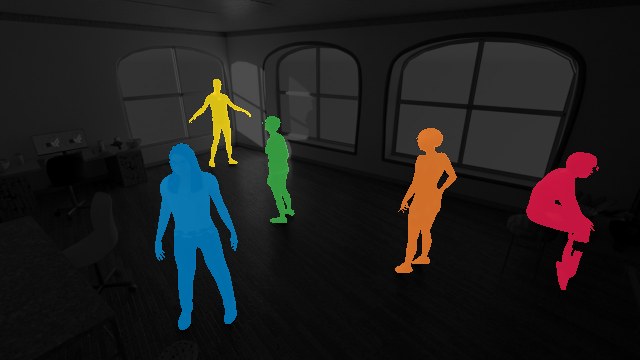}};\end{scope}
\node[dcchip] at (4.690,2.943) {every body};
\node[anchor=north west, inner sep=0pt, font=\scriptsize, align=left] at (6.900,2.963) {masks\\depth\\camera poses\\projected cameras};
\node[dchead] at (4.690,4.196) {Two lens versions of every render};
\begin{scope}\clip (4.690,4.556) rectangle ++(0.994,0.994);\node[inner sep=0pt] at (5.187,5.053) {\includegraphics[width=0.994cm]{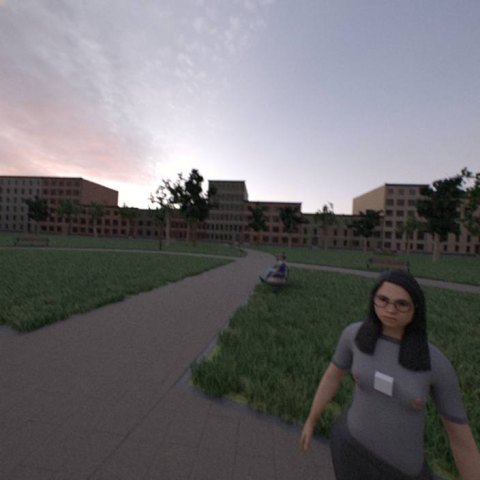}};\end{scope}
\node[anchor=north, inner sep=0pt, font=\scriptsize] at (5.187,5.610) {v1 pinhole};
\begin{scope}\clip (5.967,4.556) rectangle ++(0.994,0.994);\node[inner sep=0pt] at (6.464,5.053) {\includegraphics[width=0.994cm]{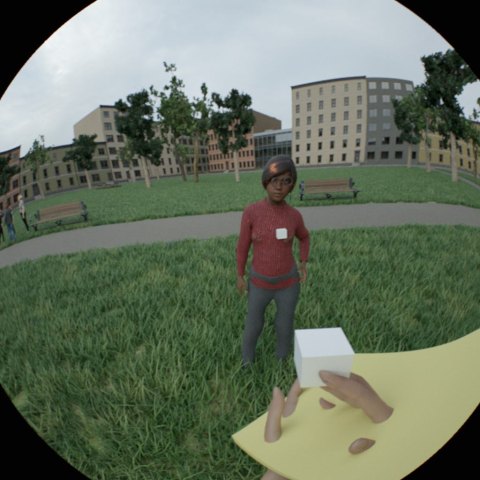}};\end{scope}
\node[anchor=north, inner sep=0pt, font=\scriptsize] at (6.464,5.610) {v2 fisheye};
\begin{scope}\clip (7.243,4.556) rectangle ++(1.767,0.994);\node[inner sep=0pt] at (8.127,5.053) {\includegraphics[width=1.767cm]{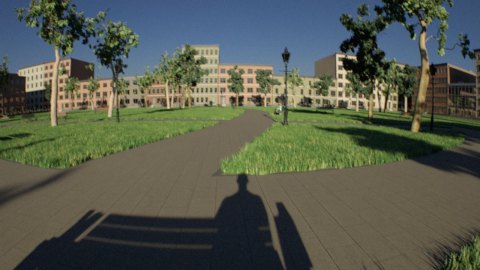}};\end{scope}
\node[anchor=north, inner sep=0pt, font=\scriptsize] at (8.127,5.610) {v2 GoPro};
\node[dchead] at (9.340,0.650) {Present};
\begin{scope}\clip (9.340,1.010) rectangle ++(1.604,1.604);\node[inner sep=0pt] at (10.142,1.812) {\includegraphics[width=1.604cm]{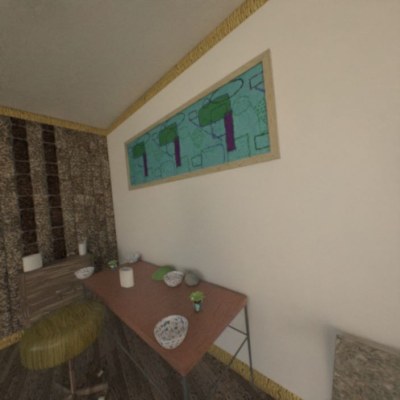}};\end{scope}
\draw[dcEGO, line width=1pt] (9.340,1.010) rectangle ++(1.604,1.604);
\begin{scope}\clip (11.024,1.010) rectangle ++(2.486,1.604);\node[inner sep=0pt] at (12.267,1.812) {\includegraphics[height=1.604cm]{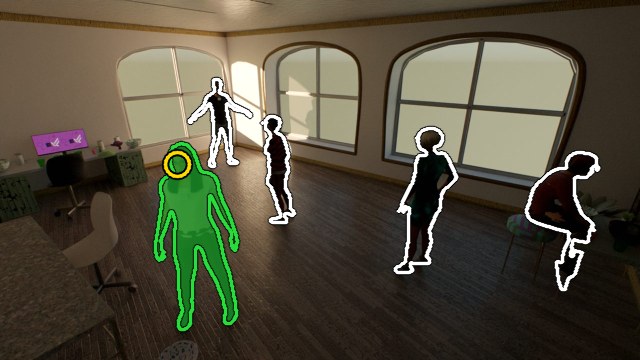}};\end{scope}
\node[dcchip, fill=dcEGO, fill opacity=1, text=black] at (9.340,1.010) {ego};
\node[dcchip, fill=dcCAM, fill opacity=1] at (11.024,1.010) {exo};
\node[dcnote] at (9.340,2.664) {every instance in view is a candidate};
\node[dchead] at (9.340,3.526) {Absent};
\begin{scope}\clip (9.340,3.886) rectangle ++(1.604,1.604);\node[inner sep=0pt] at (10.142,4.688) {\includegraphics[height=1.604cm]{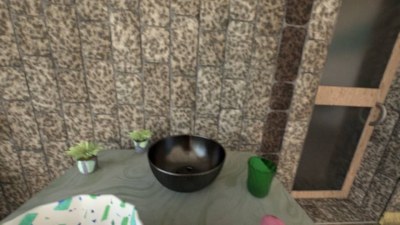}};\end{scope}
\draw[dcEGO, line width=1pt] (9.340,3.886) rectangle ++(1.604,1.604);
\begin{scope}\clip (11.024,3.886) rectangle ++(2.486,1.604);\node[inner sep=0pt] at (12.267,4.688) {\includegraphics[width=2.486cm]{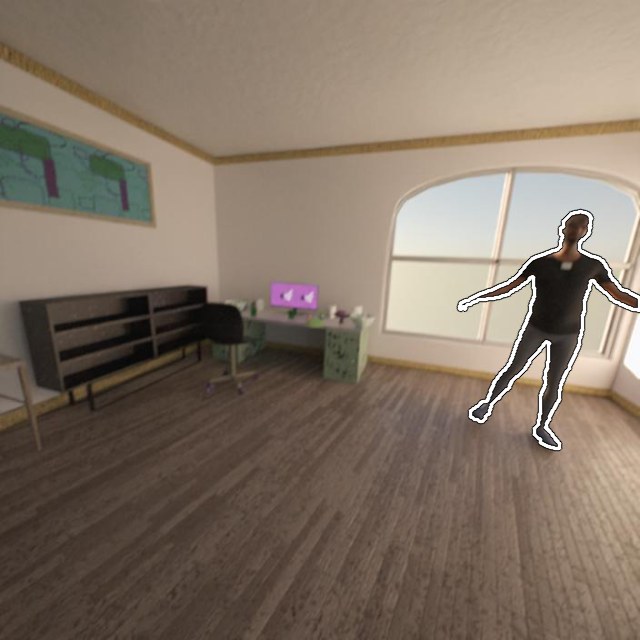}};\end{scope}
\node[dcchip, fill=dcEGO, fill opacity=1, text=black] at (9.340,3.886) {ego};
\node[dcchip, fill=dcCAM, fill opacity=1] at (11.024,3.886) {exo};
\node[dcnote] at (9.340,5.540) {the wearer is out of view};
\node[dchead] at (13.840,0.650) {L0--L1: seen domains};
\node[dcnote, text width=3.530cm, align=left] at (13.840,1.010) {L0: new recording, seen location\\L1: new location, seen domain};
\begin{scope}\clip (13.840,1.560) rectangle ++(1.735,0.976);\node[inner sep=0pt] at (14.707,2.048) {\includegraphics[width=1.735cm]{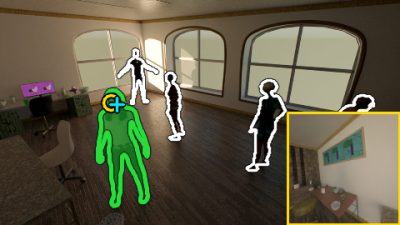}};\end{scope}
\begin{scope}\clip (15.635,1.560) rectangle ++(1.735,0.976);\node[inner sep=0pt] at (16.503,2.048) {\includegraphics[width=1.735cm]{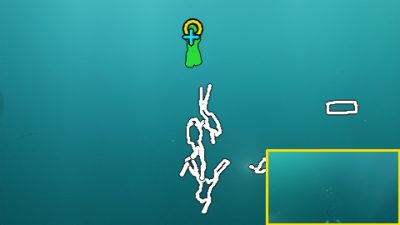}};\end{scope}
\node[dchead] at (13.840,2.712) {L2: new source or scene type};
\node[dcnote] at (13.840,3.072) {seen wearer type};
\begin{scope}\clip (13.840,3.362) rectangle ++(1.735,0.976);\node[inner sep=0pt] at (14.707,3.850) {\includegraphics[width=1.735cm]{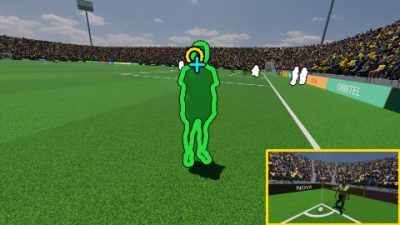}};\end{scope}
\begin{scope}\clip (15.635,3.362) rectangle ++(1.735,0.976);\node[inner sep=0pt] at (16.503,3.850) {\includegraphics[width=1.735cm]{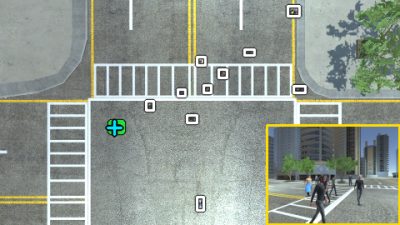}};\end{scope}
\node[dchead] at (13.840,4.514) {L3: unseen wearer type};
\begin{scope}\clip (13.840,4.874) rectangle ++(1.735,0.976);\node[inner sep=0pt] at (14.707,5.362) {\includegraphics[width=1.735cm]{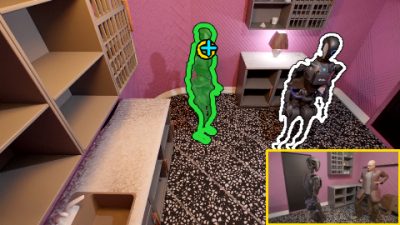}};\end{scope}
\begin{scope}\clip (15.635,4.874) rectangle ++(1.735,0.976);\node[inner sep=0pt] at (16.503,5.362) {\includegraphics[width=1.735cm]{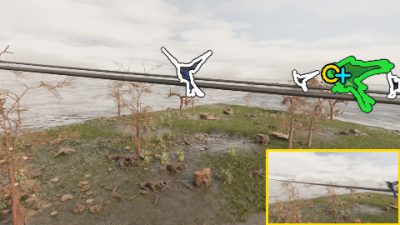}};\end{scope}
\draw[draw=black!20, fill=black!5, line width=0.5pt, rounded corners=3pt] (0,6.040) rectangle (17.460,6.710);
\node[font=\footnotesize, text=dcSTAT, inner sep=0pt] at (1.586,6.375) {\textbf{1.16M} ego--exo pairs};
\node[font=\footnotesize, text=dcSTAT, inner sep=0pt] at (5.103,6.375) {\textbf{569K} real, \textbf{589K} simulated};
\draw[black!25, line width=0.5pt] (3.172,6.140) -- (3.172,6.610);
\node[font=\footnotesize, text=dcSTAT, inner sep=0pt] at (8.074,6.375) {\textbf{17} domains};
\draw[black!25, line width=0.5pt] (7.033,6.140) -- (7.033,6.610);
\node[font=\footnotesize, text=dcSTAT, inner sep=0pt] at (10.396,6.375) {\textbf{12} wearer types};
\draw[black!25, line width=0.5pt] (9.115,6.140) -- (9.115,6.610);
\node[font=\footnotesize, text=dcSTAT, inner sep=0pt] at (12.973,6.375) {\textbf{111} test settings};
\draw[black!25, line width=0.5pt] (11.677,6.140) -- (11.677,6.610);
\node[font=\footnotesize, text=dcSTAT, inner sep=0pt] at (15.864,6.375) {human-audited labels};
\draw[black!25, line width=0.5pt] (14.268,6.140) -- (14.268,6.610);
\end{tikzpicture}
\caption{\textbf{Data engine of \dataset.} (1) Ego--exo pairs come from public recordings and our Blender, Unreal, and Isaac Sim simulations. (2) On real recordings without wearer masks, SAM~3 tracks instances; the projected ego camera (or, without camera poses, LLM votes) identifies the wearer, and human annotators audit these labels (\cref{sec:supp_audit}). Simulation supplies exact masks, depth, camera poses, and projected camera centers for every body, with each render in two lens versions. (3) A query is present if the wearer appears in the exo view (every visible instance is a candidate) and absent otherwise. (4) Splits are by scene, and test settings cover four generalization levels relative to training: new recordings at seen locations (L0), new locations in seen domains (L1), new sources or scene types with seen wearer types (L2), and unseen wearer types (L3). Real images: CoMind~\cite{comind2026}.}
\label{fig:data}
\end{figure*}

This section details dataset construction (\cref{fig:data}), label audits, realism checks, and a comparison with related benchmarks. Real queries come from existing recordings, whereas simulation extends wearer types, lenses, and scenes; the training and test lists use 0.95M of the 1.16M labeled pairs plus absent queries, with 876K queries in the training pool. Our renders place BEDLAM~\cite{black2023bedlam,pavlakos2019smplx} and MPFB~\cite{mpfb} human bodies in Infinigen~\cite{raistrick2023infinigen,raistrick2024infinigenindoors} and purpose-built scenes, include Poly Haven~\cite{polyhaven} scans and ABO~\cite{collins2022abo} objects as assets, and add RGB-to-thermal queries via a paired thermal test bed.

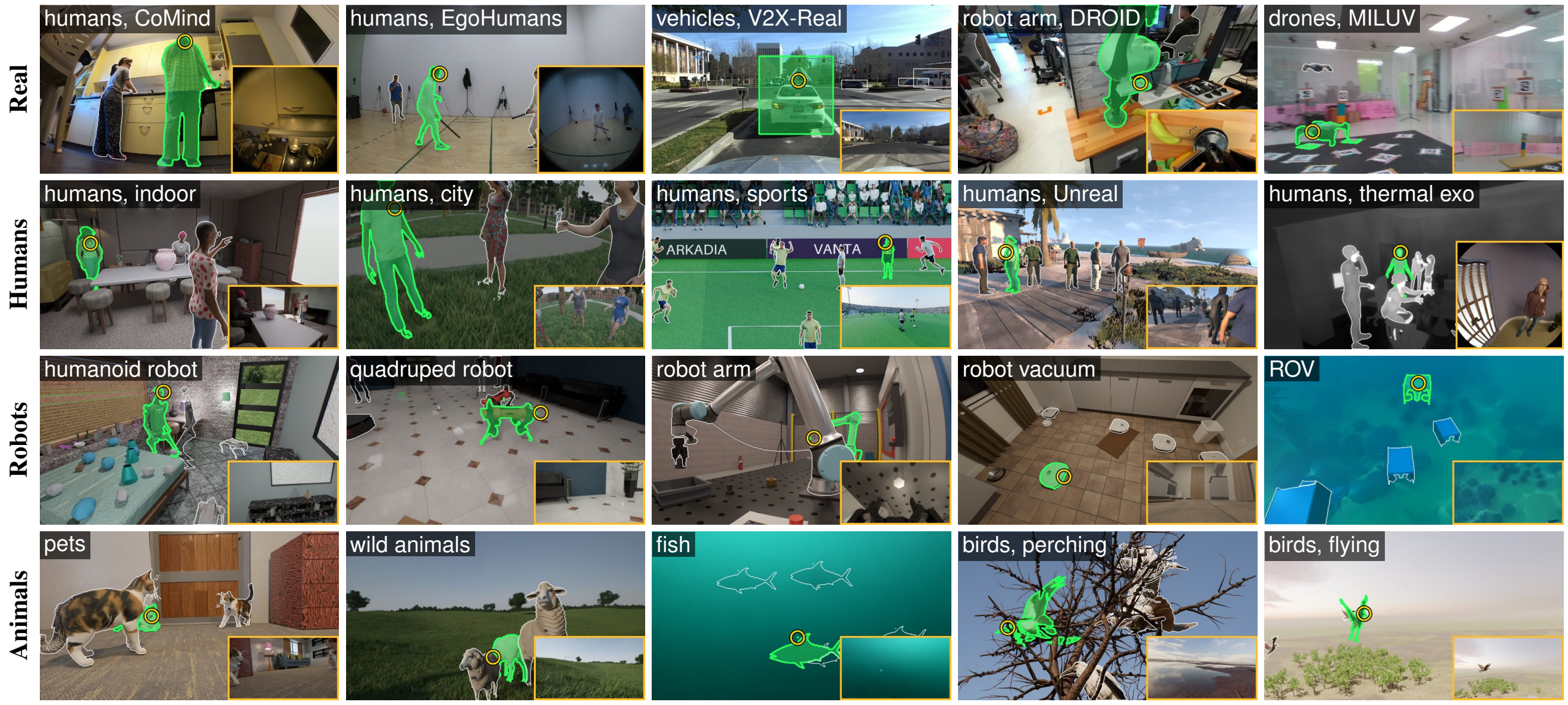
\begin{figure*}[t]
\centering
\begin{tikzpicture}[x=1cm, y=-1cm, galchip/.style={anchor=north west, inner sep=1.3pt, fill=black, fill opacity=0.62, text opacity=1, text=white, font=\scriptsize\sffamily}, galrow/.style={rotate=90, anchor=center, inner sep=0pt, font=\footnotesize\bfseries}]
\node[galrow] at (0.190,0.935) {Real};
\node[anchor=north west, inner sep=0pt] at (0.420,0.000) {\includegraphics[width=3.326cm]{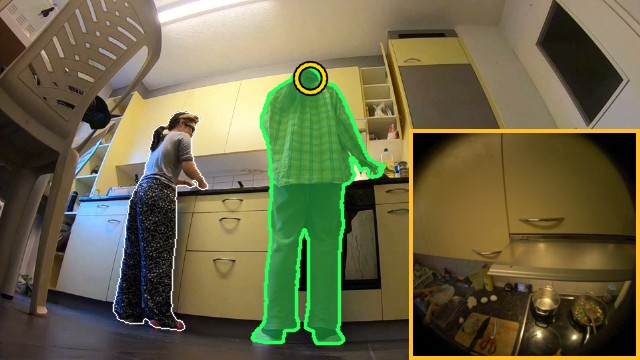}};
\node[galchip] at (0.420,0.000) {humans, CoMind};
\node[anchor=north west, inner sep=0pt] at (3.826,0.000) {\includegraphics[width=3.326cm]{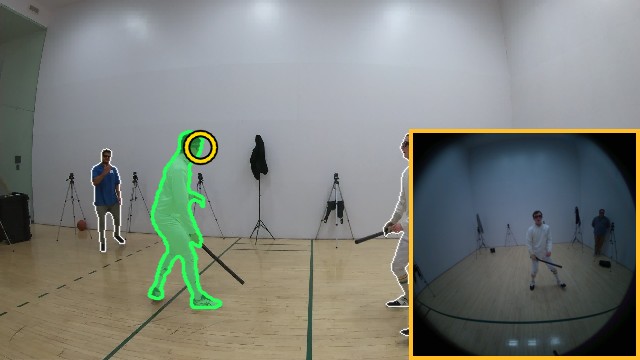}};
\node[galchip] at (3.826,0.000) {humans, EgoHumans};
\node[anchor=north west, inner sep=0pt] at (7.232,0.000) {\includegraphics[width=3.326cm]{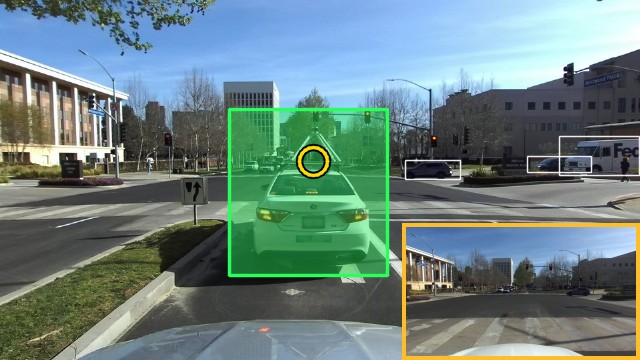}};
\node[galchip] at (7.232,0.000) {vehicles, V2X-Real};
\node[anchor=north west, inner sep=0pt] at (10.638,0.000) {\includegraphics[width=3.326cm]{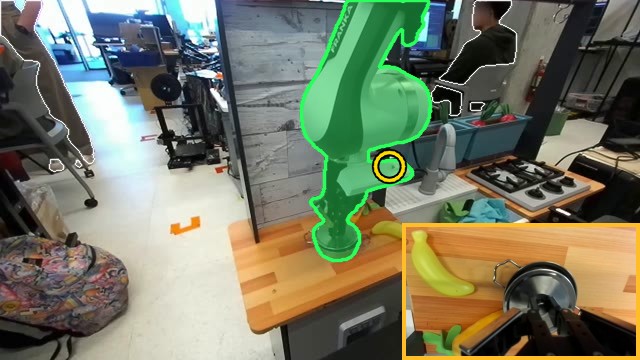}};
\node[galchip] at (10.638,0.000) {robot arm, DROID};
\node[anchor=north west, inner sep=0pt] at (14.044,0.000) {\includegraphics[width=3.326cm]{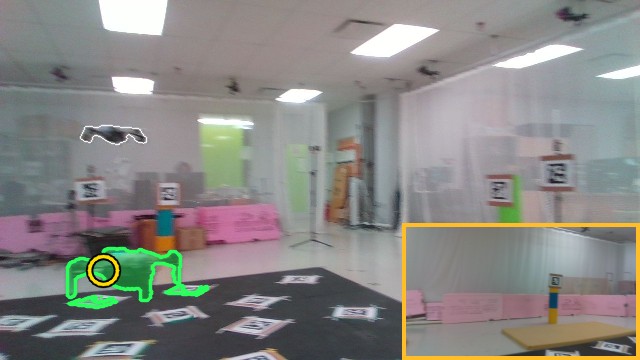}};
\node[galchip] at (14.044,0.000) {drones, MILUV};
\node[galrow] at (0.190,2.886) {Humans};
\node[anchor=north west, inner sep=0pt] at (0.420,1.951) {\includegraphics[width=3.326cm]{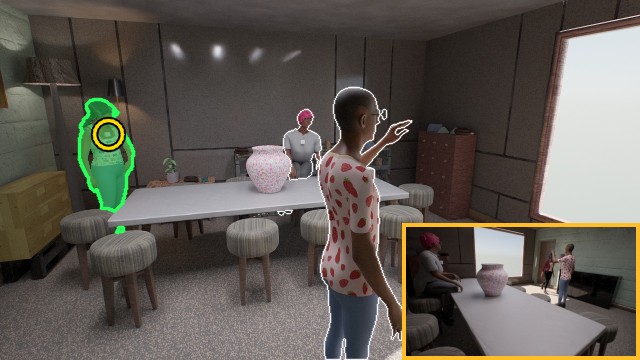}};
\node[galchip] at (0.420,1.951) {humans, indoor};
\node[anchor=north west, inner sep=0pt] at (3.826,1.951) {\includegraphics[width=3.326cm]{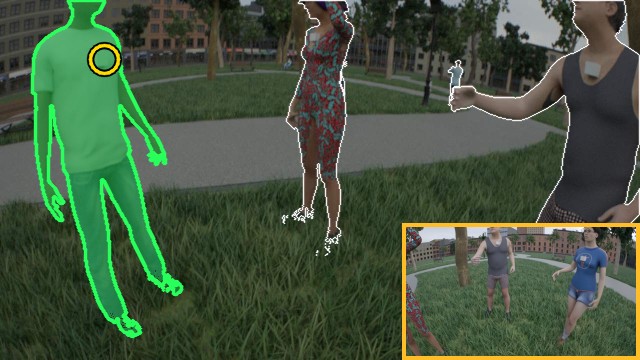}};
\node[galchip] at (3.826,1.951) {humans, city};
\node[anchor=north west, inner sep=0pt] at (7.232,1.951) {\includegraphics[width=3.326cm]{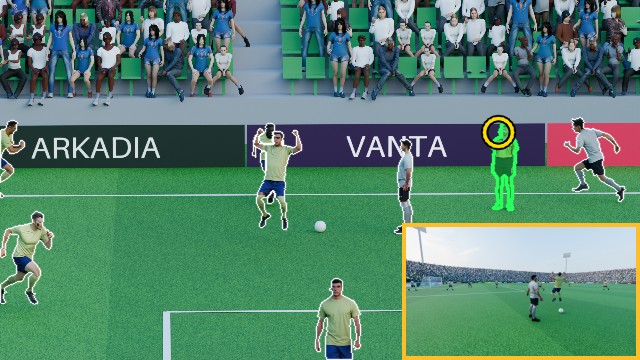}};
\node[galchip] at (7.232,1.951) {humans, sports};
\node[anchor=north west, inner sep=0pt] at (10.638,1.951) {\includegraphics[width=3.326cm]{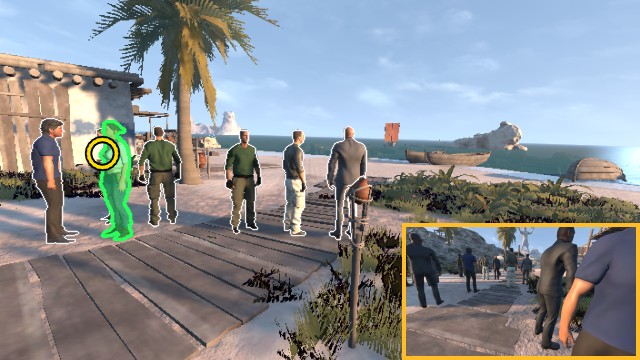}};
\node[galchip] at (10.638,1.951) {humans, Unreal};
\node[anchor=north west, inner sep=0pt] at (14.044,1.951) {\includegraphics[width=3.326cm]{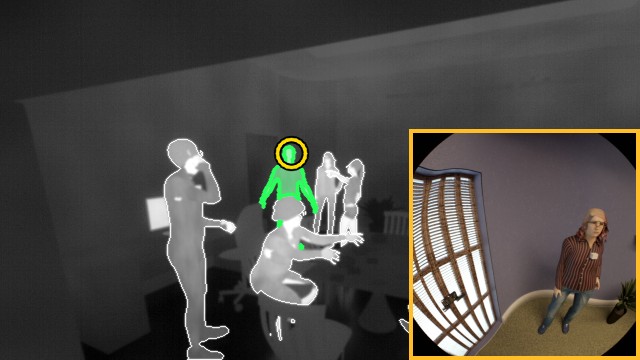}};
\node[galchip] at (14.044,1.951) {humans, thermal exo};
\node[galrow] at (0.190,4.837) {Robots};
\node[anchor=north west, inner sep=0pt] at (0.420,3.902) {\includegraphics[width=3.326cm]{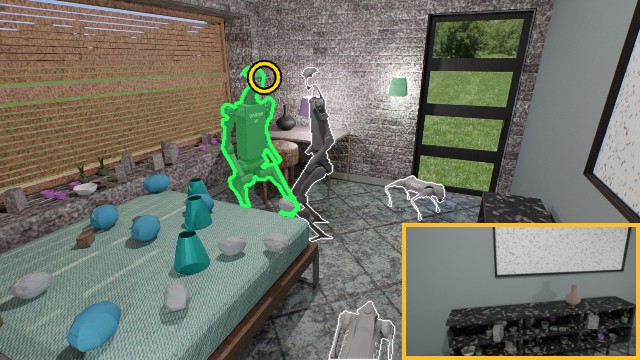}};
\node[galchip] at (0.420,3.902) {humanoid robot};
\node[anchor=north west, inner sep=0pt] at (3.826,3.902) {\includegraphics[width=3.326cm]{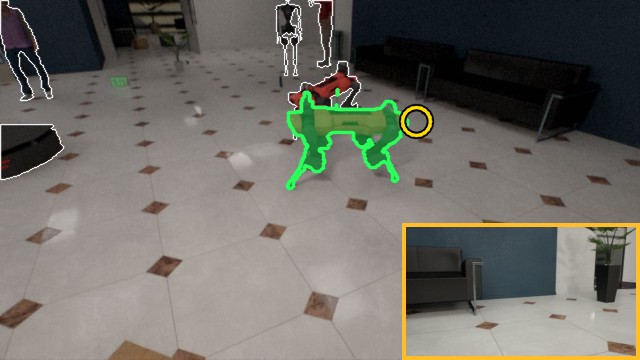}};
\node[galchip] at (3.826,3.902) {quadruped robot};
\node[anchor=north west, inner sep=0pt] at (7.232,3.902) {\includegraphics[width=3.326cm]{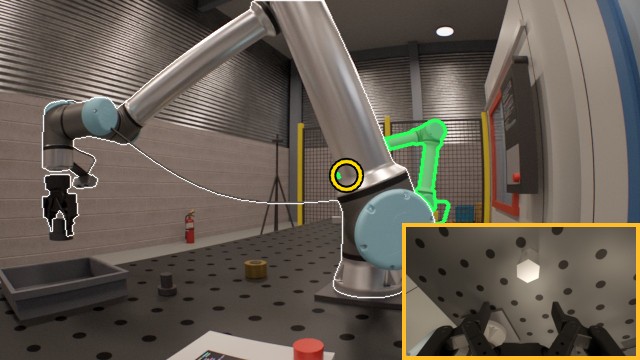}};
\node[galchip] at (7.232,3.902) {robot arm};
\node[anchor=north west, inner sep=0pt] at (10.638,3.902) {\includegraphics[width=3.326cm]{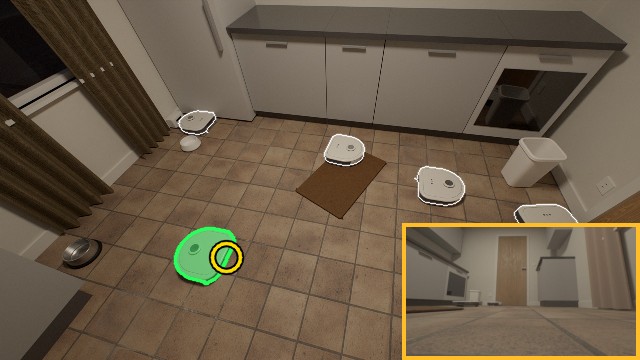}};
\node[galchip] at (10.638,3.902) {robot vacuum};
\node[anchor=north west, inner sep=0pt] at (14.044,3.902) {\includegraphics[width=3.326cm]{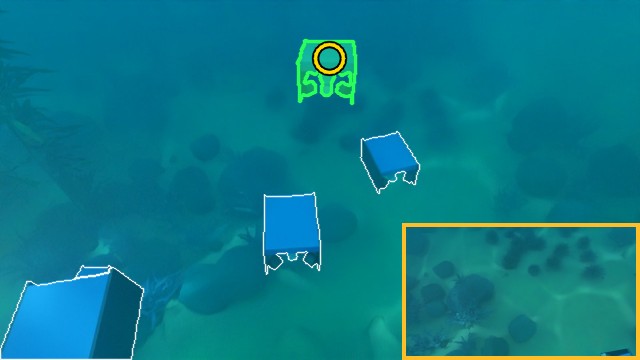}};
\node[galchip] at (14.044,3.902) {ROV};
\node[galrow] at (0.190,6.788) {Animals};
\node[anchor=north west, inner sep=0pt] at (0.420,5.853) {\includegraphics[width=3.326cm]{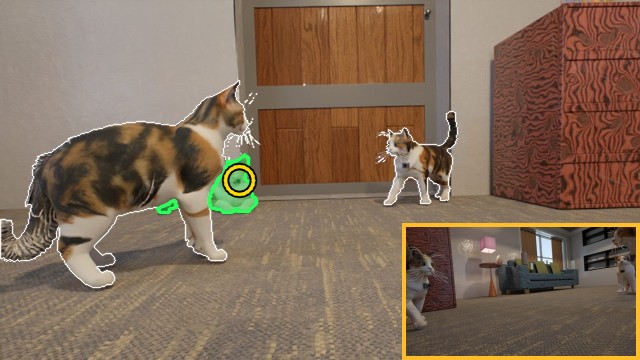}};
\node[galchip] at (0.420,5.853) {pets};
\node[anchor=north west, inner sep=0pt] at (3.826,5.853) {\includegraphics[width=3.326cm]{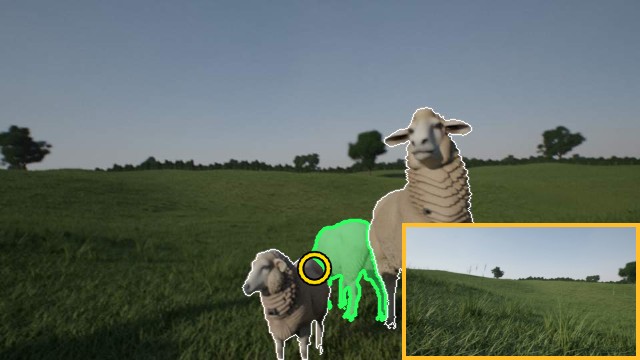}};
\node[galchip] at (3.826,5.853) {wild animals};
\node[anchor=north west, inner sep=0pt] at (7.232,5.853) {\includegraphics[width=3.326cm]{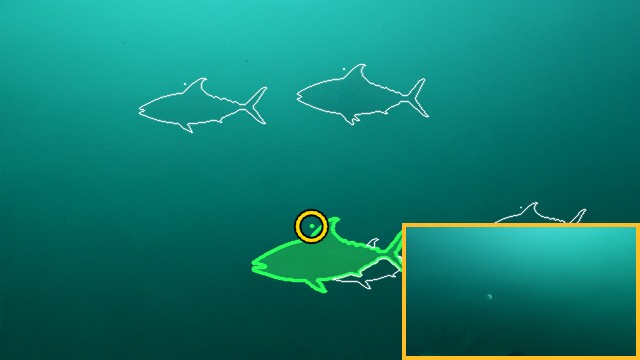}};
\node[galchip] at (7.232,5.853) {fish};
\node[anchor=north west, inner sep=0pt] at (10.638,5.853) {\includegraphics[width=3.326cm]{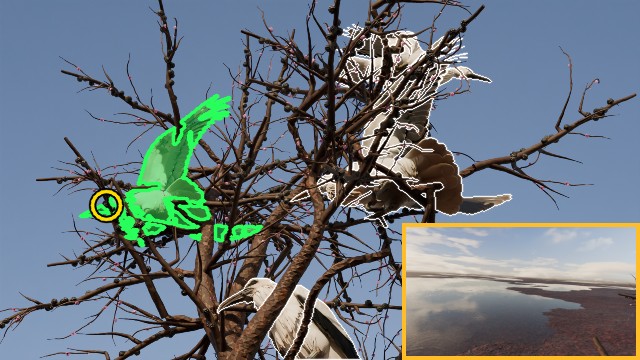}};
\node[galchip] at (10.638,5.853) {birds, perching};
\node[anchor=north west, inner sep=0pt] at (14.044,5.853) {\includegraphics[width=3.326cm]{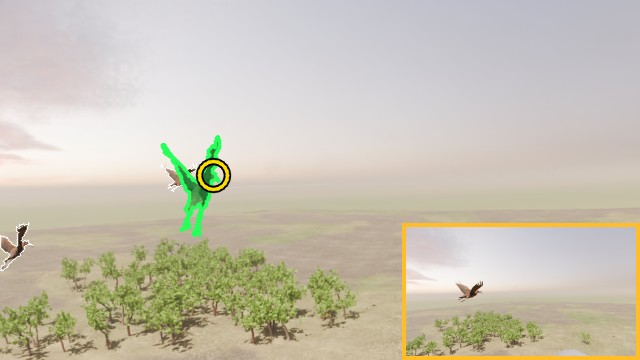}};
\node[galchip] at (14.044,5.853) {birds, flying};
\end{tikzpicture}
\caption{\textbf{Examples from \dataset.} Gold test queries from real recordings (top row) and simulated humans, robots, and animals. Each tile crops the exo image around the green-filled wearer mask, outlines other candidates in white, marks the true epipole with a yellow ring, and insets the ego image; labels give wearer type (ROV: remotely operated vehicle) and source or domain. Tiles are chosen by eye from each set's first queries in query-key hash order. Real images: CoMind~\cite{comind2026}, EgoHumans~\cite{khirodkar2023egohumans}, V2X-Real~\cite{xiang2024v2xreal}, DROID~\cite{khazatsky2024droid}, and MILUV~\cite{shalaby2026miluv}.}
\label{fig:gallery_data}
\end{figure*}

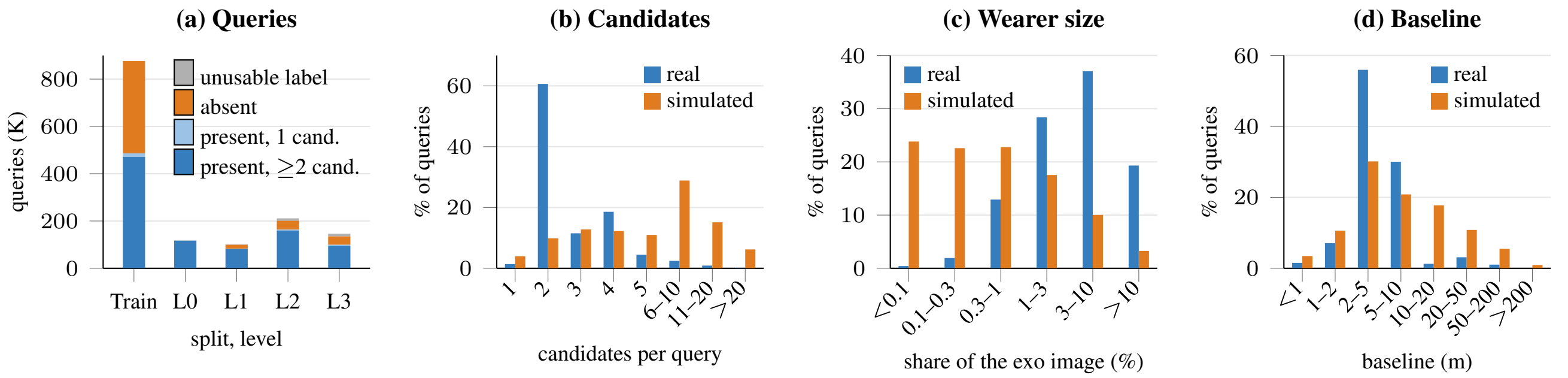
\begin{figure*}[t]
\centering
\definecolor{statREAL}{HTML}{367DBD}\definecolor{statSIM}{HTML}{E07B1F}
\definecolor{statTWO}{HTML}{367DBD}\definecolor{statONE}{HTML}{9CC3E6}\definecolor{statABS}{HTML}{E07B1F}\definecolor{statIGN}{HTML}{B0B0B0}
\begin{tikzpicture}
\pgfplotsset{statbase/.style={scale only axis, width=2.95cm, height=2.35cm, axis lines*=left, tick align=outside,
  tick style={black!60}, every tick label/.style={font=\scriptsize}, label style={font=\scriptsize}, title style={font=\footnotesize\bfseries,
  yshift=-2pt}, ymajorgrids, grid style={black!10}, xtick pos=bottom, ytick pos=left, ylabel shift=-3pt}}
\begin{axis}[name=pa, at={(0cm,0cm)}, anchor=south west, statbase, ybar stacked, bar width=7pt, ymin=0, ymax=900,
  xtick={0,1,2,3,4}, xticklabels={Train,L0,L1,L2,L3}, enlarge x limits=0.15, ylabel={queries (K)}, xlabel={split, level},
  title={(a) Queries}, legend style={font=\scriptsize, draw=none, fill=none, at={(1.0,1.0)}, anchor=north east, row sep=-2pt,
  inner sep=1pt}, legend cell align=left, reverse legend]
\addplot[fill=statTWO, draw=none] coordinates {(0,470.8) (1,116.5) (2,81.1) (3,159.6) (4,94.6)};
\addlegendentry{present, ${\geq}$2 cand.}
\addplot[fill=statONE, draw=none] coordinates {(0,15.4) (1,1.3) (2,2.1) (3,3.8) (4,4.7)};
\addlegendentry{present, 1 cand.}
\addplot[fill=statABS, draw=none] coordinates {(0,390.2) (1,0.4) (2,16.2) (3,37.4) (4,35.6)};
\addlegendentry{absent}
\addplot[fill=statIGN, draw=none] coordinates {(0,0.0) (1,0.9) (2,2.2) (3,10.4) (4,11.4)};
\addlegendentry{unusable label}
\end{axis}
\begin{axis}[name=pb, at={(4.35cm,0cm)}, anchor=south west, statbase, ybar=0pt, bar width=3.2pt, ymin=0, ymax=70,
  xtick={0,1,2,3,4,5,6,7}, xticklabels={{1},{2},{3},{4},{5},{6--10},{11--20},{${>}$20}}, xticklabel style={rotate=45, anchor=east, inner sep=1pt}, enlarge x limits=0.08,
  ylabel={\% of queries}, xlabel={candidates per query}, title={(b) Candidates}, legend style={font=\scriptsize, draw=none, fill=none, at={(1.0,1.0)}, anchor=north east, row sep=-2pt, inner sep=1pt}, legend cell align=left, legend image code/.code={\draw[#1, draw=none] (0cm,-0.08cm) rectangle (0.18cm,0.1cm);}]
\addplot[fill=statREAL, draw=none] coordinates {(0,1.38) (1,60.64) (2,11.51) (3,18.57) (4,4.44) (5,2.44) (6,0.89) (7,0.15)};
\addplot[fill=statSIM, draw=none] coordinates {(0,3.95) (1,9.86) (2,12.78) (3,12.25) (4,10.98) (5,28.85) (6,15.12) (7,6.21)};
\legend{real, simulated}
\end{axis}
\begin{axis}[name=pc, at={(8.7cm,0cm)}, anchor=south west, statbase, ybar=0pt, bar width=3.2pt, ymin=0, ymax=40,
  xtick={0,1,2,3,4,5}, xticklabels={{${<}$0.1},{0.1--0.3},{0.3--1},{1--3},{3--10},{${>}$10}}, xticklabel style={rotate=45, anchor=east, inner sep=1pt}, enlarge x limits=0.08,
  ylabel={\% of queries}, xlabel={share of the exo image (\%)}, title={(c) Wearer size}, legend style={font=\scriptsize, draw=none, fill=none, at={(0.04,1.0)}, anchor=north west, row sep=-2pt, inner sep=1pt}, legend cell align=left, legend image code/.code={\draw[#1, draw=none] (0cm,-0.08cm) rectangle (0.18cm,0.1cm);}]
\addplot[fill=statREAL, draw=none] coordinates {(0,0.42) (1,1.93) (2,12.92) (3,28.39) (4,37.03) (5,19.31)};
\addplot[fill=statSIM, draw=none] coordinates {(0,23.82) (1,22.58) (2,22.78) (3,17.53) (4,10.02) (5,3.27)};
\legend{real, simulated}
\end{axis}
\begin{axis}[name=pd, at={(13.05cm,0cm)}, anchor=south west, statbase, ybar=0pt, bar width=3.2pt, ymin=0, ymax=60,
  xtick={0,1,2,3,4,5,6,7}, xticklabels={{${<}$1},{1--2},{2--5},{5--10},{10--20},{20--50},{50--200},{${>}$200}}, xticklabel style={rotate=45, anchor=east, inner sep=1pt}, enlarge x limits=0.08,
  ylabel={\% of queries}, xlabel={baseline (m)}, title={(d) Baseline}, legend style={font=\scriptsize, draw=none, fill=none, at={(1.0,1.0)}, anchor=north east, row sep=-2pt, inner sep=1pt}, legend cell align=left, legend image code/.code={\draw[#1, draw=none] (0cm,-0.08cm) rectangle (0.18cm,0.1cm);}]
\addplot[fill=statREAL, draw=none] coordinates {(0,1.52) (1,7.09) (2,55.95) (3,30.04) (4,1.27) (5,3.11) (6,1.03) (7,0.00)};
\addplot[fill=statSIM, draw=none] coordinates {(0,3.46) (1,10.60) (2,30.14) (3,20.83) (4,17.76) (5,10.82) (6,5.47) (7,0.92)};
\legend{real, simulated}
\end{axis}
\end{tikzpicture}
\caption{\textbf{Statistics of \dataset.} (a) Training and test queries by generalization level and query state on the lists of \cref{tab:data} (876,291 training, 578,046 test; the Full tier adds the two sets that \cref{tab:data} omits). (b)--(d) Present test queries, real versus simulated: (b) candidate count; (c) wearer mask area as a share of the exo image, excluding the 15K box-annotated queries (from stored render pixel counts, and for real sets and thermal renders from the masks of a query-key hash sample of at most 2,000 queries per set, weighted by set size); (d) camera baseline where both camera poses exist (82K real, 205K simulated).}
\label{fig:stats_data}
\end{figure*}

\paragraph{Examples and statistics.}
\Cref{fig:gallery_data} shows Gold test queries from real recordings and simulated humans, robots, and animals. \Cref{fig:stats_data} splits the query lists into training and the four generalization levels and shows the distributions of candidate count, wearer size in the exo image, and camera baseline. Simulated queries have more candidates, smaller wearers, and a wider baseline range than real ones.

\paragraph{Label construction for real data.}
Label construction depends on the annotations and metadata available for each real dataset; \cref{tab:audit} reports the audits of the masks and of the projected camera centers for the listed sets. In Ego-Exo4D, SAM~3 tracks humans, queries with object tracks that SAM~3 mistook for humans are dropped (\cref{sec:supp_audit}), and the Aria~\cite{engel2023projectaria} pose projected via each GoPro's calibration identifies the wearer. In EgoHumans, the released keypoints are projected through the fisheye model and matched to tracks per frame; wearers without official poses are excluded. In CoMind, an LLM votes on 16 moments per track (requiring at least 70\% agreement and a binomial test), and Aria Machine Perception Services (MPS) trajectories solved against ViTPose~\cite{xu2022vitpose} eye keypoints provide projected camera centers. CORE4D-Real approximates them from SMPL-X~\cite{pavlakos2019smplx} eyes with a mount fitted per day. In Ego4D-TF, one wearer's camera serves as the peer exo camera, and labels are pooled over appearance clusters within a game.

\begin{table*}[t]
\centering
\caption{\textbf{Audit of real sets.} Wearer: how a track is linked to a camera. Mask audit: visual checks of sampled frames.}
\label{tab:audit}
\footnotesize
\setlength{\tabcolsep}{4pt}
\begin{tabular}{lp{2.3cm}p{4.1cm}p{3.9cm}p{3.6cm}}
\toprule
Set & Masks & Wearer & Mask audit & Projected camera centers \\
\midrule
TF2023 & manual polygons & given, identities corrected in scene 27, 45 test frames with an unannotated wearer set aside, not scored & \no & none \\
Ego-Exo4D & SAM~3 tracks & Aria pose projected with each GoPro's calibration & wearer about 38/40 clean, others about 36/40 & 97.9\% on the wearer mask \\
EgoHumans & SAM~3 tracks & official keypoints through the fisheye model, matched per frame & wearer 40/40, others 39/40 (lego, fencing); wearer 38/40 (tagging) & 7\,cm from the eye midpoint, median 11.3\,px at 1920\,px \\
CoMind & SAM~3 tracks & LLM votes per track, at least 70\% agreement & 131 of 160 views high confidence & 82.8\% good, 11.7\% fair, 5.5\% weak \\
CORE4D-Real & automatic & dataset labels & identities checked by eye & approximate: 64.5\% near the wearer, 19.4\% far \\
Ego4D-TF & SAM~3 tracks & automatic, pooled over appearance clusters, and Gold by one annotator & 142/152 tracks agree, about 6 of 400 crops show someone else & none \\
DROID & SAM~3 and LLM & single arm per view & 12 of 52 frames faulty, mostly wrist mounts & on the mask 71.0\%, within 10\,px 93.1\% \\
\bottomrule
\end{tabular}
\end{table*}

\begin{figure*}[t]
\centering
\begin{tikzpicture}[x=1cm, y=-1cm]
\node[anchor=west, inner sep=0pt, font=\footnotesize\bfseries, align=left] at (0,0.775) {Head};
\begin{scope}\clip[rounded corners=2.5pt] (1.450,0.000) rectangle ++(1.550,1.550);\node[anchor=north west, inner sep=0pt] at (1.450,0.000) {\includegraphics[height=1.550cm]{figs/lens/lens_head_v1.jpg}};\end{scope}
\node[anchor=base, inner sep=0pt, font=\scriptsize] at (2.225,1.820) {v1 pinhole};
\node[anchor=base, inner sep=0pt, font=\scriptsize, text=black!70] at (2.225,2.090) {100$^\circ$};
\begin{scope}\clip[rounded corners=2.5pt] (3.651,0.000) rectangle ++(2.756,1.550);\node[anchor=north west, inner sep=0pt] at (3.651,0.000) {\includegraphics[height=1.550cm]{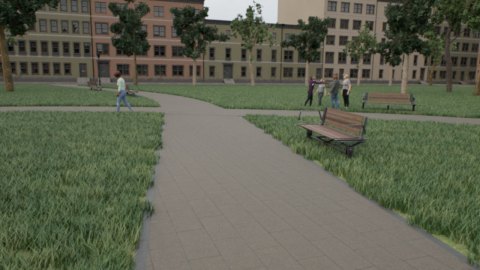}};\end{scope}
\node[anchor=base, inner sep=0pt, font=\scriptsize] at (5.029,1.820) {AR headset};
\node[anchor=base, inner sep=0pt, font=\scriptsize, text=black!70] at (5.029,2.090) {66$^\circ$};
\begin{scope}\clip[rounded corners=2.5pt] (7.057,0.000) rectangle ++(1.163,1.550);\node[anchor=north west, inner sep=0pt] at (7.057,0.000) {\includegraphics[height=1.550cm]{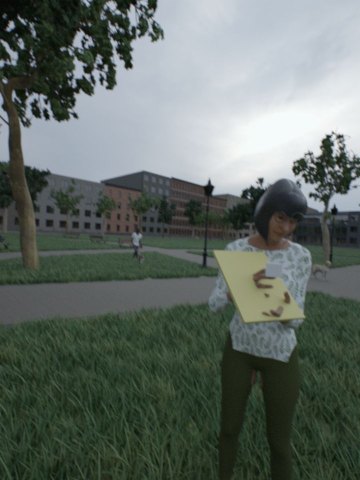}};\end{scope}
\node[anchor=base, inner sep=0pt, font=\scriptsize] at (7.638,1.820) {smart glasses};
\node[anchor=base, inner sep=0pt, font=\scriptsize, text=black!70] at (7.638,2.090) {73$^\circ$};
\begin{scope}\clip[rounded corners=2.5pt] (8.871,0.000) rectangle ++(1.550,1.550);\node[anchor=north west, inner sep=0pt] at (8.871,0.000) {\includegraphics[height=1.550cm]{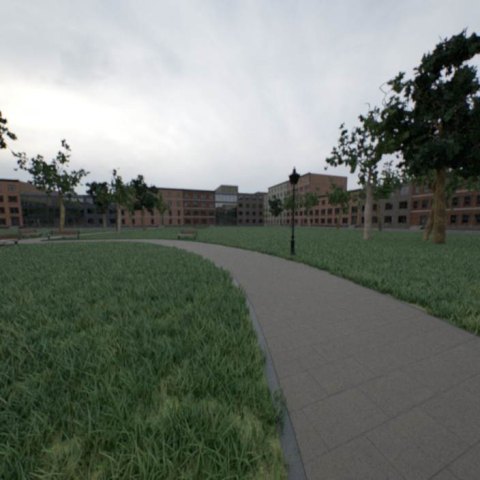}};\end{scope}
\node[anchor=base, inner sep=0pt, font=\scriptsize] at (9.646,1.820) {wide rectilinear};
\node[anchor=base, inner sep=0pt, font=\scriptsize, text=black!70] at (9.646,2.090) {105$^\circ$};
\begin{scope}\clip[rounded corners=2.5pt] (11.071,0.000) rectangle ++(1.550,1.550);\node[anchor=north west, inner sep=0pt] at (11.071,0.000) {\includegraphics[height=1.550cm]{figs/lens/lens_head_aria.jpg}};\end{scope}
\node[anchor=base, inner sep=0pt, font=\scriptsize] at (11.846,1.820) {Aria fisheye};
\node[anchor=base, inner sep=0pt, font=\scriptsize, text=black!70] at (11.846,2.090) {116$^\circ$};
\begin{scope}\clip[rounded corners=2.5pt] (13.272,0.000) rectangle ++(2.756,1.550);\node[anchor=north west, inner sep=0pt] at (13.272,0.000) {\includegraphics[height=1.550cm]{figs/lens/lens_head_gopro.jpg}};\end{scope}
\node[anchor=base, inner sep=0pt, font=\scriptsize] at (14.650,1.820) {GoPro on head};
\node[anchor=base, inner sep=0pt, font=\scriptsize, text=black!70] at (14.650,2.090) {118$^\circ$};
\node[anchor=west, inner sep=0pt, font=\footnotesize\bfseries, align=left] at (0,3.065) {Chest};
\begin{scope}\clip[rounded corners=2.5pt] (1.450,2.290) rectangle ++(2.756,1.550);\node[anchor=north west, inner sep=0pt] at (1.450,2.290) {\includegraphics[height=1.550cm]{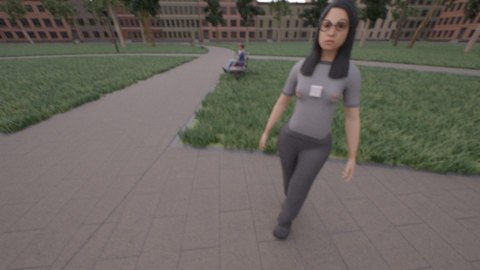}};\end{scope}
\node[anchor=base, inner sep=0pt, font=\scriptsize] at (2.828,4.110) {v1 pinhole};
\node[anchor=base, inner sep=0pt, font=\scriptsize, text=black!70] at (2.828,4.380) {110$^\circ$};
\begin{scope}\clip[rounded corners=2.5pt] (4.406,2.290) rectangle ++(2.756,1.550);\node[anchor=north west, inner sep=0pt] at (4.406,2.290) {\includegraphics[height=1.550cm]{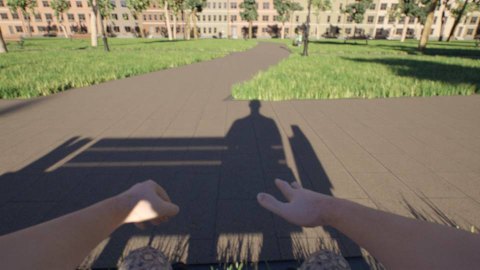}};\end{scope}
\node[anchor=base, inner sep=0pt, font=\scriptsize] at (5.783,4.110) {linear};
\node[anchor=base, inner sep=0pt, font=\scriptsize, text=black!70] at (5.783,4.380) {88$^\circ$};
\begin{scope}\clip[rounded corners=2.5pt] (7.361,2.290) rectangle ++(2.756,1.550);\node[anchor=north west, inner sep=0pt] at (7.361,2.290) {\includegraphics[height=1.550cm]{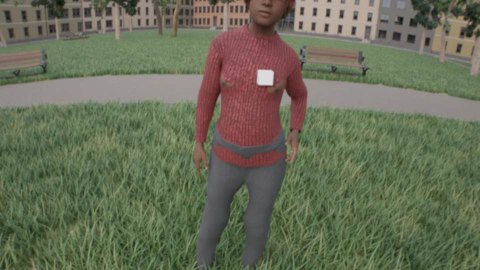}};\end{scope}
\node[anchor=base, inner sep=0pt, font=\scriptsize] at (8.739,4.110) {TF2023-style wide};
\begin{scope}\clip[rounded corners=2.5pt] (10.317,2.290) rectangle ++(2.756,1.550);\node[anchor=north west, inner sep=0pt] at (10.317,2.290) {\includegraphics[height=1.550cm]{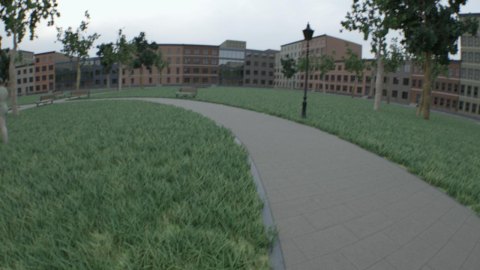}};\end{scope}
\node[anchor=base, inner sep=0pt, font=\scriptsize] at (11.694,4.110) {GoPro wide};
\node[anchor=base, inner sep=0pt, font=\scriptsize, text=black!70] at (11.694,4.380) {124$^\circ$};
\begin{scope}\clip[rounded corners=2.5pt] (13.272,2.290) rectangle ++(2.756,1.550);\node[anchor=north west, inner sep=0pt] at (13.272,2.290) {\includegraphics[height=1.550cm]{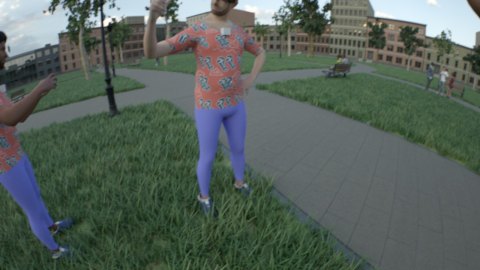}};\end{scope}
\node[anchor=base, inner sep=0pt, font=\scriptsize] at (14.650,4.110) {body camera};
\node[anchor=base, inner sep=0pt, font=\scriptsize, text=black!70] at (14.650,4.380) {132$^\circ$};
\node[anchor=west, inner sep=0pt, font=\footnotesize\bfseries, align=left] at (0,5.355) {Exo};
\begin{scope}\clip[rounded corners=2.5pt] (1.450,4.580) rectangle ++(2.756,1.550);\node[anchor=north west, inner sep=0pt] at (1.450,4.580) {\includegraphics[height=1.550cm]{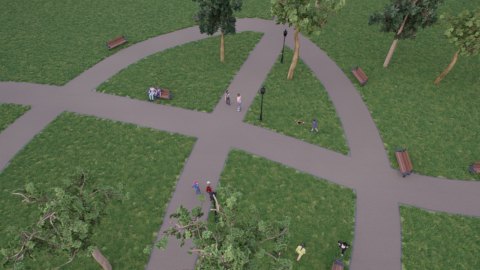}};\end{scope}
\node[anchor=base, inner sep=0pt, font=\scriptsize] at (2.828,6.400) {v1 pinhole};
\node[anchor=base, inner sep=0pt, font=\scriptsize, text=black!70] at (2.828,6.670) {77$^\circ$};
\begin{scope}\clip[rounded corners=2.5pt] (5.391,4.580) rectangle ++(2.756,1.550);\node[anchor=north west, inner sep=0pt] at (5.391,4.580) {\includegraphics[height=1.550cm]{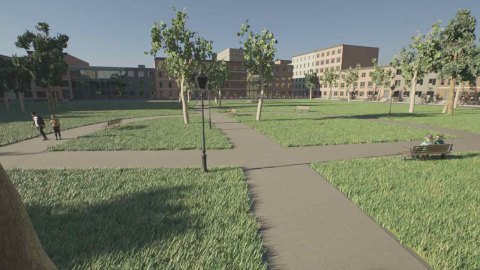}};\end{scope}
\node[anchor=base, inner sep=0pt, font=\scriptsize] at (6.769,6.400) {tripod pinhole};
\node[anchor=base, inner sep=0pt, font=\scriptsize, text=black!70] at (6.769,6.670) {94$^\circ$};
\begin{scope}\clip[rounded corners=2.5pt] (9.331,4.580) rectangle ++(2.756,1.550);\node[anchor=north west, inner sep=0pt] at (9.331,4.580) {\includegraphics[height=1.550cm]{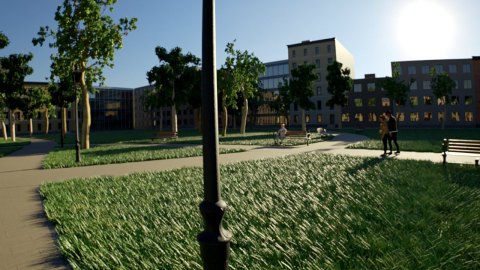}};\end{scope}
\node[anchor=base, inner sep=0pt, font=\scriptsize] at (10.709,6.400) {CCTV wide};
\begin{scope}\clip[rounded corners=2.5pt] (13.272,4.580) rectangle ++(2.756,1.550);\node[anchor=north west, inner sep=0pt] at (13.272,4.580) {\includegraphics[height=1.550cm]{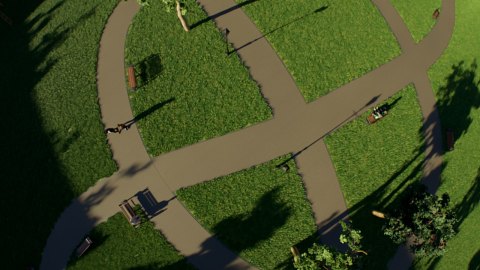}};\end{scope}
\node[anchor=base, inner sep=0pt, font=\scriptsize] at (14.650,6.400) {tripod GoPro fisheye};
\node[anchor=base, inner sep=0pt, font=\scriptsize, text=black!70] at (14.650,6.670) {116$^\circ$};
\node[anchor=west, inner sep=0pt, font=\footnotesize\bfseries, align=left] at (0,7.645) {Other\\wearers};
\begin{scope}\clip[rounded corners=2.5pt] (1.450,6.870) rectangle ++(2.756,1.550);\node[anchor=north west, inner sep=0pt] at (1.450,6.870) {\includegraphics[height=1.550cm]{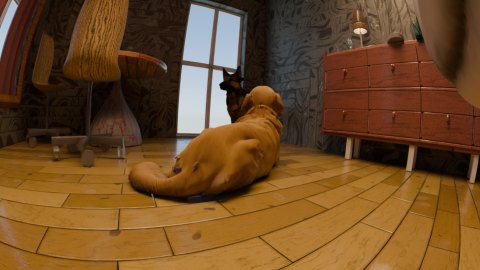}};\end{scope}
\node[anchor=base, inner sep=0pt, font=\scriptsize] at (2.828,8.690) {pet collar fisheye};
\node[anchor=base, inner sep=0pt, font=\scriptsize, text=black!70] at (2.828,8.960) {145$^\circ$};
\begin{scope}\clip[rounded corners=2.5pt] (5.391,6.870) rectangle ++(2.756,1.550);\node[anchor=north west, inner sep=0pt] at (5.391,6.870) {\includegraphics[height=1.550cm]{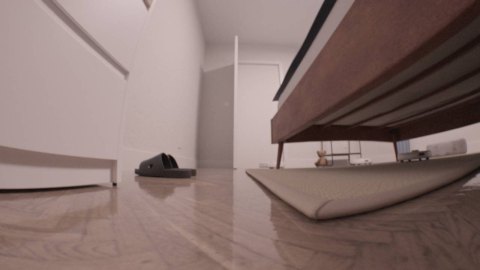}};\end{scope}
\node[anchor=base, inner sep=0pt, font=\scriptsize] at (6.769,8.690) {vacuum fisheye};
\node[anchor=base, inner sep=0pt, font=\scriptsize, text=black!70] at (6.769,8.960) {151$^\circ$};
\begin{scope}\clip[rounded corners=2.5pt] (9.331,6.870) rectangle ++(2.756,1.550);\node[anchor=north west, inner sep=0pt] at (9.331,6.870) {\includegraphics[height=1.550cm]{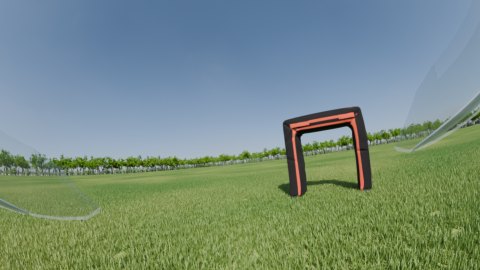}};\end{scope}
\node[anchor=base, inner sep=0pt, font=\scriptsize] at (10.709,8.690) {FPV drone fisheye};
\node[anchor=base, inner sep=0pt, font=\scriptsize, text=black!70] at (10.709,8.960) {146$^\circ$};
\begin{scope}\clip[rounded corners=2.5pt] (13.272,6.870) rectangle ++(2.756,1.550);\node[anchor=north west, inner sep=0pt] at (13.272,6.870) {\includegraphics[height=1.550cm]{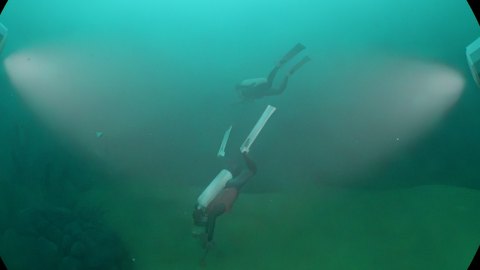}};\end{scope}
\node[anchor=base, inner sep=0pt, font=\scriptsize] at (14.650,8.690) {diver, dome port};
\node[anchor=base, inner sep=0pt, font=\scriptsize, text=black!70] at (14.650,8.960) {161$^\circ$};
\end{tikzpicture}
\caption{\textbf{Mixed lenses.} Head, chest, and exo profiles on one city scene, and other wearers in the bottom row. Labels give device and, where the profile fixes it, horizontal field of view.}
\label{fig:lens}
\end{figure*}

\paragraph{Lens profiles.}
Simulated lens profiles (\cref{fig:lens}) are fitted to real calibrations, with intrinsics jittered per device. The most common head cameras are Aria-style fisheyes (45\%), followed by GoPros, smart glasses, and AR headsets; chest cameras use wide, GoPro, body-camera, and linear lenses, and exo cameras are mostly pinholes (70\%) plus GoPro fisheyes and wide CCTV lenses. Fisheyes are rendered as equidistant projections and remapped so that RGB, masks, and depth stay aligned, while robots use their real sensor models and underwater cameras use flat and dome ports; \cref{sec:supp_lens} reports how the lenses affect accuracy.

\paragraph{Quality control.}
In our renders, we drop views with a lens inside geometry, dark views, and views dominated by one body. We review every set at full resolution and withdraw or re-render defective episodes; known defects stay out of Gold, while Full keeps every test query.

\paragraph{Pretraining overlap.}
\label{sec:supp_overlap}
We check the evaluated models' documentation for test-set overlap; because several also use unnamed or unreleased data, only named overlap can be ruled out. Pi3X and Depth Anything~3 train on parts of OmniWorld~\cite{zhou2026omniworld}, which also includes Ego-Exo4D and DROID. Depth Anything~3 lists 1,039 OmniWorld scenes as synthetic (matching only the game part's count) without saying which part it uses; its Ego-Exo4D scores match those of other backbones, so we find no sign of memorized recordings, although this cannot rule them out. The other 3D models' papers, model cards, repositories, and training configurations name none of our real or third-party test sets. On assets and related domains, \vggto and CUT3R train on BEDLAM and Depth Anything~3 on ABO objects, reused in our renders as assets rather than images; \vggto also trains on Aria Digital Twin~\cite{pan2023adt} (recorded with the device of some of our ego cameras), Aria Synthetic Environments~\cite{avetisyan2024scenescript} (simulating that device), and driving and synthetic flight data close to our vehicle and drone domains. SAM~3, which proposes Track B masks and labels DROID, was trained on Ego4D and DROID stills. On our own renders alone (one input order for all), zero-shot \vggto still leads at 54.5 seen and 53.3 unseen \ka versus 45.3 and 43.7 for the next-best backbones.

\paragraph{License and release.}
About 50\% of the pairs are our own renders; all except the Isaac Sim renders---whose assets NVIDIA does not permit us to redistribute---can be released (47\% of the pairs). Among real sources, DROID and MILUV are under CC BY 4.0. Ego-Exo4D and Ego4D require their own agreements, and TF2023 does not allow image redistribution, although its authors agreed to the release of our labels and corrections. For these three sources, we release labels, query lists, answer keys, and download and conversion scripts, so users obtain the images under each source's terms. The data card states that BEDLAM and SMPL-X bodies are licensed for non-commercial use.

\subsection{Blind audit and corrections}
\label{sec:supp_audit}

\begin{figure*}[t]
\centering
\includegraphics[width=0.82\textwidth]{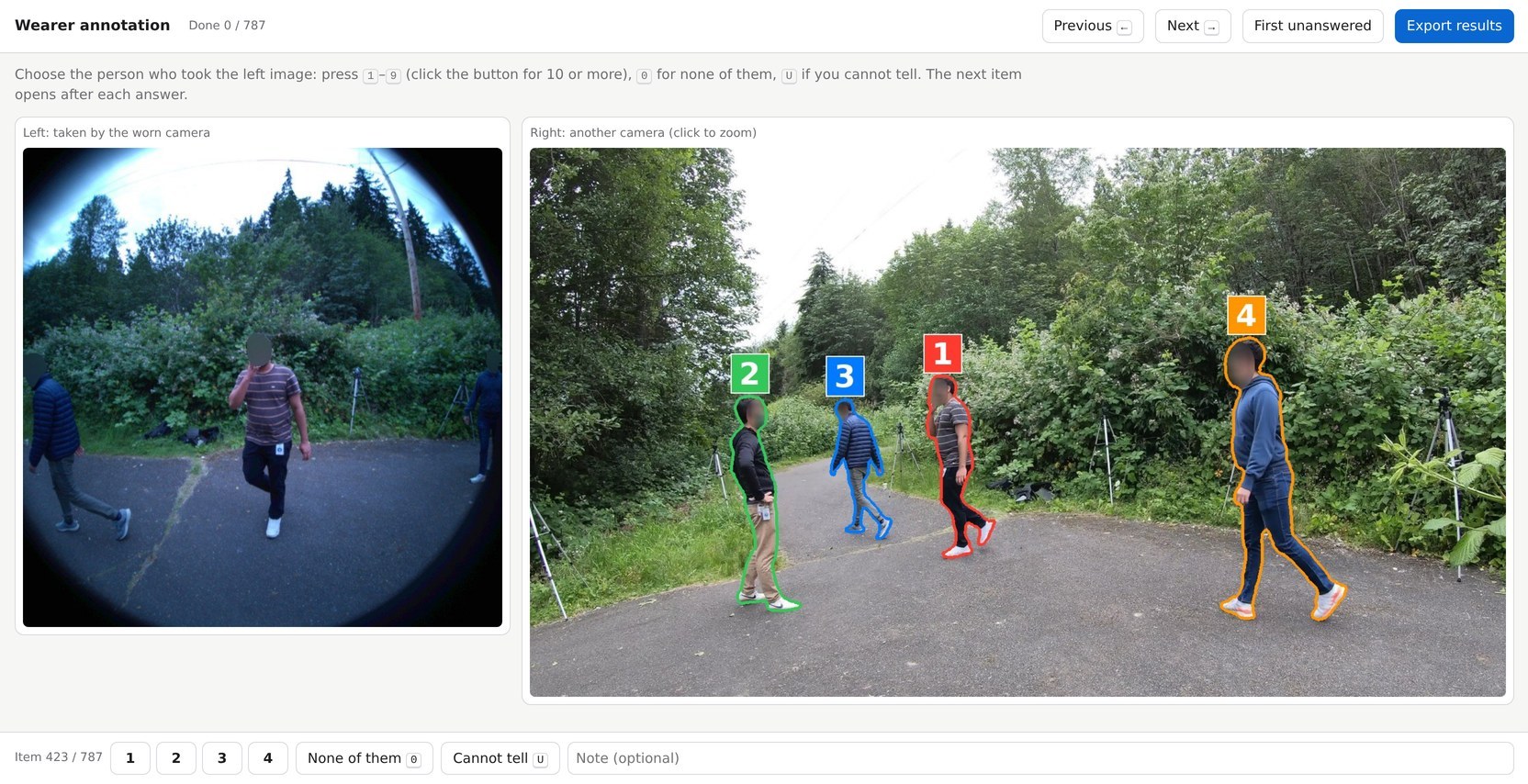}
\caption{\textbf{Blind audit interface.} The page used by the two annotators (\cref{sec:supp_audit}), shown on one EgoHumans~\cite{khirodkar2023egohumans} query with its interface text translated into English. The ego image is on the left and the exo image on the right, with every candidate outlined and numbered in random order and no label, projection, set name, or model output. Annotators answer with a number, \emph{none of them}, or \emph{cannot tell}; Ego4D-TF and TF2023 reviews added one option, \emph{visible but not outlined}.}
\label{fig:review_ui}
\end{figure*}

\paragraph{Blind audit.}
The blind audit covers human wearers only: two annotators who had not seen the automatic labels answered 787 queries of the seven real human sets, 700 drawn uniformly and 87 hard ones with another mask within $2^\circ$ of the true epipole. Candidates were outlined and numbered in random order (\cref{fig:review_ui}), whereas LLM items use numbered boxes. The annotators agree with Cohen's $\kappa$ of 0.86 on uniform and 0.92 on hard queries; where both agree, the automatic label is wrong on 2 of 631 uniform queries (0.3\%, interval 0.1--1.2\%) and 1 of 83 hard ones. \Cref{tab:audit2} compares the annotators with our models on the 580 queries of CoMind, CORE4D-Real, EgoHumans, Harmony4D, and TF2023.

\paragraph{Ego4D-TF and TF2023.}
One annotator rechecked all 702 Gold-tier Ego4D-TF queries without seeing the labels, deciding 651 and agreeing with the automatic label on 91.1\% [88.4, 92.8]. Gold follows the annotator, changing 49 labels and setting aside 60 queries that were undecided or lacked a wearer mask; these remain in the list unscored.

TF2023 has manual polygons and wearer IDs. Two annotators blind to the labels reviewed 714 frames (Cohen's $\kappa=0.90$): model-flagged segments, frames with an unannotated wearer, and 10 random frames per scene. They confirm the two-identity swap in scene 27; beyond it, 2 test labels change and 45 frames with an unannotated wearer are dropped, together 0.22\% of the test set. On random frames with an annotated wearer, 0.59\% [0.0, 1.49] of labels are wrong.

\paragraph{Corrections.}
The audit found two automatic-label faults; all Gold results in this paper use the corrected lists, and the corrections leave the main result unchanged. First, because Ego-Exo4D candidates included objects tracked as humans (such as a plate in one query), we dropped queries with any candidate track found by a person detector in under half its sampled frames, cutting the Ego-Exo4D part of Gold from 1,200 to 1,089 queries. Second, two CORE4D-Real sequences pair ego and exo frames from different moments (727 $\to$ 722 Gold queries). We keep TF2023's published polygons for comparability with PEN.

\begin{table}[!htb]
\centering
\caption{\textbf{Annotators and our models on the audit queries.} Acc against the dataset labels (manual ones for TF2023). Seconds: median time per query.}
\label{tab:audit2}
\footnotesize
\setlength{\tabcolsep}{4pt}
\begin{tabular}{lccc}
\toprule
Answered by & Queries & Acc & Seconds \\
\midrule
Annotator A & 580 & 95.2 & 4.8 \\
Annotator B & 580 & 97.9 & 5.2 \\
\midrule
\rowcolor{tabourslight}
\ours-ZS & 580 & 93.6 & 0.2 \\
\ours & 580 & 98.4 & 0.4 \\
\bottomrule
\end{tabular}
\end{table}

\subsection{Realism of the simulation}
\label{sec:supp_realism}

For humans and vehicles, where real sets of the same wearer type exist, we run four checks from prior work (\cref{tab:realism}); because simulated vehicles come from public CARLA-based sets, only human sets test our own rendering. First, for sim-to-real predictivity (Kadian~\etal~\cite{kadian2020sim2real}), we correlate 24 methods' scores across simulated and real sets per wearer type. Simulated sets rank methods as real ones do, with every simulated human domain alone reaching $\tau\geq0.79$, including Unreal sets whose engine never appears in training. Half of the 24 are close \vggto variants; keeping only the twelve baselines not built on it and its camera head gives $\tau$ of 0.821 [0.577, 0.974] for humans and 0.744 [0.444, 0.972] for vehicles. Among trained models alone, the method rankings on real and simulated human sets agree poorly ($\tau=0.33$) because scores on real sets saturate at 95--99 \ka while simulated sets still separate the models.

Second, for the real-to-virtual gap (Virtual KITTI~\cite{gaidon2016virtual}), we compare a model on baseline-matched simulated and real queries. For \ours, matching narrows the human gap from $-$10.0 to $-$4.5 \ka, with matched direction errors differing by only 0.07$^\circ$ [$-$0.11, 0.30], and its vehicle gap is not significant; the zero-shot model loses more on simulated sets, which hold more distant and more numerous candidates, although after matching its vehicle gap is no longer significant. Third, in DINOv2 feature space~\cite{oquab2024dinov2} (recommended by Stein~\etal~\cite{stein2023exposing} for Fr\'echet distances), simulated-to-real distances are comparable to those between real datasets. Fourth, simulated queries cover most real geometric values and extend far beyond them: simulated human queries have a median of 7 candidates against 2 and a baseline of 11.8\,m against 3.8\,m, which makes simulation harder. These checks have further limits: trained models saw Blender humans in training, real human geometry comes from three sets with camera poses, and no person rated the images.

Two other wearer types have real sets with weaker agreement. Simulated robot arms rank methods only moderately like DROID ($\tau=0.58$), where the largest mask nearly always wins. Our aerial renders likewise agree only moderately with MILUV's real quadcopter recordings on the ranking of the 14 zero-shot methods (\cref{sec:supp_locked}; $\tau=0.58$, or 0.68 with pose-supervised training and \ours added, in a check run once); MILUV separates zero-shot methods poorly because it has two identical candidates per query in one arena.

For the remaining non-human wearers and for drones seen from the ground, no real set exists for the task, so for several of them we compare appearance alone against real images matched in camera pitch. Across these wearer types---drones and birds seen from the ground~\cite{jiang2023antiuav,svanstrom2021dataset}, herds seen from drones~\cite{kline2025mmla} and by camera traps~\cite{swanson2015snapshot,kline2025smartwilds}, and divers and fish~\cite{pedersen2019brackish,lian2023watermask,islam2020suim}---the kernel inception distance (KID) on DINOv2 features between our renders and real images of the same wearers is 0.39--1.15 times that between two real datasets of those wearers; a ratio below one means the real datasets differ more from each other than from our renders, not that the renders look more real.

\begin{table}[!htb]
\centering
\caption{\textbf{Realism of the simulation.} Rank: correlation over 24 methods. Gap: simulated minus real. FD ratio: simulated-to-real over real-to-real Fr\'echet distance. Coverage: lowest share of real values of seven geometric quantities within the simulated 1--99\% range.}
\label{tab:realism}
\footnotesize
\setlength{\tabcolsep}{3pt}
\resizebox{\ifdim\width>\columnwidth\columnwidth\else\width\fi}{!}{%
\begin{tabular}{lcc}
\toprule
Check & Humans & Vehicles \\
\midrule
Rank, Kendall $\tau$ & 0.82 [0.68, 0.92] & 0.86 [0.71, 0.95] \\
Rank, Pearson $r$ & 0.89 [0.84, 0.95] & 0.95 [0.90, 0.98] \\
\midrule
Gap of \ours, raw & $-$10.0 [$-$11.2, $-$8.6] & 0.9 [$-$8.6, 10.0] \\
Gap of \ours, matched & $-$4.5 [$-$5.6, $-$3.2] & $-$3.9 [$-$17.0, 7.1] \\
$\aex$ gap of \ours, matched & 0.07$^\circ$ [$-$0.11, 0.30] & $-$0.27$^\circ$ [$-$1.27, 0.45] \\
Gap of \ours-ZS, raw & $-$26.3 [$-$28.5, $-$23.8] & $-$13.3 [$-$23.7, $-$1.5] \\
Gap of \ours-ZS, matched & $-$10.9 [$-$12.8, $-$8.9] & $-$10.6 [$-$20.0, 0.8] \\
\midrule
FD ratio, exo / ego & 1.14 / 1.21 & 0.97 / 1.07 \\
Coverage & 93.3\% & 94.1\% \\
\bottomrule
\end{tabular}}
\end{table}

\subsection{Related benchmarks}
Whereas prior benchmarks each score one or two of the three tasks on humans or objects, \dataset scores all three, as well as absent wearers, across 17 domains (\cref{tab:benchmarks}).

\begin{table}[!htb]
\centering
\caption{\textbf{Related benchmarks.} ID / Seg / Loc: scores \who, \where (for Ego-Exo4D correspondence, of a queried object), or \exo. Absent: scores targets out of view.}
\label{tab:benchmarks}
\footnotesize
\setlength{\tabcolsep}{2.6pt}
\resizebox{\ifdim\width>\columnwidth\columnwidth\else\width\fi}{!}{%
\begin{tabular}{lccccll}
\toprule
Benchmark & ID & Seg & Loc & Absent & Targets & Size \\
\midrule
TF2023~\cite{zhao2024pen} & \cmark & \cmark & \no & \no & humans & 69K queries \\
TF2025~\cite{zhao2026sequence} & \cmark & \no & \no & \no & humans & 2.3$\times$ TF2023 \\
EgoExoBench, wearer identification~\cite{he2025egoexobench} & \cmark & \no & \no & \no & humans & part of 7.3K QA \\
CVMHAT~\cite{han2020cvmhat} & \cmark & \no & \no & \no & humans & 11.7K pairs \\
Ego-Exo4D corr.~\cite{grauman2024egoexo4d} & \no & \cmark & \no & \cmark & objects & 1.8M masks \\

\midrule
\dataset (ours) & \cmark & \cmark & \cmark & \cmark & 12 wearer types & 1.16M pairs \\
\bottomrule
\end{tabular}}
\end{table}

\section{Protocol and Metrics}
\label{sec:supp_protocol}

This section specifies the evaluation protocol on \dataset---what inputs a method receives for each query, which queries each track scores, and which test queries enter the reported results---and defines the metrics (\cref{sec:supp_metrics}). Main results use the Full tier (all test queries), while other experiments use the Gold tier, a fixed subset drawn without model output (\cref{sec:supp_gold}).

\paragraph{Inputs.}
For each query, the protocol gives a method the ego and exo images, plus a category word for \where or the candidates as instance masks for \who (as boxes on sets that provide only boxes: top views and vehicles). True projected camera centers are never given as candidates or other inputs, because on simulated data they would disclose exact camera positions that instance masks do not provide.

\paragraph{Rules fixed after auditing the data.}
The protocol enforces eight rules: (1) peer views form separate settings; (2) Tracks A and B ignore visible wearers covering less than $1/5000$ of the image; (3) Track A requires at least two candidates; (4) the body wearing a peer exo camera is excluded as a candidate, and one wearer's head and chest cameras are never paired; (5) category words are fixed per set, as per-query words would reveal answers in mixed sets; (6) real absent queries require a projected camera center outside the image, or a label marking the wearer absent on Ego4D-TF, which has no projected camera centers; (7) candidate order is shuffled once per query and then fixed, and no method sees identifiers, except that video experiments link candidates across frames by their tracks (\cref{sec:supp_video}); and (8) settings with fewer than 30 present queries with at least two candidates are listed but omitted from \who averages, since domains weight settings equally and in such a setting one query moves accuracy by over three points. Only the two robot-arm sets' peer views fall under this rule: the first has two such queries (each worth 17 points of the whole domain) and the second none. Averaging the first anyway would lower the unseen macro averages on the Gold tier from 72.6 to 71.2 for pose-supervised training and from 78.2 to 76.4 for \ours, and the adaptation gain from 5.5 to 5.3, so the rule slightly raises our reported gain. Two L3 domains also include human wearers: paragliders and skydivers form 47\% of drone queries and humans 16\% of legged-robot ones (\cref{sec:supp_full} reports the unseen gain without them).

Among all 14 settings outside the macro averages (listed in \cref{tab:excl}), DROID is excluded as too easy: on the Full tier, only 99 present queries have at least two candidates, and the largest-mask baseline (without the ego image) identifies the wearer in 99.0\% of them (chance 46.8\%). Ego-Exo4D's held-out sites remain in the macro averages even though the largest mask wins 99.4\% of their Gold queries; removing those sites moves seen \ka by at most 0.2. In \exo, seven sets without reliable camera poses lack exo-side directions, and the \exo macro averages use 92 settings. After freezing the data, we checked all query lists for missing or duplicate keys and confirmed that test, training, and development lists are disjoint.

\begin{table}[!htb]
\centering
\caption{\textbf{Settings outside the macro averages} (Full tier). Choice: present queries with at least two candidates.}
\label{tab:excl}
\footnotesize
\setlength{\tabcolsep}{3pt}
\resizebox{\ifdim\width>\columnwidth\columnwidth\else\width\fi}{!}{%
\begin{tabular}{llrrl}
\toprule
Setting & Level & Queries & Choice & Reason \\
\midrule
ALOHA & L3 & 540 & 540 & locked set \\
TUMTraf-V2X & L3 & 294 & 294 & locked set \\
MILUV & L3 & 998 & 998 & locked set \\
DROID & L3 & 586 & 99 & too easy \\
Robot arms v1, peer & L3 & 530 & 2 & rule (8), \who only \\
Robot arms v2, peer & L3 & 446 & 0 & rule (8), \who only \\
Thermal, RGB to RGB & L1 & 878 & 878 & control \\
EgoHumans, undistorted & L2 & 51,169 & 44,039 & the same frames, undistorted \\
CSRD-R & L2 & 2,309 & 2,309 & the recordings of CVMHAT \\
CVUTD, cyclists from a drone & L3 & 100 & 100 & qualitative only \\
First wild-animal set, v1 fixed & L3 & 1,163 & 667 & earlier version, less realistic \\
\quad v1 peer & L3 & 2,929 & 906 & earlier version, less realistic \\
\quad v2 fixed & L3 & 709 & 342 & earlier version, less realistic \\
\quad v2 peer & L3 & 2,178 & 536 & earlier version, less realistic \\
\bottomrule
\end{tabular}}
\end{table}

\paragraph{Absent wearers in the score.}
On the Ego-Exo4D correspondence leaderboard, the top-IoU entry in a participant report~\cite{egoexo4d2025challenge} judges visibility at chance (balanced accuracy 50\%), which scoring only visible frames cannot reveal.

\paragraph{Track B diagnostics.}
Outside the main score, diagnostic counterfactual queries swap the ego image for one from another scene, making the correct answer \emph{absent}. In the dataset, absent queries record whether the wearer is off-frame or fully occluded, though real data contain only the off-frame case. Because natural absence rates vary widely across domains (football 55\%, flocks 41\%, underwater 24\%, indoors 3\%), \giou fixes the absent-query share.

\subsection{Metrics}
\label{sec:supp_metrics}

\begin{table*}[t]
\centering
\caption{\textbf{All metrics.} Where each metric is reported: the main paper or listed supplement figures and tables.}
\label{tab:metrics}
\footnotesize
\setlength{\tabcolsep}{4pt}
\begin{tabular}{llp{9.0cm}l}
\toprule
Track & Metric & Definition & Reported in \\
\midrule
A & Acc, \ka & Share of correct picks, and $(\text{Acc}-c)/(1-c)$ with $c$ the mean of $1/N$ in the setting & main \\
A & Per-query \ka & Mean of $(\text{hit}-1/N)/(1-1/N)$, used for slices & \cref{fig:cross,tab:crossover,tab:crossdom} \\
A & Acc one-to-one & Acc after one-to-one assignment & \cref{tab:full,tab:comp} \\
A & PEN Acc, mIoU & Share of queries whose chosen mask has an IoU above 0.5, and the mean IoU, within PEN's detector pool (TF2023 only) & main \\
\midrule
B & mIoU-P, \giou & Mean IoU on present queries, and $0.8\,$mIoU-P $+0.2\,$N-acc & main \\
B & Pr@0.5 & Share of present queries with an IoU above 0.5 & main \\
B & VA & Mean of the accuracies of the presence verdict on absent and on present queries & main \\
B & N-acc, T-acc & Accuracy of the presence verdict on absent and on present queries & \cref{tab:presence} \\
\midrule
C & $\aex$ & Median of the exo-side direction error, with directions keeping their sign & main \\
C & $\aeg$, RE & Median of the ego-side direction error and the rotation error & \cref{tab:full,tab:exopooled} \\
C & MA@$\epsilon$ & Share with $\max(\aex,\aeg)<\epsilon$, $\epsilon=5^\circ,15^\circ$ & main \\
C & AUC@30, RRA@30, RTA@30 & Area under the accuracy curve of the larger of the rotation and translation-direction errors up to $30^\circ$, and shares of pairs with each error below $30^\circ$, as in VGGT, on signed translation directions, for comparison with the literature & \cref{tab:exopooled} \\
C & BLE & Median of $b\sin(\min(\aex,90^\circ))$, with $b$ the true baseline, in body widths (human and diver 0.50, vehicle 1.90, animal 0.80, bird 0.30, drone 0.50, robot 0.50, arm 0.15, and vacuum 0.35\,m); a set takes the width of the first wearer type it lists, and the pet sets, without a width, stay out of the BLE averages & main \\
\bottomrule
\end{tabular}
\end{table*}

\Cref{tab:metrics} lists the metrics and where each is reported. The minimum in \cref{eq:offset} runs over at most 3,000 sampled pixels per mask, and evaluation slices use query-level geometry, camera, wearer, and scene metadata. For scene-level confidence intervals, our bootstrap redraws each set's test scenes 2{,}000 times and takes the 2.5th and 97.5th percentiles of the recomputed value. Because a simulated set's two lens versions share scenes, each draw samples each scene once for both; paired differences between methods use the same draws with models and adapters fixed.

\subsection{Gold tier}
\label{sec:supp_gold}

Like ScanNet-1500~\cite{dai2017scannet,sarlin2020superglue,sun2021loftr}, the Gold tier is a fixed evaluation subset drawn without model output by taking each setting's first queries in query-key hash order. Per setting, it retains up to 600 queries with at least two candidates on real sets and up to 400 on simulated ones, up to 100 absent queries, and single-candidate queries at their natural share up to 100; the three locked sets enter it whole (\cref{sec:supp_locked}). Gold excludes real queries with low-confidence labels and episodes with a known defect, so DROID keeps 39 of its 99 present queries with at least two candidates. After the corrections in \cref{sec:supp_audit}, Gold holds 53,679 queries: 44,341 with at least two candidates, 1,058 with one, 8,220 absent, and 60 set aside by the Ego4D-TF review. As in \cref{sec:stats} and \cref{tab:data}, counts cover list members and thus include queries with unusable labels that are never scored: these 60 in Gold and 28,782 of the 636,194 test queries.

\begin{table}[!htb]
\centering
\caption{\textbf{Comparison of Gold and Full tiers} (zero-shot \vggto, 107 settings).}
\label{tab:gold}
\footnotesize
\begin{tabular}{lc}
\toprule
Check & Result \\
\midrule
Setting Acc, mean / maximum absolute difference & 1.1 / 4.1 \\
Macro difference, seen / unseen (14 readout rules) & 0.6 / 0.6 \\
Kendall $\tau$ of the readout-rule ranking, seen / unseen & 0.82 / 1.00 \\
Setting median $\aex$, mean / maximum difference & 1.4$^\circ$ / 11.5$^\circ$ \\
Setting bootstrap half-width, median / 90th pct. & 4.3 / 7.8 \\
\bottomrule
\end{tabular}
\end{table}

Identification agrees closely between Gold and Full (\cref{tab:gold,tab:goldfull}), so Gold tracks Full on queries with at least two candidates apart from the defects it removes on purpose. Because Gold caps absent and single-candidate queries per setting, metrics pooling them with others, such as MA@$\epsilon$, weight query states differently across tiers. Absent queries make up 18.8\% of unseen Track C queries on Gold versus 28.2\% on Full, which also retains the 5.8\% of queries whose wearer is too small to count or not clearly absent (rules 2 and 6 of \cref{sec:supp_protocol}); most of Full's extra absent queries come from peer settings, where the exo camera rarely sees the wearer. On absent queries, \ours scores about 20 points lower in MA@5, which explains its lower MA@5 on the Full tier; on present queries, and when Full is reweighted to Gold's mix of query states, the two tiers agree within 0.5 points. Across readout comparisons, rules that swap ranks differ by less than half a point, and the largest differences between tiers occur in small settings with heavy-tailed errors. Because one-to-one assignment uses every queried ego image in a frame, Gold-only runs compute it on fewer ego images and score about 2.4 points lower than Full.

\section{Readout Analysis}
\label{sec:supp_analysis}
We analyze our geometric readout by addressing four questions: when depth and direction each help, how pose error limits identification, what makes a domain difficult, and how our readout compares with other ways to identify or segment the wearer.

\subsection{Depth, direction, and camera position}
\label{sec:supp_read}

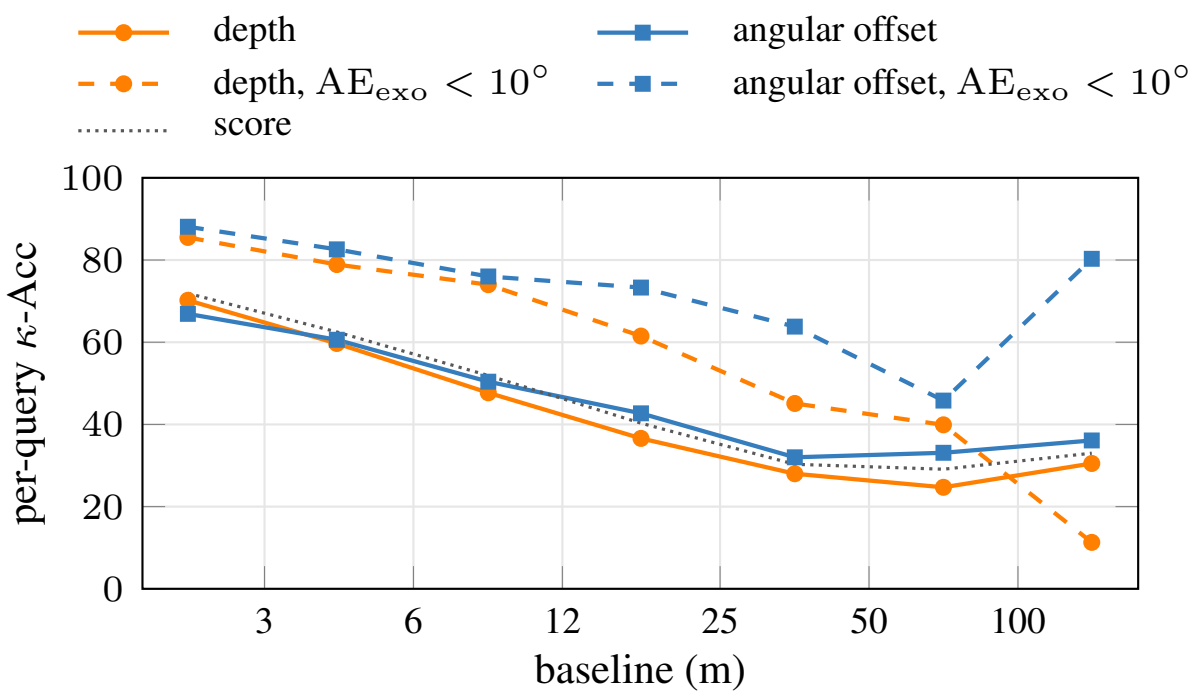
\begin{figure}[!htb]
\centering
\begin{tikzpicture}
\begin{axis}[width=0.98\columnwidth, height=4.3cm, xmode=log, xmin=1.7, xmax=175, ymin=0, ymax=100,
  xtick={3,6,12,25,50,100}, xticklabels={3,6,12,25,50,100}, minor xtick={}, ytick={0,20,40,60,80,100},
  xlabel={baseline (m)}, ylabel={per-query \ka}, tick label style={font=\scriptsize}, label style={font=\footnotesize},
  xlabel shift=-3pt, ylabel shift=-4pt, grid=major, grid style={gray!20},
  legend style={font=\scriptsize, draw=none, fill=none, at={(0.5,1.03)}, anchor=south, cells={anchor=west}, row sep=-1pt, column sep=6pt}, legend columns=2]
\addplot[thick, color=orange, mark=*, mark size=1.2pt] coordinates {(2.1,70.2) (4.2,59.7) (8.5,47.7) (17.3,36.6) (35.4,28.0) (70.7,24.7) (141,30.5)};
\addlegendentry{depth}
\addplot[thick, color=cvprblue, mark=square*, mark size=1.1pt] coordinates {(2.1,66.9) (4.2,60.6) (8.5,50.4) (17.3,42.7) (35.4,32.0) (70.7,33.1) (141,36.1)};
\addlegendentry{angular offset}
\addplot[thick, dashed, color=orange, mark=*, mark size=1.2pt, mark options={solid}] coordinates {(2.1,85.5) (4.2,78.9) (8.5,74.0) (17.3,61.5) (35.4,45.1) (70.7,39.9) (141,11.3)};
\addlegendentry{depth, $\aex<10^\circ$}
\addplot[thick, dashed, color=cvprblue, mark=square*, mark size=1.1pt, mark options={solid}] coordinates {(2.1,88.1) (4.2,82.6) (8.5,76.0) (17.3,73.3) (35.4,63.8) (70.7,45.8) (141,80.3)};
\addlegendentry{angular offset, $\aex<10^\circ$}
\addplot[semithick, color=black!65, densely dotted, mark=none] coordinates {(2.1,71.9) (4.2,62.5) (8.5,51.9) (17.3,40.3) (35.4,30.3) (70.7,29.1) (141,33.0)};
\addlegendentry{score}
\end{axis}
\end{tikzpicture}
\caption{\textbf{The depth readout loses accuracy faster with distance than the angular offset.} Results use zero-shot \vggto with both input orders, macro-averaged over domains with metric baselines (up to 13 per bin).}
\label{fig:cross}
\end{figure}

\begin{table*}[t]
\centering
\caption{\textbf{Gold versus Full tier by method} (Gold / Full, excluding Ego4D-TF). $^\ast$: angular offset only, as in \cref{tab:who}. As for the other 3D models, the exo-first \vggto row does not reflect ego cameras predicted behind the exo camera, unlike \cref{tab:components} (\cref{sec:supp_impl}).}
\label{tab:goldfull}
\footnotesize
\setlength{\tabcolsep}{4pt}
\begin{tabular}{lcccc}
\toprule
 & \tabhead{2}{\ka} & \tabhead{2}{MA@5} \\
\cmidrule(lr){2-3}\cmidrule(lr){4-5}
Method & Seen & Unseen & Seen & Unseen \\
\midrule
\multicolumn{5}{l}{\emph{Without training on our data, exo image first}} \\
DUSt3R & 26.9 / 26.5 & 24.7 / 25.0 & 9.5 / 9.3 & 8.5 / 8.1 \\
MASt3R & 40.5 / 39.6 & 33.5 / 33.8 & 16.1 / 15.9 & 14.3 / 14.2 \\
CUT3R & 31.4 / 31.2 & 24.1 / 24.6 & 12.2 / 11.6 & 8.8 / 8.9 \\
VGGT & 38.5 / 38.4 & 35.9 / 36.2 & 11.9 / 12.0 & 13.5 / 13.2 \\
Pi3X & 48.6 / 49.0 & 50.1 / 50.4 & 14.2 / 14.1 & 22.6 / 22.0 \\
MapAnything & 52.1 / 51.3 & 49.8 / 50.1 & 20.6 / 21.0 & 20.2 / 19.7 \\
Depth Anything~3 & 50.7 / 50.7 & 48.2 / 48.2 & 29.9 / 29.2 & 24.0 / 23.4 \\
Reloc3r$^\ast$ & 34.5 / 33.4 & 32.7 / 32.9 & 14.2 / 14.1 & 12.7 / 12.2 \\
RoMa + E$^\ast$ & 28.7 / 28.5 & 24.8 / 25.1 & 12.3 / 11.7 & 9.8 / 9.8 \\
MASt3R matches + E$^\ast$ & 41.7 / 41.4 & 36.9 / 37.4 & 24.1 / 23.9 & 22.5 / 21.9 \\
\vggto, dense tokens + PnP & 19.4 / 19.2 & 16.0 / 16.0 & 2.2 / 1.8 & 1.5 / 1.7 \\
\vggto, camera head & 61.0 / 60.2 & 58.2 / 58.4 & 28.9 / 28.6 & 28.8 / 27.9 \\
\midrule
\rowcolor{tabourslight}
\ours-ZS & 60.7 / 60.5 & 58.6 / 58.9 & 28.6 / 28.5 & 28.9 / 28.1 \\
\rowcolor{tabours}
\ours & 87.4 / 87.1 & 78.3 / 77.9 & 72.7 / 71.8 & 53.2 / 50.6 \\
\bottomrule
\end{tabular}
\end{table*}

\paragraph{Depth and angular offset by distance.}
Across camera baselines and mask sizes, we compare \emph{depth} (camera distance alone), \emph{offset} (angular offset $\theta_i$), and \emph{score} ($s_i$ from \cref{eq:mix}): mask depth helps for nearby, large wearers, whereas angular offset alone suits small, distant ones. In \cref{fig:cross}, baseline bins beyond 50\,m cover at most 4 domains and dashed points beyond 100\,m cover only one; \cref{tab:crossover} includes a setting in a bin when it has at least 30 queries. By mask size, depth leads above about 0.3\% of the image, while angular offset leads below it, by 9.5 points under 0.1\%. The same distance trend holds within ground domains spanning several distances (\cref{tab:crossdom}), and the zero-shot baselines in \cref{tab:who} show the same trade-off: read with angular offset alone, they lose 3.1--6.6 \ka on seen settings but gain 3.9--16.6 on birds and 3.4--13.3 on drones.

\paragraph{Queries with an accurate pose.}
On queries with an accurate pose ($\aex<10^\circ$, dashed lines), angular offset outperforms depth at every distance, leading by 2.6 points below 3\,m and 11.8 at 12--25\,m. Although this subset illustrates what angular offset can achieve rather than average performance, it shows that depth's close-range edge comes from poor-pose queries. Zero-shot depth thus hedges against pose error rather than being a stronger cue on its own, unlike the trained model's depth (see the true-epipole columns of \cref{tab:readtype}, which keep predicted intrinsics, depth, and baseline). Depth mainly separates candidates along the same ray.

\begin{figure*}[t]
\centering
\includegraphics[width=\textwidth]{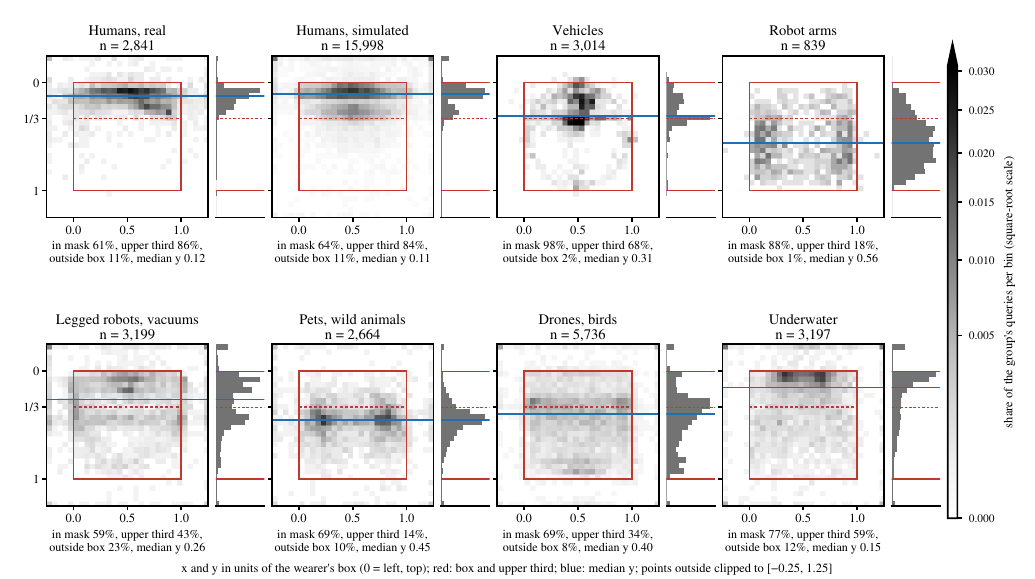}
\caption{\textbf{Where the camera sits on its wearer} (true epipoles of 37,488 queries).}
\label{fig:camloc}
\end{figure*}

\paragraph{Where the camera sits.}
Plotting true epipoles inside each wearer's bounding box (\cref{fig:camloc}) shows that no fixed position on the mask suits every wearer. On humans, cameras sit high (median 0.11--0.12 of the box height below its top), with 84--86\% of the epipoles in the box's upper third yet only 61--64\% inside the mask itself, whereas robot arms carry cameras lower (median 0.56).

\paragraph{Pose quality.}
\Cref{tab:pose0} splits queries by zero-shot pose error. Within the tolerance of \cref{sec:readout}, angular offset identifies the wearer on 96--100\% of queries in every domain with tolerance metadata except sports (94.4\%); the guarantee strictly assumes one input order and all mask pixels, whereas our readout fuses both input orders and samples pixels.

\begin{table*}[t]
\centering
\caption{\textbf{Depth and angular offset by distance and mask size.} Per-query \ka macro-averaged over domains for zero-shot \vggto using both input orders. Accurate pose: $\aex<10^\circ$.}
\label{tab:crossover}
\footnotesize
\setlength{\tabcolsep}{5pt}
\begin{tabular}{lrcccccc}
\toprule
 & & \tabhead{3}{All queries} & \tabhead{3}{Accurate pose} \\
\cmidrule(lr){3-5}\cmidrule(lr){6-8}
Baseline (m) & Queries & Depth & Offset & Score & Depth & Offset & Score \\
\midrule
$<$3 & 7,703 & 70.2 & 66.9 & 71.9 & 85.5 & 88.1 & 89.6 \\
3--6 & 5,844 & 59.7 & 60.6 & 62.5 & 78.9 & 82.6 & 83.8 \\
6--12 & 4,471 & 47.7 & 50.4 & 51.9 & 74.0 & 76.0 & 80.0 \\
12--25 & 5,571 & 36.6 & 42.7 & 40.3 & 61.5 & 73.3 & 69.0 \\
25--50 & 3,270 & 28.0 & 32.0 & 30.3 & 45.1 & 63.8 & 55.6 \\
50--100 & 2,370 & 24.7 & 33.1 & 29.1 & 39.9 & 45.8 & 43.2 \\
$\geq$100 & 839 & 30.5 & 36.1 & 33.0 & 11.3 & 80.3 & 35.5 \\
\midrule
Wearer area (\% of image) & & & & & & & \\
\midrule
$<$0.1 & 7,839 & 37.9 & 47.4 & 42.0 & 52.2 & 70.3 & 60.8 \\
0.1--0.3 & 5,830 & 46.3 & 47.7 & 49.2 & 70.4 & 68.9 & 74.4 \\
0.3--1 & 6,081 & 59.7 & 58.1 & 62.0 & 84.6 & 80.1 & 87.2 \\
1--3 & 4,390 & 63.8 & 63.4 & 67.4 & 84.6 & 89.3 & 91.0 \\
3--10 & 3,641 & 76.0 & 71.5 & 79.7 & 91.5 & 93.7 & 96.5 \\
$\geq$10 & 965 & 88.5 & 74.3 & 87.4 & 100.0 & 100.0 & 100.0 \\
\bottomrule
\end{tabular}
\end{table*}

\begin{table}[t]
\centering
\caption{\textbf{Lead of depth over angular offset within each domain} by baseline distance in meters (per-query \ka, zero-shot \vggto). Bins with fewer than 150 queries are omitted. Underwater pools all wearers across the underwater sets.}
\label{tab:crossdom}
\footnotesize
\setlength{\tabcolsep}{3pt}
\resizebox{\ifdim\width>\columnwidth\columnwidth\else\width\fi}{!}{%
\begin{tabular}{lcccccc}
\toprule
Domain & $<$3 & 3--6 & 6--12 & 12--25 & 25--50 & $\geq$50 \\
\midrule
Indoor & +16.4 & +11.4 & \no & \no & \no & \no \\
Real humans & +5.9 & +10.1 & +4.1 & \no & \no & \no \\
Pets & +3.4 & $-$0.1 & \no & \no & \no & \no \\
Nature & +17.3 & +21.2 & +8.7 & +1.3 & \no & \no \\
City & +13.9 & \no & +8.6 & +0.6 & \no & \no \\
Sports & \no & +17.3 & +9.0 & +0.9 & $-$0.3 & +1.3 \\
Vehicles & \no & \no & +0.4 & $-$1.1 & $-$8.7 & $-$6.8 \\
Thermal & +8.3 & $-$7.7 & \no & \no & \no & \no \\
Underwater & +3.1 & $-$6.7 & $-$16.0 & \no & \no & \no \\
Wild animals & +3.2 & +9.2 & +0.9 & $-$0.3 & $-$2.8 & $-$1.0 \\
Birds & $-$22.1 & $-$16.5 & $-$37.3 & $-$30.8 & $-$21.5 & \no \\
Drones & $-$9.9 & $-$33.1 & \no & \no & \no & $-$27.2 \\
Legged robots & +8.9 & +3.9 & $-$3.5 & \no & \no & \no \\
Robot arms & $-$4.0 & \no & \no & \no & \no & \no \\
Vacuums & +3.4 & $-$3.1 & \no & \no & \no & \no \\
\bottomrule
\end{tabular}}
\end{table}

\begin{table}[t]
\centering
\caption{\textbf{Identification accuracy by pose quality} (zero-shot \vggto with the combined score). Accurate: $\aex<10^\circ$. \ka is pooled per domain; Unreal lacks metadata for the tolerance, and Underwater pools all wearers across the underwater sets.}
\label{tab:pose0}
\footnotesize
\setlength{\tabcolsep}{3pt}
\resizebox{\ifdim\width>\columnwidth\columnwidth\else\width\fi}{!}{%
\begin{tabular}{lccccc}
\toprule
 & Accurate & \tabhead{2}{\ka} & In tolerance & Offset right \\
\cmidrule(lr){3-4}
Domain & (\%) & accurate & otherwise & (\%) & in tolerance (\%) \\
\midrule
Indoor & 83.6 & 94.6 & 71.5 & 51.4 & 99.3 \\
Real humans & 81.4 & 96.0 & 82.0 & 56.3 & 99.5 \\
Pets & 88.4 & 99.2 & 72.4 & 73.7 & 99.9 \\
Nature & 77.1 & 79.9 & 52.4 & 20.6 & 96.1 \\
City & 72.8 & 80.0 & 35.3 & 31.3 & 96.6 \\
Unreal & 71.6 & 83.7 & 37.6 & \no & \no \\
Sports & 66.5 & 77.9 & 24.0 & 25.0 & 94.4 \\
Vehicles & 53.8 & 81.1 & 14.5 & 22.9 & 98.3 \\
Thermal & 30.4 & 91.9 & 1.7 & 17.9 & 100.0 \\
Underwater & 10.9 & 67.4 & 6.7 & 6.4 & 98.1 \\
Wild animals & 46.3 & 53.7 & 6.5 & 11.3 & 96.1 \\
Birds & 45.0 & 45.7 & 10.0 & 27.6 & 97.9 \\
Drones & 30.2 & 50.7 & 14.3 & 24.3 & 98.5 \\
Legged robots & 74.6 & 92.5 & 39.4 & 55.2 & 99.2 \\
Robot arms & 55.4 & 89.2 & $-$5.9 & 43.1 & 99.4 \\
Vacuums & 79.6 & 97.2 & 49.8 & 68.2 & 99.3 \\
\midrule
Macro & 60.5 & 80.0 & 32.0 & 35.7 & \no \\
\bottomrule
\end{tabular}}
\end{table}

\subsection{The nearest mask at the true epipole}
\label{sec:supp_pose}

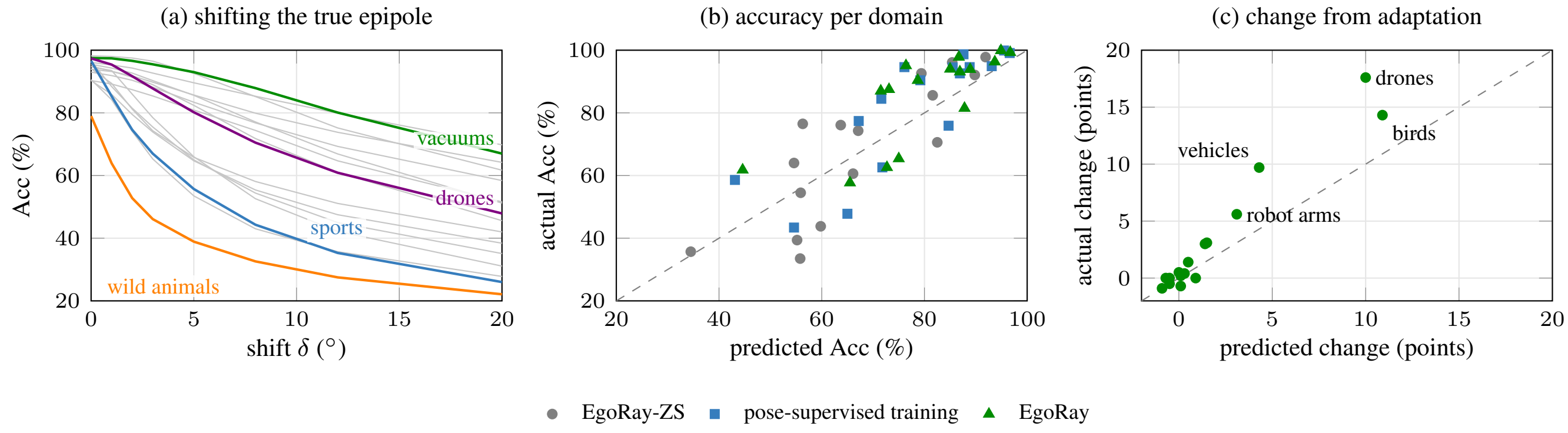
\begin{figure*}[t]
\centering
\begin{tikzpicture}
\begin{axis}[width=0.355\textwidth, height=4.4cm, tick label style={font=\scriptsize}, label style={font=\footnotesize}, title style={font=\footnotesize, yshift=-3pt}, grid=major, grid style={gray!20}, xlabel shift=-3pt, ylabel shift=-4pt, title={(a) shifting the true epipole}, xmin=0, xmax=20, ymin=20, ymax=100, xtick={0,5,10,15,20},
  xlabel={shift $\delta$ ($^\circ$)}, ylabel={Acc (\%)}]
\addplot[gray!45, thin, forget plot] coordinates {(0,96.1) (1,95.3) (2,94.3) (3,92.7) (5,89.6) (8,85.3) (12,79.9) (20,69.8)};
\addplot[gray!45, thin, forget plot] coordinates {(0,96.6) (1,88.4) (2,81.5) (3,75.5) (5,65.0) (8,54.4) (12,45.0) (20,35.1)};
\addplot[gray!45, thin, forget plot] coordinates {(0,94.1) (1,89.4) (2,81.2) (3,74.3) (5,65.8) (8,58.1) (12,51.1) (20,42.0)};
\addplot[gray!45, thin, forget plot] coordinates {(0,90.3) (1,89.2) (2,87.4) (3,85.4) (5,80.9) (8,74.4) (12,64.7) (20,51.6)};
\addplot[gray!45, thin, forget plot] coordinates {(0,94.9) (1,93.3) (2,91.4) (3,89.0) (5,82.4) (8,71.8) (12,61.0) (20,45.6)};
\addplot[gray!45, thin, forget plot] coordinates {(0,90.6) (1,84.2) (2,74.1) (3,65.3) (5,53.5) (8,43.0) (12,35.7) (20,27.8)};
\addplot[gray!45, thin, forget plot] coordinates {(0,98.2) (1,98.0) (2,97.5) (3,96.5) (5,92.6) (8,85.2) (12,75.2) (20,61.7)};
\addplot[gray!45, thin, forget plot] coordinates {(0,93.9) (1,93.2) (2,91.8) (3,89.8) (5,85.7) (8,79.9) (12,73.8) (20,64.2)};
\addplot[gray!45, thin, forget plot] coordinates {(0,93.0) (1,91.9) (2,90.4) (3,88.2) (5,83.4) (8,76.8) (12,69.3) (20,58.4)};
\addplot[gray!45, thin, forget plot] coordinates {(0,95.5) (1,94.3) (2,92.7) (3,90.3) (5,85.6) (8,77.1) (12,66.9) (20,51.2)};
\addplot[gray!45, thin, forget plot] coordinates {(0,90.3) (1,85.9) (2,79.5) (3,73.9) (5,64.7) (8,55.4) (12,47.5) (20,38.4)};
\addplot[gray!45, thin, forget plot] coordinates {(0,94.7) (1,93.5) (2,86.0) (3,78.5) (5,66.0) (8,52.6) (12,42.2) (20,31.1)};
\addplot[green!55!black, thick, forget plot] coordinates {(0,97.5) (1,97.4) (2,96.6) (3,95.5) (5,93.0) (8,87.9) (12,80.1) (20,67.0)};
\addplot[violet, thick, forget plot] coordinates {(0,97.4) (1,95.4) (2,91.7) (3,87.8) (5,80.2) (8,70.5) (12,60.9) (20,47.9)};
\addplot[cvprblue, thick, forget plot] coordinates {(0,96.6) (1,85.3) (2,74.6) (3,67.0) (5,55.7) (8,44.3) (12,35.3) (20,26.0)};
\addplot[orange, thick, forget plot] coordinates {(0,79.0) (1,63.9) (2,52.8) (3,46.1) (5,38.9) (8,32.6) (12,27.5) (20,22.1)};
\node[font=\scriptsize, text=green!55!black, anchor=south east, inner sep=1pt, fill=white, fill opacity=0.8, text opacity=1] at (axis cs:19.8,68.5) {vacuums};
\node[font=\scriptsize, text=violet, anchor=south east, inner sep=1pt, fill=white, fill opacity=0.8, text opacity=1] at (axis cs:19.8,49.4) {drones};
\node[font=\scriptsize, text=cvprblue, anchor=south west, inner sep=1pt, fill=white, fill opacity=0.8, text opacity=1] at (axis cs:10.5,38.5) {sports};
\node[font=\scriptsize, text=orange, anchor=south west, inner sep=1pt, fill=white, fill opacity=0.8, text opacity=1] at (axis cs:0.6,21.0) {wild animals};
\end{axis}
\begin{axis}[width=0.355\textwidth, height=4.4cm, tick label style={font=\scriptsize}, label style={font=\footnotesize}, title style={font=\footnotesize, yshift=-3pt}, grid=major, grid style={gray!20}, xlabel shift=-3pt, ylabel shift=-4pt, xshift=0.338\textwidth, title={(b) accuracy per domain}, xmin=20, xmax=100, ymin=20, ymax=100, xtick={20,40,60,80,100}, ytick={20,40,60,80,100},
  xlabel={predicted Acc (\%)}, ylabel={actual Acc (\%)},
  legend style={font=\scriptsize, draw=none, fill=none, at={(0.5,-0.36)}, anchor=north, cells={anchor=west}, column sep=6pt}, legend columns=3]
\addplot[only marks, mark=*, mark size=1.5pt, gray] coordinates {(82.5,70.6) (55.2,39.4) (67.1,74.3) (59.8,43.8) (79.4,92.6) (66.1,60.6) (56.3,76.5) (91.9,97.8) (85.4,96.1) (81.6,85.6) (54.6,64.0) (55.8,33.5) (63.7,76.1) (89.8,92.1) (55.9,54.5) (34.5,35.7)};
\addlegendentry{\ours-ZS}
\addplot[only marks, mark=square*, mark size=1.4pt, cvprblue] coordinates {(84.7,75.9) (54.6,43.4) (85.5,94.5) (65.0,47.8) (87.6,98.7) (88.8,94.6) (76.1,94.6) (96.6,99.1) (95.5,99.9) (86.9,92.6) (71.6,84.5) (71.8,62.6) (79.2,90.4) (93.1,94.9) (67.2,77.4) (43.1,58.6)};
\addlegendentry{pose-supervised training}
\addplot[only marks, mark=triangle*, mark size=1.9pt, green!55!black] coordinates {(87.8,81.5) (65.5,57.7) (85.0,94.0) (75.0,65.4) (86.7,97.8) (88.9,93.9) (76.4,95.1) (96.7,99.3) (94.9,99.9) (86.9,93.1) (73.1,87.5) (72.7,62.6) (78.7,90.4) (93.7,96.3) (71.5,87.0) (44.6,61.8)};
\addlegendentry{\ours}
\addplot[gray, dashed, domain=20:100, samples=2, forget plot] {x};
\end{axis}
\begin{axis}[width=0.355\textwidth, height=4.4cm, tick label style={font=\scriptsize}, label style={font=\footnotesize}, title style={font=\footnotesize, yshift=-3pt}, grid=major, grid style={gray!20}, xlabel shift=-3pt, ylabel shift=-4pt, xshift=0.676\textwidth, title={(c) change from adaptation}, xmin=-2, xmax=20, ymin=-2, ymax=20,
  xlabel={predicted change (points)}, ylabel={actual change (points)}, clip=false]
\addplot[only marks, mark=*, mark size=1.5pt, green!55!black] coordinates {(3.1,5.6) (10.9,14.3) (-0.5,-0.5) (10.0,17.6) (-0.9,-0.9) (0.1,-0.7) (0.3,0.4) (0.1,0.2) (-0.7,0.0) (-0.0,0.5) (1.4,3.0) (0.9,0.0) (-0.5,0.0) (0.5,1.4) (4.3,9.7) (1.5,3.1)};
\addplot[gray, dashed, domain=-2:20, samples=2] {x};
\node[font=\scriptsize, anchor=west, inner sep=3pt] at (axis cs:10.0,17.6) {drones};
\node[font=\scriptsize, anchor=north west, inner sep=3pt] at (axis cs:10.9,14.3) {birds};
\node[font=\scriptsize, anchor=south east, inner sep=3pt] at (axis cs:4.3,9.7) {vehicles};
\node[font=\scriptsize, anchor=west, inner sep=3pt] at (axis cs:3.1,5.6) {robot arms};
\end{axis}
\end{tikzpicture}
\caption{\textbf{Pose error predicts identification accuracy} (Acc). (a) Accuracy of selecting the mask nearest to the true epipole after shifting the epipole by $\delta$, one curve per domain. (b) One point per model and domain. Predicted accuracy ($x$): each test set's curve, computed as in (a), read at the model's $\aex$ on each query. Actual accuracy ($y$): the model's accuracy on the same queries. Both are averaged over a set's queries and then over the domain's sets. (c) One point per domain: predicted ($x$) versus actual ($y$) change in accuracy from test-time adaptation (pose-supervised training to \ours). Dashed lines mark equality. Panels (b, c) include only queries with camera poses, so real humans and vehicles differ slightly from \cref{tab:pose}.}
\label{fig:dose}
\end{figure*}

To isolate how pose error limits identification, \cref{tab:pose} reports the accuracy of picking the mask nearest to the true epipole in every domain with camera poses, and \cref{fig:dose} traces how it drops as the true epipole is rotated by $\delta$ in eight random directions. Across all models and domains, the accuracies these curves predict at each model's pose errors correlate with the actual ones ($r=0.84$ over 48 points). Human, vehicle, and wild-animal domains sit above the diagonal, consistent with our score using mask depth, whereas underwater wearers, drones, robot arms, and birds sit below it, where erroneous epipoles tend to lean toward another candidate. Accuracy changes from test-time adaptation track these curve predictions ($r=0.97$; Spearman's $\rho=0.90$, since both actual and predicted changes are near zero outside four domains), and the actual gains are 1.5 times the predicted ones, against 1.2 times for pose-supervised training ($r=0.81$). This true-epipole oracle is not an upper bound on our readout: where humans occlude one another (indoor, nature, and real humans), our combined score outperforms it (e.g., 97.8 versus 90.3 indoors).

\begin{table}[!htb]
\centering
\caption{\textbf{Pose estimation is the main bottleneck} (Acc). Error: median $\aex$ on queries where \ours errs. Gap: median angle from the true epipole to the nearest other candidate. Top views lack camera poses, and Underwater pools all wearers across the underwater sets.}
\label{tab:pose}
\footnotesize
\setlength{\tabcolsep}{4pt}
\begin{tabular}{lcccc}
\toprule
Domain & \ours & True epipole & Error & Gap \\
\midrule
Birds & 57.7 & 96.6 & 7.8$^\circ$ & 5.6$^\circ$ \\
Wild animals & 61.7 & 79.0 & 6.0$^\circ$ & 0.9$^\circ$ \\
Underwater & 62.6 & 95.5 & 22.4$^\circ$ & 11.5$^\circ$ \\
Drones & 65.4 & 97.4 & 14.4$^\circ$ & 11.4$^\circ$ \\
Robot arms & 81.6 & 96.1 & 19.1$^\circ$ & 17.3$^\circ$ \\
Sports & 87.5 & 96.6 & 4.9$^\circ$ & 3.2$^\circ$ \\
Vehicles & 87.5 & 94.7 & 9.3$^\circ$ & 3.1$^\circ$ \\
Unreal & 90.4 & 90.3 & 5.0$^\circ$ & \no \\
Legged robots & 93.1 & 93.0 & 20.5$^\circ$ & 11.9$^\circ$ \\
Thermal & 93.9 & 94.9 & 23.2$^\circ$ & 7.6$^\circ$ \\
City & 94.0 & 94.1 & 2.0$^\circ$ & 3.4$^\circ$ \\
Nature & 95.1 & 90.6 & 3.6$^\circ$ & 2.6$^\circ$ \\
Vacuums & 96.3 & 97.5 & 45.5$^\circ$ & 17.9$^\circ$ \\
Indoor & 97.8 & 90.3 & 2.6$^\circ$ & 10.1$^\circ$ \\
Real humans & 98.9 & 93.9 & 2.9$^\circ$ & 13.3$^\circ$ \\
Pets & 99.3 & 98.2 & 13.3$^\circ$ & 13.8$^\circ$ \\
Top view & 99.4 & \no & \no & \no \\
\bottomrule
\end{tabular}
\end{table}

\subsection{Shared environment and angular gap}
\label{sec:supp_env}

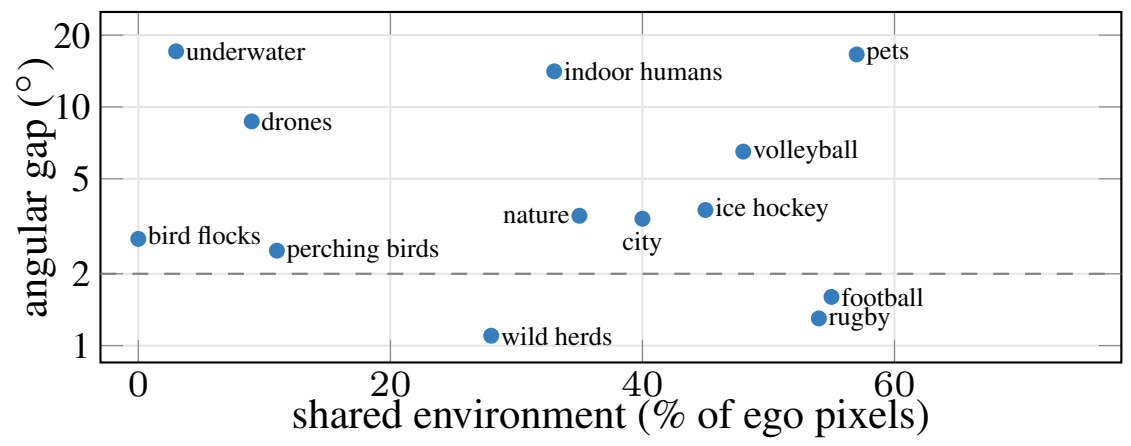
\begin{figure}[!htb]
\centering
\begin{tikzpicture}
\begin{axis}[width=\columnwidth, height=3.9cm, ymode=log, xmin=-3, xmax=78, ymin=0.85, ymax=25,
  xlabel={shared environment (\% of ego pixels)}, ylabel={angular gap ($^\circ$)},
  ytick={1,2,5,10,20}, yticklabels={1,2,5,10,20}, xtick={0,20,40,60}, tick label style={font=\scriptsize}, label style={font=\footnotesize},
  xlabel shift=-3pt, ylabel shift=-4pt, grid=major, grid style={gray!20}]
\addplot[only marks, mark=*, mark size=1.3pt, color=cvprblue] coordinates {
 (33,14.1) (57,16.6) (35,3.5) (40,3.4) (48,6.5) (45,3.7) (55,1.6) (54,1.3) (28,1.1) (11,2.5) (9,8.7) (0,2.8) (3,17.1)};
\addplot[dashed, gray, forget plot] coordinates {(-3,2) (78,2)};
\tiny
\node[anchor=west] at (axis cs:33,14.1) {indoor humans};
\node[anchor=west] at (axis cs:57,16.6) {pets};

\node[anchor=east] at (axis cs:35,3.5) {nature};
\node[anchor=north] at (axis cs:40,3.3) {city};
\node[anchor=west] at (axis cs:48,6.5) {volleyball};
\node[anchor=west] at (axis cs:45,3.7) {ice hockey};
\node[anchor=west] at (axis cs:55,1.6) {football};
\node[anchor=west] at (axis cs:54,1.3) {rugby};
\node[anchor=west] at (axis cs:28,1.1) {wild herds};
\node[anchor=west] at (axis cs:11,2.5) {perching birds};
\node[anchor=west] at (axis cs:9,8.7) {drones};
\node[anchor=west] at (axis cs:0,2.9) {bird flocks};
\node[anchor=west] at (axis cs:3,17.1) {underwater};
\end{axis}
\end{tikzpicture}
\caption{\textbf{Two sources of difficulty.} Each point is a simulated v1 set. Below the dashed line, the median angular gap is under $2^\circ$.}
\label{fig:env}
\end{figure}

We measure difficulty along two axes (\cref{tab:env}) to characterize what an ego--exo pair offers a geometric model. First, \emph{shared environment} is the share of ego pixels showing parts of the environment also visible in the exo view, computed by reprojecting depth up to the visibility range. Second, \emph{angular gap} is $\gamma_\mathrm{other}$ (\cref{sec:readout})---the angle from the true epipole to the nearest other candidate---which indicates how accurate the ego-camera direction must be. The two axes vary separately across sets (\cref{fig:env}): shared environment can coexist with small angular gaps, whereas underwater scenes have large angular gaps with almost no shared environment (near zero in 40\% of pairs).

\begin{table*}[!tp]
\centering
\caption{\textbf{Environment statistics per simulated set} (v1, shared environment and corners from 30 pairs per set). Facing: axis angle above $135^\circ$. $|\Delta h|$: camera height difference. Shared${}\approx0$: share of pairs with shared environment below 1\%. Corners: median number of Shi--Tomasi corners in the ego image (at most 3{,}000) that fall on the shared environment. Candidates, Gap, Distance, $|\Delta h|$, and Shared: medians; $<2^\circ$, $<5^\circ$, and Facing: shares of pairs.}
\label{tab:env}
\footnotesize
\setlength{\tabcolsep}{4pt}
\begin{tabular}{lcccccccccc}
\toprule
Set & Candidates & Gap ($^\circ$) & $<2^\circ$ & $<5^\circ$ & Distance (m) & Facing & $|\Delta h|$ (m) & Shared & Shared${}\approx0$ & Corners \\
\midrule
Indoor humans & 4 & 14.1 & 2\% & 12\% & 3.3 & 9\% & 0.7 & 33\% & 0\% & 346 \\
Pets & 3 & 16.6 & 0\% & 6\% & 3.1 & 7\% & 2.0 & 57\% & 0\% & 464 \\

Nature & 9 & 3.5 & 21\% & 67\% & 14 & 8\% & 4.3 & 35\% & 3\% & 668 \\
City & 6 & 3.4 & 22\% & 68\% & 14 & 16\% & 1.8 & 40\% & 0\% & 1{,}220 \\
Beach volleyball & 7 & 6.5 & 10\% & 38\% & 20 & 20\% & 6.5 & 48\% & 0\% & 1{,}180 \\
Ice hockey & 11 & 3.7 & 25\% & 61\% & 24 & 24\% & 1.5 & 45\% & 3\% & 241 \\
Football & 18 & 1.6 & 58\% & 87\% & 66 & 18\% & 16.5 & 55\% & 3\% & 1{,}240 \\
Rugby & 26 & 1.3 & 66\% & 92\% & 73 & 18\% & 15.9 & 54\% & 0\% & 1{,}180 \\
Wild herds & 21 & 1.1 & 77\% & 97\% & 48 & 13\% & 1.9 & 28\% & 0\% & 1{,}170 \\
Perching birds & 7 & 2.5 & 42\% & 75\% & 7.6 & 21\% & 1.5 & 11\% & 17\% & 102 \\
Drones, air operations & 5 & 8.7 & 10\% & 27\% & 175 & 16\% & 25 & 9\% & 43\% & 64 \\
Bird flocks & 14 & 2.8 & 29\% & 78\% & 34 & 6\% & 17 & 0\% & 60\% & 0 \\
Underwater & 4 & 17.1 & 1\% & 8\% & 4.7 & 16\% & 1.2 & 3\% & 40\% & 6 \\
\bottomrule
\end{tabular}
\end{table*}

\subsection{Segmentation by ranking proposals}
\label{sec:supp_seg}

Holding the predicted epipole fixed, we compare epipole-prompted segmentation against ranking SAM~3 proposals; ranking proposals yields the highest mean IoU (\cref{tab:heads}).

\begin{table}[!htbp]
\centering
\caption{\textbf{Segmentation methods} given the same localization (zero-shot \vggto, nine v1 sets, 500 present queries each). Point prompts: the predicted epipole.}
\label{tab:heads}
\footnotesize
\setlength{\tabcolsep}{4pt}
\begin{tabular}{lc}
\toprule
Method & mIoU \\
\midrule
SAM~3 point prompt & 19.9 \\
SAM~2.1~\cite{ravi2025sam2} point prompt & 22.6 \\
DINOv3~\cite{simeoni2025dinov3} region growing & 7.4 \\
DINOv2~\cite{oquab2024dinov2} region growing & 8.7 \\
Grounding DINO~\cite{liu2024groundingdino} box + SAM~2.1 & 51.2 \\
SAM~3 proposals + angular offset & 55.0 \\
SAM~3 proposals + score & 55.8 \\
\midrule
Oracle proposal & 78.6 \\
\bottomrule
\end{tabular}
\end{table}

\subsection{Readouts by group}
\label{sec:supp_readgroup}

\Cref{tab:full} compares the angular offset, depth, and combined score readouts group by group on the Full tier. Based on the development list, where alternative readouts did not justify a change, we adopt the score with $\alpha=3$.

\begin{table*}[!tp]
\centering
\caption{\textbf{Zero-shot \vggto by group} (Acc, Full tier). Depth and Offset: one input order. Two orders: angular offset with both. $\aex$, $\aeg$: median errors in degrees. Const.\ $\aex$: for a constant direction.}
\label{tab:full}
\scriptsize
\setlength{\tabcolsep}{4pt}
\begin{tabular}{lcccccccccc}
\toprule
Group & Random & Largest & Depth & Offset & Two orders & Score & One-to-one & $\aex\downarrow$ & $\aeg\downarrow$ & Const.\ $\aex\downarrow$ \\
\midrule
TF2023 & 25.4 & 30.7 & 95.5 & 88.7 & 88.3 & 94.8 & 95.3 & \no & \no & \no \\
Ego-Exo4D, seen sites & 40.0 & 80.4 & 96.9 & 90.9 & 91.4 & 97.0 & 97.0 & 6.7 & 8.5 & 13.5 \\
Ego-Exo4D, held-out sites & 44.7 & 99.7 & 99.7 & 96.5 & 96.4 & 99.6 & 99.6 & 5.2 & 6.1 & 12.5 \\
Harmony4D & 50.0 & 50.0 & 90.2 & 89.9 & 89.9 & 91.1 & 94.5 & 1.4 & 1.9 & 10.2 \\
EgoHumans & 29.4 & 33.3 & 95.5 & 85.0 & 85.4 & 94.8 & 97.8 & 5.5 & 6.0 & 17.8 \\
CoMind & 50.0 & 50.0 & 91.5 & 94.0 & 94.2 & 95.1 & 98.6 & \no & 15.1 & \no \\
CORE4D-Real & 50.0 & 51.7 & 97.7 & 93.9 & 93.8 & 97.1 & 97.1 & \no & 10.0 & \no \\
Ego4D-TF (peer) & 27.8 & 27.8 & 90.9 & 86.3 & 86.4 & 90.3 & 92.7 & \no & \no & \no \\
Top view & 8.5 & 15.3 & 82.0 & 82.8 & 85.1 & 84.8 & 85.5 & \no & \no & \no \\
\midrule
Indoor, fixed & 20.0 & 20.8 & 95.5 & 87.3 & 87.2 & 94.9 & 98.0 & 3.0 & 4.1 & 23.4 \\
Indoor, peer & 27.2 & 28.1 & 90.5 & 79.5 & 79.3 & 89.8 & 92.5 & 6.9 & 6.8 & 48.1 \\
Nature, fixed & 10.3 & 10.5 & 74.8 & 70.0 & 70.8 & 77.6 & 83.7 & 3.3 & 6.8 & 13.8 \\
Nature, peer & 16.7 & 16.9 & 74.1 & 61.9 & 61.9 & 75.4 & 77.1 & 8.2 & 7.8 & 39.8 \\
City, fixed & 15.5 & 15.7 & 71.9 & 70.9 & 71.8 & 75.1 & 83.5 & 3.5 & 5.8 & 21.4 \\
City, peer & 25.6 & 28.0 & 71.5 & 70.5 & 71.1 & 74.8 & 81.1 & 8.5 & 8.7 & 40.4 \\
Sports, fixed & 9.9 & 9.2 & 60.9 & 55.7 & 56.1 & 62.1 & 64.5 & 11.0 & 12.5 & 16.0 \\
Sports, peer & 14.7 & 16.0 & 63.5 & 60.7 & 61.0 & 67.3 & 69.3 & 7.0 & 7.1 & 47.9 \\
Unreal, fixed & 15.2 & 11.2 & 77.5 & 70.6 & 71.0 & 77.7 & 81.9 & 3.6 & 6.0 & 18.2 \\
Unreal, peer & 25.5 & 22.7 & 74.9 & 67.1 & 67.2 & 74.8 & 78.9 & 10.7 & 10.4 & 52.1 \\
Thermal to thermal & 20.6 & 21.0 & 59.2 & 59.9 & 60.3 & 60.8 & 68.1 & 5.9 & 7.3 & 21.6 \\
RGB to thermal & 20.6 & 20.8 & 25.7 & 23.8 & 23.5 & 26.6 & 29.0 & 129.1 & 82.3 & 21.7 \\
RGB to RGB (control) & 20.1 & 20.7 & 95.4 & 83.3 & 83.9 & 94.5 & 98.6 & 4.3 & 5.0 & 21.8 \\
\midrule
Divers and ROVs, fixed & 20.2 & 20.6 & 26.9 & 31.6 & 31.6 & 28.9 & 33.6 & 79.8 & 76.7 & 24.3 \\
Divers and ROVs, peer & 26.9 & 27.5 & 32.4 & 37.1 & 36.7 & 34.5 & 40.3 & 70.4 & 71.6 & 56.9 \\
Fish, fixed & 18.6 & 16.0 & 21.6 & 25.8 & 26.0 & 23.5 & 27.7 & 82.5 & 87.2 & 16.8 \\
Fish, peer & 30.4 & 28.3 & 40.1 & 51.7 & 51.5 & 46.8 & 58.0 & 53.5 & 53.0 & 60.9 \\
\midrule
Pets, fixed & 36.7 & 37.0 & 98.1 & 97.8 & 98.4 & 98.4 & 100.0 & 2.7 & 3.5 & 14.6 \\
Pets, peer & 46.2 & 46.3 & 95.8 & 92.1 & 90.9 & 96.5 & 98.8 & 5.8 & 5.8 & 59.8 \\
Wild animals, fixed & 6.6 & 6.7 & 19.0 & 19.2 & 19.0 & 20.2 & 21.3 & 31.1 & 34.1 & 9.1 \\
Wild animals, peer & 15.8 & 15.9 & 48.2 & 47.4 & 47.1 & 51.1 & 53.9 & 9.1 & 9.6 & 38.1 \\

Birds, fixed & 13.9 & 14.3 & 20.0 & 32.5 & 33.1 & 26.1 & 28.7 & 40.0 & 44.7 & 11.9 \\
Birds, peer & 26.1 & 25.5 & 40.5 & 71.7 & 71.9 & 53.8 & 60.0 & 9.6 & 9.6 & 73.4 \\
Drones, fixed & 19.6 & 19.5 & 25.6 & 36.3 & 38.0 & 32.1 & 36.8 & 30.8 & 35.5 & 11.3 \\
Drones, peer & 31.3 & 31.7 & 48.3 & 70.1 & 69.9 & 55.5 & 63.5 & 14.0 & 14.0 & 70.4 \\
Humanoids and quadrupeds, fixed & 27.7 & 26.6 & 86.4 & 84.9 & 84.9 & 87.9 & 92.8 & 2.7 & 3.7 & 21.6 \\
Humanoids and quadrupeds, peer & 35.2 & 36.4 & 85.1 & 81.0 & 81.2 & 86.5 & 91.2 & 6.8 & 6.9 & 42.6 \\
Robot arms & 44.8 & 45.0 & 67.9 & 68.8 & 69.6 & 70.8 & 77.9 & 10.8 & 23.2 & 15.1 \\
Robot vacuums, fixed & 28.1 & 28.1 & 92.9 & 93.8 & 93.6 & 93.4 & 99.0 & 2.9 & 3.8 & 23.2 \\
Robot vacuums, peer & 41.7 & 41.7 & 90.4 & 88.0 & 87.7 & 91.9 & 98.1 & 9.2 & 9.2 & 82.3 \\
Vehicles, roadside & 13.9 & 20.4 & 46.7 & 47.1 & 50.2 & 52.0 & 52.6 & 43.4 & 46.0 & 16.7 \\
Vehicle to vehicle & 20.0 & 27.0 & 61.2 & 69.4 & 69.8 & 67.2 & 67.2 & 2.8 & 2.5 & 21.2 \\
DROID (real arm) & 46.8 & 99.0 & 98.0 & 96.0 & 98.0 & 100.0 & 100.0 & 3.4 & 5.5 & 20.4 \\
\midrule
Macro, seen & 24.0 & 27.9 & 67.7 & 64.7 & 64.8 & 69.1 & 73.4 & \no & \no & \no \\
Macro, unseen & 25.1 & 25.9 & 64.7 & 66.7 & 67.2 & 67.7 & 71.5 & \no & \no & \no \\
\bottomrule
\end{tabular}
\end{table*}

\subsection{Readout variants and shortcut controls}
\label{sec:supp_comp}

\begin{table*}[!tp]
\centering
\caption{\textbf{Readouts by wearer type} (\ka). Bold: best of the three readouts.}
\label{tab:readtype}
\scriptsize
\setlength{\tabcolsep}{2.2pt}
\resizebox{\ifdim\width>\textwidth\textwidth\else\width\fi}{!}{%
\begin{tabular}{lcccccccccccc}
\toprule
 & \tabhead{6}{\ours} & \tabhead{6}{\ours-ZS} \\
\cmidrule(lr){2-7}\cmidrule(lr){8-13}
 & \tabhead{3}{Predicted epipole} & \tabhead{3}{True epipole} & \tabhead{3}{Predicted epipole} & \tabhead{3}{True epipole} \\
\cmidrule(lr){2-4}\cmidrule(lr){5-7}\cmidrule(lr){8-10}\cmidrule(lr){11-13}
Wearer & Offset & Upper third & Mask minimum & Offset & Upper third & Mask minimum & Offset & Upper third & Mask minimum & Offset & Upper third & Mask minimum \\
\midrule
Humans & 88.6 & \textbf{93.6} & 93.1 & 84.5 & \textbf{94.9} & 94.7 & 64.1 & \textbf{71.7} & 71.1 & 75.2 & \textbf{77.6} & 75.9 \\
Vehicles & 83.7 & \textbf{86.2} & 85.5 & 71.0 & 77.9 & \textbf{79.8} & 49.7 & \textbf{50.2} & \textbf{50.2} & \textbf{71.4} & 65.5 & 64.2 \\
Robot arms & 63.9 & 60.1 & \textbf{66.9} & 89.6 & 83.3 & \textbf{91.6} & 44.8 & 40.9 & \textbf{47.0} & \textbf{92.5} & 77.7 & 80.8 \\
Drones and birds & 52.6 & \textbf{52.7} & 51.7 & \textbf{81.1} & 79.8 & 80.3 & \textbf{40.1} & 26.9 & 25.5 & \textbf{74.2} & 45.0 & 43.9 \\
Animals & 72.8 & 77.2 & \textbf{77.8} & 66.7 & 75.1 & \textbf{76.7} & 57.8 & \textbf{61.9} & 61.8 & 66.9 & \textbf{67.4} & 67.2 \\
Underwater & 49.5 & 50.6 & \textbf{51.1} & 83.1 & 85.5 & \textbf{85.8} & \textbf{16.6} & 13.8 & 13.0 & \textbf{77.9} & 34.9 & 30.0 \\
Robots and vacuums & 89.1 & 91.6 & \textbf{92.0} & 90.2 & 95.8 & \textbf{96.8} & 79.2 & 83.2 & \textbf{83.3} & 87.4 & \textbf{88.3} & \textbf{88.3} \\
\midrule
Macro, seen & 82.9 & 87.3 & \textbf{87.4} & 85.8 & 93.6 & \textbf{94.2} & 54.8 & \textbf{61.2} & 60.7 & \textbf{78.8} & 71.3 & 69.5 \\
Macro, unseen & 75.6 & 78.0 & \textbf{78.2} & 81.0 & 86.3 & \textbf{87.3} & 57.8 & \textbf{58.6} & \textbf{58.6} & \textbf{76.2} & 68.7 & 67.8 \\
\bottomrule
\end{tabular}}
\end{table*}

\Cref{tab:readtype} compares, across wearer types, the angular offset, the score at the centroid of the mask's upper third, and the score minimized over the mask (\cref{eq:mix}). Minimizing over the mask gives a small macro gain for \ours and helps wrist-mounted cameras, whereas zero-shot depth remains unreliable even with true epipoles.

Because distinct ego images of one exo frame come from different wearers, querying several of them allows a one-to-one Hungarian assignment~\cite{kuhn1955hungarian}, using cost $s_{qi}$ divided by the median of $s_{qj}$ over query $q$'s candidates $j$. We report this assignment separately (\cref{tab:comp,tab:full}) because it depends on which ego images of a frame are queried; it also holds only within a mount type, since one human can wear head and chest cameras simultaneously.

\Cref{tab:comp} compares readout components with alternatives, and \cref{tab:short} tests for shortcuts. Swapping the ego image for one from another scene drops accuracy to chance, confirming that selection relies on the cross-image relation. Inpainting all candidates out of the exo image keeps the angular-offset accuracy within 5 points of the original in the indoor, bird, and robot-arm domains, showing that most identification accuracy survives without the candidates' appearance; in nature scenes, however, it drops by 12--17 points.

\begin{table}[!htbp]
\centering
\caption{\textbf{Readout components and alternatives} on \vggto. Upper block: Acc on the Full tier, macro-averaged over all settings. Lower block: each alternative \vs the rule it replaces; dev.: development list. Rank-16 model: LoRA rank 16 with a shorter schedule (\cref{tab:train}). Appearance rows cover the 16,857 test queries with features.}
\label{tab:comp}
\footnotesize
\setlength{\tabcolsep}{3pt}
\begin{tabular}{@{}>{\raggedright\arraybackslash\hangindent=1em\hangafter=1}p{0.52\columnwidth}l@{}}
\toprule
Variant & Result \\
\midrule
Angular offset, one input order & 66.4 \\
\quad + two input orders & 66.9 (V2X-Sim 59.4 $\to$ 67.8) \\
\quad + mask depth, $\alpha=3$ & 68.6 \\
\quad + one-to-one & 72.5 \\
\midrule
Keypoint anchors (ViTPose) + depth, dev.\ \ka & 70.3 \vs 70.5 \\
Attention readout~\cite{kang2025attention} alone, dev.\ \ka & about 9 \\
Attention reranking of the top angular-offset candidates, dev.\ \ka & 57.7 \vs 67.8 \\
Appearance alone (DINOv3), Acc & about 24 (chance) \\
Appearance on near ties / added, Acc & 68.0 / 59.7 \vs 68.1 \\
Weighting by the disagreement of the two input orders, dev.\ Acc & 76.9--77.4 \vs 77.4 \\
Learned combination of cues, rank-16 model, Gold unseen Acc & 78.4 \vs 78.7 \\
Heat-map head, rank-16 model, Gold seen / unseen Acc & 87.7 / 79.2 \vs 87.9 / 79.3 \\
\bottomrule
\end{tabular}
\end{table}

\begin{table}[!htbp]
\centering
\caption{\textbf{Shortcut controls} (zero-shot \vggto, Acc by angular offset). Random ego: an ego image from another scene. Erased: every candidate inpainted out of the exo image.}
\label{tab:short}
\footnotesize
\setlength{\tabcolsep}{3pt}
\begin{tabular}{lcccc}
\toprule
Setting & Chance & Original & Random ego & Erased \\
\midrule
Indoor, fixed & 21.2 & 90.7 & 22.8 & 90.2 \\
Indoor, peer & 31.2 & 89.5 & 28.2 & 89.0 \\
Nature, fixed & 9.4 & 76.5 & 9.8 & 60.0 \\
Nature, peer & 16.5 & 65.7 & 18.8 & 53.5 \\
Birds, fixed & 13.8 & 46.0 & 13.8 & 41.2 \\
Birds, peer & 24.4 & 91.2 & 26.5 & 89.2 \\
Robot arms & 44.3 & 65.7 & 46.5 & 67.0 \\
\bottomrule
\end{tabular}
\end{table}

\paragraph{Use of the ego image.}
\label{sec:supp_egouse}
Following PEN, we measure how strongly models rely on the ego image with EgoRate, $1-(\text{Acc}_\text{noise}-c)/(\text{Acc}-c)$, where $\text{Acc}_\text{noise}$ is the accuracy when the ego image is replaced by uniform noise and $c$ is the chance level (\cref{sec:eval}). Across the 62 sets other than the three locked ones, EgoRate reaches 0.963 zero-shot, 0.975 after pose-supervised training, and 0.974 for \ours, confirming that accuracy drops nearly to chance without the ego image. On TF2023, \ours scores 1.01 (above PEN's 0.96), whereas the zero-shot model reaches 0.85 because its readout still favors certain candidates when the ego image is uninformative. Ego-Exo4D is the exception (EgoRate at most 0.19 for \ours): because the wearer is usually the largest candidate in its exo views, 89--92\% of queries stay correct with noise as the ego image (chance 41.8\%), so its scores say little about the ego view.

\section{Test-time Adaptation}
\label{sec:supp_adapt}

In our test-time adaptation, the candidate constraint drives the accuracy gains on unseen settings while the anchor preserves camera directions. Below, we ablate both, examine which queries change, and evaluate low-data and online regimes (\cref{sec:supp_impl} covers optimization and model selection).

\subsection{Ablations and mechanism}
\label{sec:supp_tta}

In our test-time adaptation ablations (\cref{tab:tta}), every change to the final configuration keeps unseen \ka within 0.7 points except dropping the anchor, which costs 5.0 [4.5, 5.5] points as top views collapse (99.4 to 48.7) and lowers seen and unseen MA@5. Under the same LoRA, schedule, and anchor, entropy minimization~\cite{wang2021tent}, pseudo-labeling~\cite{liang2020shot}, and two-order consistency give little unseen gain compared with the candidate constraint (\cref{tab:tta}). Without retuning, the same adaptation also helps VGGT~\cite{wang2025vggt}, Pi3X~\cite{wang2025pi3}, and Depth Anything~3~\cite{lin2025da3} after pose-supervised training (\cref{sec:supp_bb}). On the Gold tier, adaptation cuts pose-supervised training's unseen lateral error from 5.6 to 3.3 body widths while keeping the seen one at 0.5. Adaptation lowers \ka on the development list (whose domains are all seen) from 91.8 to 88.6 and MA@5 on L0 from 95.8 to 91.7 (\cref{sec:supp_levels}); it also reduces the presence accuracy (VA) of CoMind and CORE4D-Real, two real sets with automatic masks, by 11.4 and 4.9 points (\cref{sec:supp_where}). Sets that informed no design decision still gain (\cref{sec:supp_clean}), including MILUV (\cref{sec:supp_locked}).

\begin{table}[!htbp]
\centering
\caption{\textbf{Test-time adaptation ablations.} Indented rows change one component of the configuration above them. Bold: best per column. Hard minimum: the minimum over the candidates in place of the soft minimum in \cref{eq:cand}.}
\label{tab:tta}
\scriptsize
\setlength{\tabcolsep}{2.4pt}
\resizebox{\ifdim\width>\columnwidth\columnwidth\else\width\fi}{!}{%
\begin{tabular}{lcccc}
\toprule
 & \tabhead{2}{\ka} & \tabhead{2}{MA@5} \\
\cmidrule(lr){2-3}\cmidrule(lr){4-5}
Configuration & Seen & Unseen & Seen & Unseen \\
\midrule
\multicolumn{5}{l}{\emph{From pose-supervised training}} \\
No adaptation & 87.8\ci{1.3} & 72.6\ci{1.0} & 74.5\ci{2.2} & 50.7\ci{0.9} \\
\rowcolor{tabours}
Final configuration (\ours) & 87.4\ci{1.2} & 78.2\ci{1.0} & 72.7\ci{2.3} & 53.2\ci{0.9} \\
\quad hard minimum & 87.5\ci{1.3} & \textbf{78.3}\ci{0.9} & 72.2\ci{2.3} & 53.3\ci{1.0} \\
\quad no anchor & 85.7\ci{1.3} & 73.1\ci{0.9} & 50.0\ci{2.4} & 38.3\ci{0.9} \\
\quad anchor without the epipole term & 86.8\ci{1.2} & 77.7\ci{0.9} & 64.0\ci{2.8} & 48.4\ci{0.8} \\
\quad 100 steps & 87.6\ci{1.2} & 77.5\ci{1.0} & 73.5\ci{2.2} & \textbf{53.4}\ci{0.9} \\
\quad 400 steps & 87.0\ci{1.2} & 77.8\ci{1.0} & 71.4\ci{2.3} & 52.8\ci{0.9} \\
\quad + two-order consistency & 87.4\ci{1.3} & \textbf{78.3}\ci{0.9} & 72.6\ci{2.2} & 53.3\ci{1.0} \\
\quad focal length frozen & 87.1\ci{1.3} & 78.1\ci{1.0} & 69.5\ci{2.5} & 53.1\ci{0.9} \\
\midrule
\multicolumn{5}{l}{\emph{Other objectives, same schedule and anchor}} \\
Entropy, TENT-style~\cite{wang2021tent} & \textbf{88.0}\ci{1.4} & 72.5\ci{1.0} & 74.5\ci{2.2} & 51.1\ci{0.9} \\
Pseudo-labeling, SHOT-style~\cite{liang2020shot} & 87.7\ci{1.4} & 73.2\ci{1.0} & 74.4\ci{2.1} & 50.8\ci{0.9} \\
Two-order consistency only & 87.9\ci{1.4} & 73.2\ci{0.9} & \textbf{74.6}\ci{2.1} & 51.4\ci{0.9} \\
\midrule
\multicolumn{5}{l}{\emph{From \vggto without training}} \\
No adaptation (\ours-ZS) & 60.7\ci{1.2} & 58.6\ci{1.1} & 28.6\ci{1.6} & 28.9\ci{0.9} \\
Adaptation, 800 steps & 73.0\ci{2.0} & 69.4\ci{1.1} & 34.2\ci{1.8} & 32.6\ci{0.7} \\
\quad no anchor & 72.4\ci{1.9} & 61.6\ci{1.1} & 28.8\ci{1.3} & 27.4\ci{0.8} \\
\bottomrule
\end{tabular}}
\end{table}

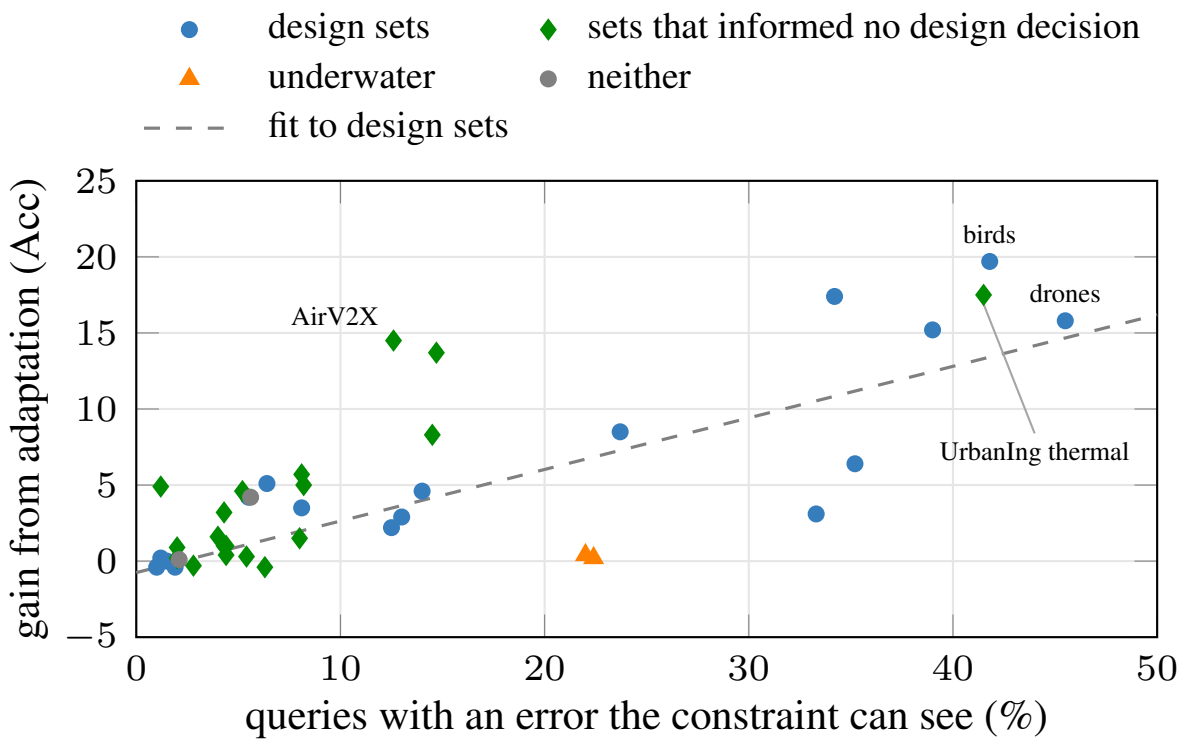
\begin{figure}[!htb]
\centering
\begin{tikzpicture}
\begin{axis}[width=\columnwidth, height=4.6cm, xmin=0, xmax=50, ymin=-5, ymax=25, xtick={0,10,...,50}, ytick={-5,0,...,25},
  xlabel={queries with an error the constraint can see (\%)}, ylabel={gain from adaptation (Acc)},
  tick label style={font=\scriptsize}, label style={font=\footnotesize}, xlabel shift=-3pt, ylabel shift=-4pt, grid=major, grid style={gray!20},
  legend style={font=\scriptsize, draw=none, fill=none, at={(0.5,1.03)}, anchor=south, cells={anchor=west}, row sep=-1pt, column sep=4pt}, legend columns=2,
  clip=false]
\addplot[only marks, mark=*, mark size=1.4pt, color=cvprblue] coordinates {(23.7,8.5) (13.0,2.9) (34.2,17.4) (41.8,19.7) (1.0,-0.4) (1.9,-0.4) (45.5,15.8) (39.0,15.2) (8.1,3.5) (5.5,4.2) (1.2,0.2) (1.5,-0.0) (35.2,6.4) (33.3,3.1) (6.4,5.1) (14.0,4.6) (12.5,2.2)};
\addlegendentry{design sets}
\addplot[only marks, mark=diamond*, mark size=1.8pt, color=green!55!black] coordinates {(12.6,14.5) (5.2,4.6) (4.3,3.2) (4.4,1.0) (4.4,0.4) (14.5,8.3) (8.2,5.0) (8.0,1.5) (5.4,0.3) (6.3,-0.4) (2.0,0.2) (2.0,0.9) (2.8,-0.3) (41.5,17.5) (14.7,13.7) (1.2,4.9) (8.1,5.7) (4.0,1.6) (4.3,1.0)};
\addlegendentry{sets that informed no design decision}
\addplot[only marks, mark=triangle*, mark size=1.9pt, color=orange] coordinates {(22.4,0.2) (22.0,0.4)};
\addlegendentry{underwater}
\addplot[only marks, mark=*, mark size=1.4pt, color=gray] coordinates {(2.1,0.1) (5.6,4.2)};
\addlegendentry{neither}
\addplot[semithick, dashed, gray, domain=0:50, samples=2] {-0.7550 + 0.3390*x};
\addlegendentry{fit to design sets}
\node[font=\tiny, anchor=south, inner sep=1.5pt] at (axis cs:41.8,20.3) {birds};
\node[font=\tiny, anchor=south, inner sep=1.5pt] at (axis cs:45.5,16.4) {drones};
\node[font=\tiny, anchor=south east, inner sep=1.5pt] at (axis cs:12.3,14.9) {AirV2X};
\draw[gray!70, thin] (axis cs:41.5,16.9) -- (axis cs:44.0,8.4);
\node[font=\tiny, anchor=north, inner sep=1.5pt] at (axis cs:44.0,8.4) {UrbanIng thermal};
\end{axis}
\end{tikzpicture}
\caption{\textbf{Adaptation gains where the constraint sees the errors.} Points: sets with at least 20 wrong picks. An error the constraint can see leaves the epipole predicted after pose-supervised training over $2^\circ$ from every candidate.}
\label{fig:gain}
\end{figure}

\Cref{fig:gain} shows where test-time adaptation helps and \cref{tab:carry} how gains carry over to held-out scenes. Because adapting one LoRA across all domains at once adds nothing on unseen settings and costs 4.9 points on seen ones, each deployment adapts on its own stream. Across sets, gains correlate with the share of queries whose pick is wrong and whose epipole lies over $2^\circ$ from every candidate ($r=0.77$; partial $r=0.50$ given the share of all errors). Restricted to queries whose epipole misses every candidate (a selection that needs no labels), the constraint still gains 4.9 of the 5.5 unseen points, showing the shared LoRA also fixes errors it cannot see (31\% of them, versus 34\% of visible ones).

\Cref{tab:mech} breaks down adaptation by domain and tests two simpler explanations. First, which masks serve as candidates matters: although using the whole exo image as the sole candidate helps drones with poor starting poses, evaluating with the angular offset alone (which excludes whole-image depth) shows \ours still leads this control by 5.2 [3.2, 7.3] points on drones and 7.1 [5.1, 8.8] on birds. Second, fitting one exo-camera rotation per set with the same candidate loss and no LoRA gains 0.0 [$-$0.3, 0.3] unseen \ka overall (1.2 [0.4, 1.9] on drones and 0.4 [$-$0.3, 1.1] on birds, while costing vehicles 3.5 [2.2, 4.9]); because these rotations are small (median 0.87$^\circ$ across sets), a single shared rotation does not explain the gain. The anchor preserves directions and prevents collapse, with the epipole term giving the main direction constraint in the final configuration (\cref{tab:mech,tab:tta}).

\begin{table}[!htbp]
\centering
\caption{\textbf{Adaptation carries over to held-out scenes} (Acc, pinhole (v1) lens, held-out half of the scenes). PST: pose-supervised training.}
\label{tab:carry}
\footnotesize
\setlength{\tabcolsep}{4pt}
\begin{tabular}{lccc}
\toprule
Setting & Queries & PST & Adapted on other scenes \\
\midrule
Birds, peer & 116 & 60.1 & 83.0 \\
Drones, peer & 134 & 50.5 & 75.4 \\
Drones, fixed & 142 & 30.8 & 42.3 \\
Birds, fixed & 110 & 33.9 & 47.0 \\
Wild animals, fixed & 158 & 44.3 & 49.6 \\
Football, fixed & 158 & 79.3 & 80.8 \\
Robot arms & 203 & 68.6 & 75.5 \\
Vehicles, drone view & 114 & 90.4 & 91.5 \\
Football, peer & 175 & 80.2 & 81.5 \\
Wild animals, peer & 148 & 58.1 & 60.6 \\
\bottomrule
\end{tabular}
\end{table}

\begin{table}[!htbp]
\centering
\caption{\textbf{Mechanism of test-time adaptation.} Upper block: changes to the candidates or queries used during adaptation; evaluation always uses the true candidates.}
\label{tab:mech}
\footnotesize
\setlength{\tabcolsep}{2.5pt}
\resizebox{\ifdim\width>\columnwidth\columnwidth\else\width\fi}{!}{%
\begin{tabular}{lcccccc}
\toprule
 & \tabhead{4}{\ka} & \tabhead{2}{MA@5} \\
\cmidrule(lr){2-5}\cmidrule(lr){6-7}
Variant & Unseen & Drones & Birds & Vehicles & Seen & Unseen \\
\midrule
\multicolumn{7}{l}{\emph{Change against pose-supervised training without adaptation}} \\
True candidates (\ours) & +5.5\ci{0.6} & +24.0\ci{3.8} & +18.1\ci{1.9} & +10.3\ci{3.0} & $-$1.8\ci{0.5} & +2.5\ci{0.4} \\
Without the wearer & $-$19.7\ci{0.9} & +18.4\ci{3.7} & +9.9\ci{2.0} & $-$22.8\ci{3.0} & $-$37.8\ci{1.1} & $-$26.6\ci{0.8} \\
Random candidates & $-$12.6\ci{0.9} & +20.7\ci{3.5} & +11.6\ci{2.2} & $-$20.0\ci{4.5} & $-$46.9\ci{1.2} & $-$31.1\ci{0.8} \\
Candidates of another frame & $-$8.6\ci{1.0} & +21.1\ci{3.5} & +9.4\ci{2.2} & $-$6.9\ci{3.6} & $-$53.8\ci{1.9} & $-$32.7\ci{0.8} \\
Whole image as the only candidate & $-$11.5\ci{1.2} & +19.7\ci{3.4} & +8.6\ci{2.5} & $-$26.0\ci{4.4} & $-$58.2\ci{1.8} & $-$33.7\ci{0.9} \\
Violating queries only & +4.9\ci{0.5} & +23.9\ci{3.0} & +18.6\ci{1.9} & +8.9\ci{2.6} & $-$0.6\ci{0.3} & +2.9\ci{0.4} \\
Satisfied queries only & +1.8\ci{0.4} & +3.5\ci{2.3} & +8.1\ci{1.4} & +5.1\ci{2.1} & +0.0\ci{0.2} & +0.7\ci{0.2} \\
\midrule
\multicolumn{7}{l}{\emph{Change against \ours when one anchor term is dropped}} \\
Rotation term & +0.1\ci{0.3} & +0.2\ci{1.2} & $-$0.2\ci{1.1} & $-$0.5\ci{0.9} & $-$0.8\ci{0.3} & $-$0.1\ci{0.2} \\
Focal term & +0.0\ci{0.3} & $-$0.3\ci{1.1} & +0.3\ci{0.9} & +0.2\ci{1.2} & $-$0.9\ci{0.4} & +0.2\ci{0.2} \\
Epipole term & $-$0.5\ci{0.5} & $-$0.6\ci{1.2} & +0.1\ci{1.1} & $-$0.3\ci{1.4} & $-$8.7\ci{1.5} & $-$4.8\ci{0.5} \\
\bottomrule
\end{tabular}}
\end{table}

\subsection{Adaptation on fewer queries and online}
\label{sec:supp_stream}

\begin{figure}[!htb]
\centering
\begin{tikzpicture}
\begin{axis}[width=\columnwidth, height=3.6cm, xmin=0.5, xmax=10.5, ymin=-0.5, ymax=8.5,
  xtick={1,...,10}, ytick={0,2,4,6,8}, xlabel={tenth of each set's stream, in arrival order}, ylabel={gain in \ka},
  tick label style={font=\scriptsize}, label style={font=\footnotesize}, xlabel shift=-3pt, ylabel shift=-4pt, grid=major, grid style={gray!20},
  legend style={font=\tiny, draw=none, fill=none, at={(0.02,0.98)}, anchor=north west, cells={anchor=west}, row sep=-1.5pt}]
\addplot[thick, color=cvprblue, mark=*, mark size=1pt, error bars/.cd, y dir=both, y explicit] coordinates {(1,0.4) += (0,0.4) -= (0,0.3) (2,2.3) += (0,1.2) -= (0,1.0) (3,2.6) += (0,1.3) -= (0,1.3) (4,2.4) += (0,1.3) -= (0,1.3) (5,4.4) += (0,2.7) -= (0,2.4) (6,3.1) += (0,1.5) -= (0,1.4) (7,3.4) += (0,2.0) -= (0,1.9) (8,2.6) += (0,2.2) -= (0,2.0) (9,3.7) += (0,2.8) -= (0,2.6) (10,3.7) += (0,3.0) -= (0,2.5)};
\addlegendentry{online, each query scored before adapting on it}
\addplot[thick, dashed, color=orange, mark=*, mark size=1pt] coordinates {(1,4.7) (2,4.0) (3,4.3) (4,4.1) (5,5.3) (6,3.7) (7,3.6) (8,3.7) (9,4.3) (10,3.6)};
\addlegendentry{all queries of the set (\ours)}
\end{axis}
\end{tikzpicture}
\caption{\textbf{Online adaptation builds up its gain early in each stream}. Gain over pose-supervised training, pooled over 42,109 queries. Bars: 95\% scene-level confidence intervals.}
\label{fig:online}
\end{figure}
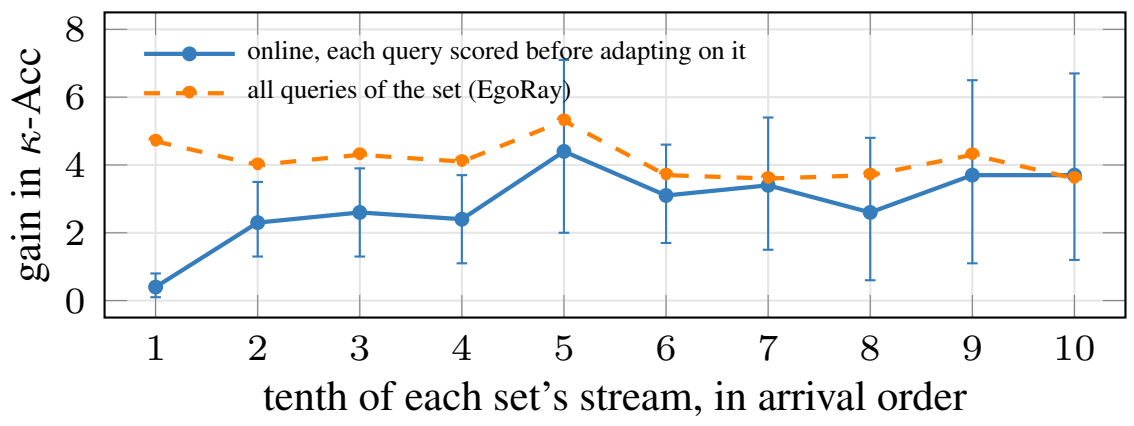

Both few-query and online adaptation retain substantial gains (\cref{tab:stream}). Few-query adaptation updates on a small subset and evaluates on the rest; online adaptation scores each query before updating on it (one update per two queries in arrival order), building its gain over the first three tenths of each stream (\cref{fig:online}). One LoRA per exo camera slightly outperforms one per set on the same queries (\cref{tab:stream}). Adapting on half of each set's scenes and testing on the other keeps 4.5 [3.8, 5.1] of the 5.5 unseen points that adaptation on all Gold queries gains on the same test queries (covering 59 of the 62 sets because three have no adaptation queries in one half). Although adaptation uses no wearer labels, it defaults to each set's Gold Track A queries (visible wearers by definition), so about 90\% of queries scored on the Full tier are new to it. Adapting instead on each set's full test stream---all 579,466 queries with at least two candidates in the 62 sets, regardless of label, including 62,214 absent queries and 20,753 with unusable labels---still yields 4.4 unseen points versus 5.5 on Gold queries.

\begin{table}[!htbp]
\centering
\caption{\textbf{Few-query, online, and per-camera adaptation} (change relative to pose-supervised training). Unseen $-$all: difference from adapting on all queries, evaluated on the same queries. Per exo camera: restricted to cameras with at least 25 queries.}
\label{tab:stream}
\footnotesize
\setlength{\tabcolsep}{2.5pt}
\resizebox{\ifdim\width>\columnwidth\columnwidth\else\width\fi}{!}{%
\begin{tabular}{lccccc}
\toprule
 & \tabhead{3}{\ka} & \tabhead{2}{MA@5} \\
\cmidrule(lr){2-4}\cmidrule(lr){5-6}
Adapted on & Seen & Unseen & Unseen $-$all & Seen & Unseen \\
\midrule
25 queries per set & $-$0.4\ci{0.5} & +3.6\ci{0.6} & $-$1.9\ci{0.5} & $-$1.6\ci{0.4} & +0.3\ci{0.5} \\
50 & $-$0.3\ci{0.5} & +4.4\ci{0.6} & $-$1.1\ci{0.5} & $-$1.7\ci{0.5} & +0.8\ci{0.5} \\
100 & $-$0.4\ci{0.5} & +4.9\ci{0.8} & $-$0.8\ci{0.4} & $-$2.0\ci{0.4} & +1.5\ci{0.4} \\
200 & $-$0.1\ci{0.7} & +4.7\ci{0.8} & $-$0.6\ci{0.6} & $-$1.8\ci{0.5} & +2.6\ci{0.6} \\
Online, one pass & $-$0.6\ci{0.6} & +3.8\ci{0.6} & $-$1.7\ci{0.5} & $-$1.4\ci{0.5} & +2.2\ci{0.5} \\
One LoRA per exo camera & +0.8\ci{2.3} & +4.7\ci{0.9} & +1.4\ci{1.1} & +1.0\ci{1.1} & +1.8\ci{1.1} \\
\midrule
All Gold queries of the set (\ours) & $-$0.5\ci{0.5} & +5.5\ci{0.6} & \no & $-$1.8\ci{0.5} & +2.5\ci{0.4} \\
\bottomrule
\end{tabular}}
\end{table}

\subsection{Examples}

\begin{figure*}[t]
\centering
\qtile{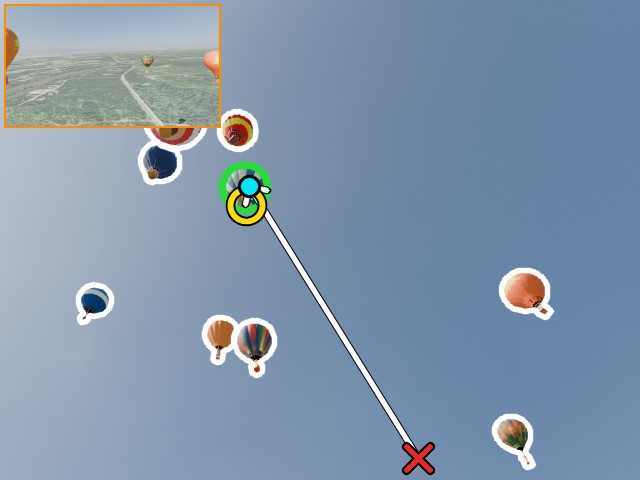}{Drones (balloons)}\hfill\qtile{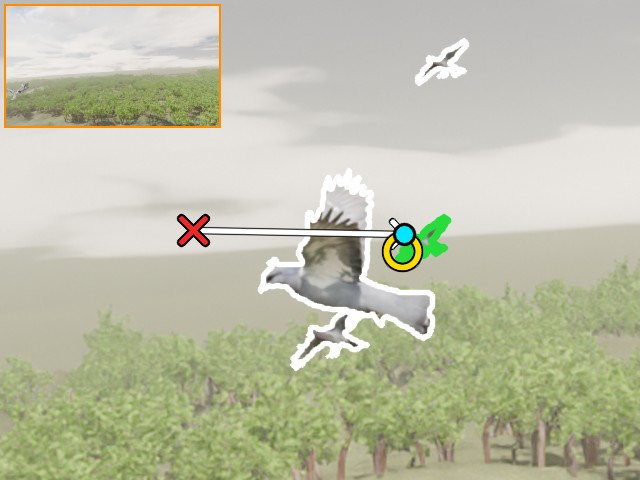}{Birds}\hfill\qtile{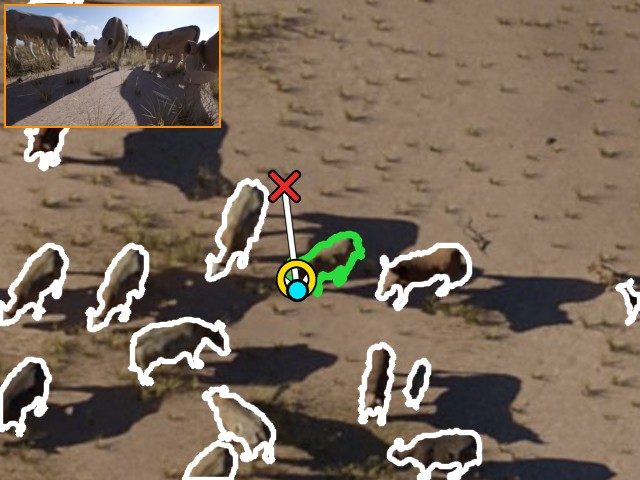}{Wild animals}\\[2pt]
\qtile{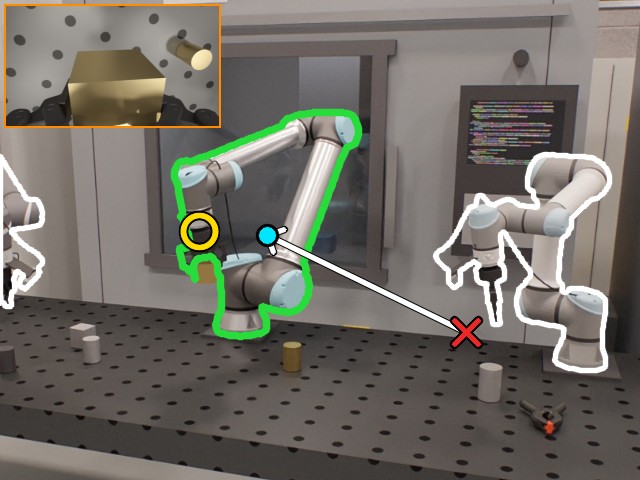}{Robot arms}\hspace{0.01\textwidth}\qtile{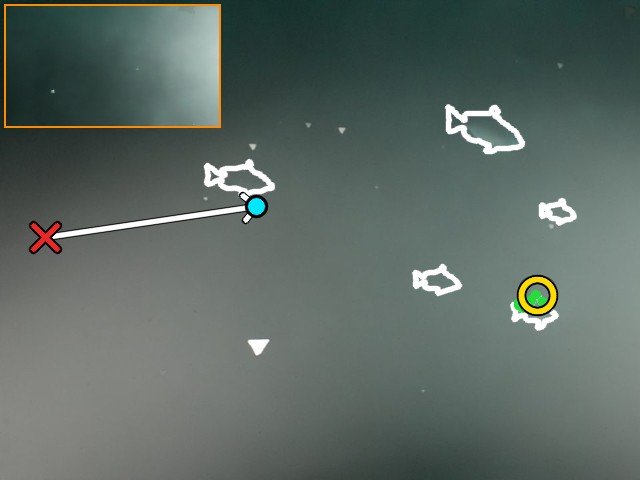}{Underwater (failure)}
\caption{\textbf{Adaptation moves the epipole onto a candidate} (\ours). Red cross: epipole before adaptation. Cyan dot: adapted epipole. Yellow ring: true epipole. Green: the wearer.}
\label{fig:qual}
\end{figure*}

\Cref{fig:qual} illustrates the effect of adaptation using the first queries in hash order whose predicted epipole moves from over $2^\circ$ away from every candidate to within $1.5^\circ$ of the wearer, or onto another candidate for the underwater failure.

\section{Implementation and Model Selection}
\label{sec:supp_impl}

This section details how to reproduce our models and baselines, their speed, and our design and hyperparameter choices.

\paragraph{Inference.}
Given the exo image first, \vggto can place the ego camera behind the exo camera. Because a visible wearer lies in front of it, our readout reflects such predictions in \who through the exo image plane and negates their depth, in every row of \cref{tab:components} as well, whereas \where answers \emph{absent} when the predicted ego camera lies behind the exo camera (\cref{sec:supp_where}). Like the other 3D models, the exo-first \vggto rows of \cref{tab:goldfull,tab:realsim,tab:bbft} do not reflect them. Mutual localization averages and normalizes unit directions from both input orders. In \where, the presence threshold $\theta^\star$ is $12^\circ$ zero-shot and chosen on the development list after pose-supervised training.

\paragraph{Pose-supervised training.}
LoRA adapters (scale 1, no dropout) update the query-key-value and output projections of frame and global attention in \vggto (1.14B parameters), with its encoder and heads frozen. Training runs 65K steps (520K samples) with AdamW~\cite{loshchilov2019adamw} (learning rate $10^{-4}$, weight decay 0.01, one-cycle schedule, 5\% warm-up, batch size 4 per GPU on two GPUs). Images are resized at native aspect ratio to about 1,024 $16\times16$-pixel patches, padded to a common canvas, and color-jittered separately. Each batch randomly picks one of the two input orders, draws absent queries (still pose-supervised) a quarter as often as the others, and samples the nine training groups (four real datasets, five simulated domains) proportionally to the square root of their size. Each sample keeps at most 32 candidates with 96 sampled pixels each, and mask depth is the median over the sampled pixels, or over all mask pixels in \who, which avoids background depth on thin bodies.

Each term of \cref{eq:loss} is averaged over the queries in each GPU's batch that carry its label; focal terms apply only to pinhole (v1) renders, where the true focal length is exact, and are averaged over horizontal and vertical focal lengths. Whereas the readout evaluates the score over every mask pixel and computes the angular offset over at most 3,000 sampled pixels, the losses evaluate the score at the centroid of the mask's upper third and the angular offset at the nearest of the 96 sampled pixels; taking the score at the best of those 96 pixels instead changes \ours's \ka by at most 0.3 points on seen and unseen settings, with both intervals including zero.

\paragraph{Test-time adaptation.}
Adaptation runs 200 AdamW steps per test set (batch size 2, learning rate $2\times10^{-5}$, gradient clipping at 1) using the starting model's temperatures. When adaptation starts from \vggto without pose-supervised training, a new adapter is trained for 800 steps with the initial temperatures of pose-supervised training, $\tau_a=0.05$\,rad and $\tau_m=0.1$. The last term of \cref{eq:anchor} is averaged over qualifying queries and divided by $\tau_a$ so that a small drift costs about as much as an equal-angle violation. For the joint input of \cref{sec:supp_joint}, multi-view training continues for 12K steps with color jitter only on one exo image and up to three ego images of its frame; adaptation uses up to six ego images per pass, while box-annotated sets keep pair input.

\paragraph{Reproduction and speed.}
An external colleague reran six sets (12 settings) on other hardware, matching our results within 0.5 points before adaptation and differing afterward by +0.1 on average (standard deviation 1.0). On one A800 (batch size one), both input orders take 0.17--0.21\,s per pair, the readout adds at most 0.06\,s on one CPU core, and adaptation takes about 160\,s per set.

\subsection{Baselines and input orders}

Feature-matching baselines (\cref{tab:who} and \cref{sec:supp_locked,sec:supp_pen}) estimate an essential matrix via the five-point solver~\cite{nister2004efficient} in MAGSAC++~\cite{barath2020magsacpp} (1\,px threshold, at least 15 inliers), forming rays from \vggto's predicted intrinsics rather than calibrated ones and using angular offset alone. Unsolved pairs stay at chance in \who, affecting SIFT and SuperPoint with LightGlue most (\cref{sec:supp_locked,sec:supp_pen}). Models follow their released input preparation: official SuperPoint with LightGlue, DUSt3R, and MASt3R keep native aspect ratios, VGGT uses pad mode, and CUT3R (needing one shape for both views) takes the ego image center-cropped to the exo shape. With the values of $\alpha$ chosen on the development list ($\alpha=3$ for Pi3X, $\alpha=1$ for MapAnything), two input orders change neither model by more than about one point, so \cref{tab:who} uses one order (exo image first) for every baseline.

\subsection{Model selection}
\label{sec:supp_design}

\Cref{tab:design} records each design decision and the data behind it. On the development list (13 sets and 3,600 queries, all from seen domains), we chose the readout (apart from the later switch to the mask minimum), all thresholds, and the rank, budget, and color jitter of pose-supervised training, and we chose the backbone from zero-shot records. We use rank 64 and twice the rank-16 budget; color jitter improves accuracy, whereas geometric jitter (mirroring, roll, and zoom) does not. We added the focal term after predicted focal lengths collapsed on Gold sets, where omitting it lowers seen MA@5 by about 40\% at rank 16 and half the final budget. After fixing the design, we tested two alternatives from \cref{sec:supp_comp} on Gold with the rank-16 model, both fitted on the development list, and adopted neither: a conditional logit over readout cues and a heat-map head predicting the wearer's mask from frozen tokens and each patch's position relative to the predicted epipole (score weight $\beta=0.2$).

Because tuning test-time adaptation on the test stream is a known pitfall~\cite{zhao2023ttab}, we chose its components and step count by cross-evaluation on the Gold sets that prompted them (\cref{tab:crosseval}), while \cref{tab:design} records how the learning rate and the 800 steps from \vggto were chosen. We hash each set's scenes into two halves, choosing on one and scoring on the other under a rule fixed beforehand, although whole-set results for the options were already known. Each choice varies one element of a reference configuration: a hard-minimum candidate constraint, the anchor, 200 steps, and a null proposal for \where. A component stays only if its paired 95\% gain interval on the deciding half exceeds zero---in \ka, in VA for the null proposal, and in MA@5 for the anchor (without significantly lowering \ka)---and a value such as the step count changes only if it beats the default with such an interval. Both halves agree except on the anchor, which we keep for direction accuracy.

\begin{table*}[t]
\centering
\caption{\textbf{Design record.} $^{a}$Bird flocks, drones, football, perching birds, underwater, wild herds, and the first robot-arm set (all v1), plus nature (v2). $^{b}$Those of $^{a}$ except perching birds and underwater, plus CSRD virtual, CVMHAT, V2X-Sim roadside, and city. $^{c}$The first robot-arm set, bird flocks, city, drones, football, underwater, and V2X-Sim roadside (all v1), plus nature (v2).}
\label{tab:design}
\footnotesize
\setlength{\tabcolsep}{3pt}
\begin{tabular}{>{\raggedright\arraybackslash}p{0.25\textwidth}>{\raggedright\arraybackslash}p{0.24\textwidth}>{\raggedright\arraybackslash}p{0.25\textwidth}>{\raggedright\arraybackslash}p{0.2\textwidth}}
\toprule
Decision & Prompted by & Chosen on & Test sets seen \\
\midrule
Readout: angular offset, mask depth, $\alpha$ & zero-shot records of depth degradation at long range and candidates on one ray indoors (\cref{tab:crossover,tab:pose}) & development list & zero-shot records of all sets then available, as motivation only \\
Minimum over the mask in \cref{eq:mix}, instead of the centroid of its upper third & cameras sit low on robot arms, animals, and drones (\cref{fig:camloc}) & after freezing, and macro averages move by at most 0.6 points on the Gold tier (\cref{sec:supp_comp}) & all, as a check after freezing \\
Two input orders & ego cameras predicted behind the V2X-Sim roadside camera with the exo image first & development list & V2X-Sim roadside \\
Backbone: \vggto & VGGT fails when the cameras face each other on Harmony4D & zero-shot comparison, and after freezing, VGGT and Pi3X trained the same way stay below it (\cref{sec:supp_bb}) & zero-shot records of TF2023 and Harmony4D, then all \\
Wearer loss in training & the readout scores candidates through the predicted epipole & part of the first recipe, ablated only afterwards (\cref{sec:supp_ftdata}) & all, as ablations \\
Focal term in training & the predicted focal length collapsed in nature & added after the diagnosis, ablated only afterwards (\cref{sec:supp_impl}) & nature, volleyball, indoor, and bird flocks, then all for the check \\
Joint input & little shared environment in underwater pairs (\cref{tab:env}) & number of ego images on test sets, final variant on the development list & birds, drones, and underwater (eight settings) \\
\midrule
Candidate constraint & true-epipole diagnosis (\cref{tab:pose}) and bird and drone errors far from every candidate & cross-evaluation (\cref{tab:crosseval}) & the eight$^{a}$ \\
Soft minimum & \no & cross-evaluation, against the hard minimum & the ten$^{b}$ \\
Anchor & without it, rotation error in nature doubles and the exo focal length grows & cross-evaluation (one of two halves), kept for directions, and on the development list alone, MA@5 falls from 73.6 to 43.6 without it & the eight$^{a}$ \\
No consistency term, focal length free & simpler & cross-evaluation & the eight$^{a}$ \\
200 steps (800 from \vggto) & \no & cross-evaluation, and 800 on all Gold sets (also the best of 200, 400, and 800 on the development list) & all \\
Learning rate (batch size set once) & \no & development list, with $10^{-5}$ and $4\times10^{-5}$ checked on all Gold sets after freezing, within 1.1 points & all, as a check after freezing \\
No null proposal for \where & \no & cross-evaluation & the eight$^{c}$ \\
One LoRA per test set & a test set is one deployment & fixed in advance, and checked after freezing on a stream of all sets (\cref{sec:supp_tta}), per exo camera, and on 25--200 queries per set (\cref{sec:supp_stream}) & all, as a check \\
\bottomrule
\end{tabular}
\end{table*}

\begin{table}[!htb]
\centering
\caption{\textbf{Cross-evaluation of adaptation.} Parentheses: chosen minus closest alternative (\ka). Minimum: hard minus soft. Anchor (\ka / MA@5) and null proposal (VA): on minus off. A component stays only if its gain on the deciding half has a paired 95\% interval above zero, so a positive difference can still leave it off.}
\label{tab:crosseval}
\footnotesize
\setlength{\tabcolsep}{2.5pt}
\resizebox{\ifdim\width>\columnwidth\columnwidth\else\width\fi}{!}{%
\begin{tabular}{llll}
\toprule
Decision & Half A & Half B & Final \\
\midrule
Objective & constraint (+6.3) & constraint (+5.1) & constraint \\
Minimum & soft ($-$0.1) & soft ($-$0.2) & soft \\
Anchor & on ($-$0.3 / +3.2) & off ($-$0.9 / +4.2) & on \\
Consistency term, focal length & off, free & off, free & off, free \\
Steps & 200 & 200 & 200 \\
Null proposal for \where & off (+0.3) & off (+0.7) & off \\
\bottomrule
\end{tabular}}
\end{table}

\subsection{Adaptation with hyperparameters chosen on half of the scenes}
\label{sec:supp_clean}

With hyperparameters chosen by cross-evaluation on the opposite scene half (\cref{sec:supp_design}), adaptation adds 7.2 [6.3, 8.1] unseen \ka over pose-supervised training when adapting on the scored half and 6.4 [5.5, 7.4] without seeing the scored scenes, keeping most of \ours's gain (\cref{tab:cleangain}). Directions worsen because the selection on one half drops the anchor: the anchored half gains 5.8 unseen MA@5, whereas the other loses 4.6. Under this protocol, entropy minimization, pseudo-labeling, and two-order consistency add at most 1.0 unseen \ka versus 7.2 for the candidate constraint. Adapting on the scored half gains most on bird flocks (21.9) and drones (20.8) and loses on underwater wearers ($-$1.5 [$-$2.5, $-$0.5]). On the 27 sets that informed no decision of \ours, adaptation adds 2.2 [1.7, 2.7] \ka from a starting score of 89.2. Because adaptation lowers \ka on the seen-only development list (\cref{sec:supp_tta}), that list can rank safeguards but not decide whether to adapt.

\begin{table}[t]
\centering
\caption{\textbf{Adaptation with hyperparameters chosen on half of the scenes} (18 Gold sets from the families that informed the design). Each fold uses one configuration chosen by cross-evaluation on the other scene half, adapting on the scored or other half. Indented rows: changes against pose-supervised training. Lower block: differences from \ours.}
\label{tab:cleangain}
\scriptsize
\setlength{\tabcolsep}{2pt}
\resizebox{\ifdim\width>\columnwidth\columnwidth\else\width\fi}{!}{%
\begin{tabular}{@{}lcccc@{}}
\toprule
 & \tabhead{2}{\ka} & \tabhead{2}{MA@5} \\
\cmidrule(lr){2-3}\cmidrule(lr){4-5}
 & Seen & Unseen & Seen & Unseen \\
\midrule
Pose-supervised training & 72.0\ci{3.3} & 61.0\ci{1.5} & 51.1\ci{2.5} & 32.9\ci{1.2} \\
\quad adapted on the scored half & $-$0.7\ci{0.8} & +7.2\ci{0.9} & $-$7.0\ci{2.4} & +0.4\ci{1.8} \\
\quad adapted on the other half & $-$1.4\ci{0.9} & +6.4\ci{0.9} & $-$7.1\ci{2.7} & +1.7\ci{1.3} \\
\quad \ours, all Gold queries & +0.0\ci{0.8} & +7.8\ci{0.8} & $-$1.1\ci{0.6} & +5.8\ci{0.6} \\
\midrule
Scored half $-$ \ours & $-$0.7\ci{0.7} & $-$0.6\ci{0.5} & $-$5.8\ci{2.5} & $-$5.4\ci{1.8} \\
Other half $-$ \ours & $-$1.4\ci{1.1} & $-$1.4\ci{0.5} & $-$6.0\ci{2.8} & $-$4.1\ci{1.3} \\
\bottomrule
\end{tabular}}
\end{table}

\section{Detailed Results}
\label{sec:supp_tables}

\ours leads every domain of \cref{tab:who}. Below, we break down the main tables by domain, generalization level, task, and individual real set, and report results on sets that informed no design decisions.

\subsection{Domain results and intervals}
\label{sec:supp_full}

\Cref{tab:mainci} provides scene-level confidence intervals for the main tables, \cref{tab:fullci} reports per-domain results with intervals for our models, and \cref{tab:compdom} breaks down \cref{tab:components} by domain. Resampling domains instead of scenes gives an unseen adaptation gain of 5.5 [2.1, 10.2] because four domains carry most of it, and resampling domains then scenes within them gives [2.0, 10.1], excluding zero. Leaving out any one domain yields an unseen gain of 4.0 (without drones) to 6.1, and excluding the human wearers in the drone and legged-robot domains (\cref{sec:supp_protocol}) produces 5.9. Nature has only two test scenes; in \cref{tab:fullci}, test scenes are counted across sets because v1 and v2 share scenes, and the half-widths of 0.0 in the top-view domain come from scores near the ceiling.

\begin{table*}[!tp]
\centering
\caption{\textbf{Intervals of the main results} (Full tier). Tabs.~\ref{tab:who}, \ref{tab:where}, and \ref{tab:exo} with the half-width of the 95\% scene-level confidence interval as superscript ($\pm$). $^\ast$: angular offset only, as in \cref{tab:who}.}
\label{tab:mainci}
\footnotesize
\setlength{\tabcolsep}{4pt}
\begin{tabular}{lcccccc}
\toprule
 & \tabhead{2}{\Who, \ka} & \tabhead{2}{\Where, \giou} & \tabhead{2}{\Exo, MA@5} \\
\cmidrule(lr){2-3}\cmidrule(lr){4-5}\cmidrule(lr){6-7}
Method & Seen & Unseen & Seen & Unseen & Seen & Unseen \\
\midrule
Largest mask & 6.5\ci{0.5} & 0.9\ci{0.4} & \no & \no & \no & \no \\
\midrule
DUSt3R~\cite{wang2024dust3r} & 26.5\ci{1.4} & 24.6\ci{1.3} & \no & \no & \no & \no \\
MASt3R~\cite{leroy2024mast3r} & 39.6\ci{1.6} & 33.4\ci{1.1} & \no & \no & 15.9\ci{1.2} & 14.2\ci{0.7} \\
CUT3R~\cite{wang2025cut3r} & 31.2\ci{1.2} & 24.4\ci{1.0} & \no & \no & \no & \no \\
VGGT~\cite{wang2025vggt} & 38.4\ci{1.1} & 35.9\ci{1.2} & \no & \no & 12.0\ci{0.9} & 13.2\ci{0.7} \\
Pi3X~\cite{wang2025pi3} & 49.0\ci{0.8} & 50.1\ci{1.0} & 49.2\ci{0.8} & 42.3\ci{0.6} & 14.1\ci{0.6} & 22.0\ci{0.6} \\
MapAnything~\cite{keetha2025mapanything} & 51.3\ci{1.3} & 49.6\ci{1.0} & 48.4\ci{1.0} & 41.1\ci{0.6} & 21.0\ci{1.1} & 19.7\ci{0.7} \\
Depth Anything~3~\cite{lin2025da3} & 50.7\ci{1.4} & 48.1\ci{1.2} & 51.6\ci{0.9} & 42.2\ci{0.6} & 29.2\ci{1.5} & 23.4\ci{0.7} \\
Reloc3r~\cite{dong2025reloc3r}$^\ast$ & 33.4\ci{1.3} & 32.7\ci{1.2} & \no & \no & 14.1\ci{1.0} & 12.2\ci{0.7} \\
RoMa~\cite{edstedt2024roma} + E$^\ast$ & 28.5\ci{1.3} & 24.2\ci{0.9} & \no & \no & \no & \no \\
MASt3R matches + E$^\ast$ & 41.4\ci{1.4} & 37.1\ci{1.0} & \no & \no & \no & \no \\
\vggto, dense tokens + PnP & 19.2\ci{1.0} & 15.5\ci{0.8} & \no & \no & \no & \no \\
\midrule
\rowcolor{tabourslight}
\ours-ZS & 60.5\ci{1.1} & 58.7\ci{0.9} & 56.3\ci{0.9} & 46.5\ci{0.6} & 28.5\ci{1.5} & 28.1\ci{0.7} \\
\rowcolor{tabours}
\ours$^\ddagger$ & 87.1\ci{1.2} & 77.7\ci{0.9} & 67.8\ci{1.4} & 55.2\ci{0.6} & 71.8\ci{2.0} & 50.6\ci{0.8} \\
\midrule
\ours $-$ \ours-ZS & +26.6\ci{0.8} & +19.0\ci{0.6} & +11.6\ci{0.8} & +8.6\ci{0.4} & +43.3\ci{1.5} & +22.5\ci{0.8} \\
\bottomrule
\end{tabular}
\end{table*}

\begin{table}[t]
\centering
\caption{\textbf{\Who per domain} (\ka). Scenes: test scenes per domain. PST: pose-supervised training. Gain: \ours minus PST. Extension: \cref{sec:supp_joint}.}
\label{tab:fullci}
\footnotesize
\setlength{\tabcolsep}{2.5pt}
\resizebox{\ifdim\width>\columnwidth\columnwidth\else\width\fi}{!}{%
\begin{tabular}{lrccccc}
\toprule
Domain & Scenes & \ours-ZS & PST & \ours$^\ddagger$ & Gain & Extension$^\ddagger$ \\
\midrule
\multicolumn{7}{l}{\emph{Seen (L0--L1)}} \\
Real humans & 170 & 93.2\ci{2.0} & 98.9\ci{0.6} & 98.6\ci{1.0} & $-$0.3\ci{0.8} & 98.6\ci{0.8} \\
Indoor & 54 & 90.4\ci{2.3} & 98.3\ci{0.8} & 97.0\ci{1.1} & $-$1.2\ci{0.8} & 97.6\ci{0.9} \\
Nature & 2 & 72.9\ci{1.3} & 93.7\ci{1.3} & 94.3\ci{1.3} & +0.5\ci{0.0} & 95.2\ci{0.1} \\
City & 20 & 67.6\ci{3.3} & 93.0\ci{1.1} & 92.3\ci{1.6} & $-$0.7\ci{0.9} & 94.6\ci{1.3} \\
Divers & 8 & 11.2\ci{3.6} & 52.8\ci{6.7} & 53.1\ci{6.1} & +0.3\ci{2.2} & 63.0\ci{8.6} \\
Thermal & 12 & 29.1\ci{3.6} & 90.1\ci{3.2} & 88.8\ci{4.1} & $-$1.3\ci{1.3} & 92.9\ci{3.0} \\
\midrule
\multicolumn{7}{l}{\emph{Unseen source (L2)}} \\
Real humans & 459 & 93.4\ci{1.5} & 98.8\ci{0.6} & 98.1\ci{0.5} & $-$0.7\ci{0.4} & 98.9\ci{0.6} \\
Top view & 16 & 85.1\ci{0.8} & 98.9\ci{0.0} & 99.4\ci{0.0} & +0.5\ci{0.0} & 99.3\ci{0.0} \\
Sports & 96 & 59.7\ci{1.4} & 82.7\ci{1.3} & 86.1\ci{1.0} & +3.3\ci{0.9} & 88.4\ci{0.9} \\
Unreal & 37 & 70.4\ci{4.5} & 88.2\ci{3.4} & 88.2\ci{3.0} & +0.0\ci{0.7} & 88.7\ci{3.4} \\
\midrule
\multicolumn{7}{l}{\emph{Unseen wearer (L3)}} \\
Vehicles & 488 & 50.2\ci{4.4} & 75.2\ci{3.4} & 85.5\ci{3.6} & +10.3\ci{2.9} & 85.2\ci{2.9} \\
Legged & 113 & 79.0\ci{3.6} & 89.2\ci{2.3} & 89.9\ci{2.9} & +0.7\ci{1.5} & 92.9\ci{2.0} \\
Robot arms & 50 & 47.0\ci{8.8} & 56.7\ci{8.3} & 66.9\ci{7.5} & +10.1\ci{5.2} & 48.6\ci{8.2} \\
Vacuums & 59 & 87.5\ci{2.4} & 91.9\ci{1.9} & 94.1\ci{1.6} & +2.2\ci{1.1} & 95.6\ci{1.6} \\
Pets & 80 & 95.6\ci{2.3} & 98.3\ci{1.3} & 98.3\ci{1.4} & +0.1\ci{1.1} & 99.4\ci{0.7} \\
Wild animals & 48 & 28.1\ci{2.6} & 53.8\ci{2.8} & 57.2\ci{2.8} & +3.5\ci{2.4} & 54.8\ci{2.9} \\
Fish & 36 & 14.8\ci{4.3} & 49.3\ci{4.4} & 49.1\ci{5.0} & $-$0.2\ci{2.2} & 58.1\ci{5.6} \\
Birds & 24 & 25.4\ci{3.0} & 30.0\ci{3.9} & 48.2\ci{3.2} & +18.1\ci{1.9} & 57.5\ci{3.8} \\
Drones & 14 & 25.6\ci{3.4} & 31.2\ci{5.0} & 55.2\ci{3.1} & +24.0\ci{3.7} & 63.1\ci{5.1} \\
\midrule
\multicolumn{7}{l}{\emph{Macro}} \\
Seen & \no & 60.7\ci{1.2} & 87.8\ci{1.2} & 87.4\ci{1.3} & $-$0.5\ci{0.5} & 90.3\ci{1.5} \\
Unseen & \no & 58.6\ci{1.1} & 72.6\ci{1.0} & 78.2\ci{0.9} & +5.5\ci{0.6} & 79.3\ci{1.1} \\
\bottomrule
\end{tabular}}
\end{table}

\begin{table*}[!tp]
\centering
\caption{\textbf{Components of \ours per domain} (\ka). Rows and $\Delta$ as in \cref{tab:components}, domains as in \cref{tab:fullci}. The last row keeps pair input on box-annotated sets.}
\label{tab:compdom}
\footnotesize
\setlength{\tabcolsep}{3pt}
\begin{tabular*}{\textwidth}{@{\extracolsep{\fill}}l*{10}{c}}
\toprule
 & \tabhead{6}{Seen (L0--L1)} & \tabhead{4}{Unseen source (L2)} \\
\cmidrule(lr){2-7}\cmidrule(lr){8-11}
Component & Real & Indoor & Nature & City & Divers & Therm. & Real & Top & Sports & Unreal \\
\midrule
\vggto camera head, angular offset & 87.4 & 77.6 & 61.2 & 61.6 & 12.9 & 27.1 & 82.9 & 82.8 & 52.2 & 60.1 \\
+ camera distance (mask depth) & 93.1 & 90.9 & 73.7 & 67.5 & 11.3 & 28.7 & 92.6 & 82.7 & 58.9 & 69.8 \\
\quad $\Delta$ & +5.7 & +13.3 & +12.5 & +5.9 & $-$1.7 & +1.7 & +9.6 & $-$0.1 & +6.6 & +9.6 \\
+ two input orders (\ours-ZS) & 93.2 & 90.4 & 72.9 & 67.6 & 11.2 & 29.1 & 93.4 & 85.1 & 59.7 & 70.4 \\
\quad $\Delta$ & +0.1 & $-$0.4 & $-$0.8 & +0.0 & $-$0.1 & +0.4 & +0.9 & +2.4 & +0.8 & +0.6 \\
+ pose-supervised training, pose labels only & 97.6 & 94.2 & 89.4 & 88.0 & 53.3 & 89.2 & 97.4 & 98.8 & 83.6 & 83.1 \\
\quad $\Delta$ & +4.4 & +3.8 & +16.5 & +20.5 & +42.1 & +60.0 & +4.0 & +13.7 & +23.8 & +12.7 \\
+ wearer loss (pose-supervised training) & 98.9 & 98.3 & 93.7 & 93.0 & 52.8 & 90.1 & 98.8 & 98.9 & 82.7 & 88.2 \\
\quad $\Delta$ & +1.3 & +4.1 & +4.3 & +5.0 & $-$0.4 & +0.9 & +1.4 & +0.1 & $-$0.8 & +5.1 \\
+ adaptation (\ours)$^\ddagger$ & 98.6 & 97.0 & 94.3 & 92.3 & 53.1 & 88.8 & 98.1 & 99.4 & 86.1 & 88.2 \\
\quad $\Delta$ & $-$0.3 & $-$1.2 & +0.5 & $-$0.7 & +0.3 & $-$1.3 & $-$0.7 & +0.5 & +3.3 & +0.0 \\
+ other cameras$^\ddagger$ & 98.6 & 97.6 & 95.2 & 94.6 & 63.0 & 92.9 & 98.9 & 99.3 & 88.4 & 88.7 \\
\quad $\Delta$ & +0.0 & +0.6 & +0.9 & +2.2 & +9.9 & +4.1 & +0.7 & $-$0.1 & +2.3 & +0.5 \\
\bottomrule
\end{tabular*}\\[6pt]
\begin{tabular*}{\textwidth}{@{\extracolsep{\fill}}lccccccccccc}
\toprule
 & \tabhead{9}{Unseen wearer (L3)} & \tabhead{2}{Macro} \\
\cmidrule(lr){2-10}\cmidrule(lr){11-12}
Component & Veh. & Legged & Arm & Vac. & Pet & Wild & Fish & Bird & Drone & Seen & Unseen \\
\midrule
\vggto camera head, angular offset & 47.5 & 73.5 & 40.9 & 84.7 & 90.7 & 23.7 & 19.6 & 40.4 & 38.0 & 54.6 & 56.7 \\
+ camera distance (mask depth) & 47.8 & 78.7 & 44.8 & 87.7 & 95.2 & 29.4 & 16.4 & 25.1 & 26.8 & 60.9 & 58.1 \\
\quad $\Delta$ & +0.3 & +5.2 & +3.9 & +3.0 & +4.5 & +5.8 & $-$3.2 & $-$15.4 & $-$11.2 & +6.2 & +1.4 \\
+ two input orders (\ours-ZS) & 50.2 & 79.0 & 47.0 & 87.5 & 95.6 & 28.1 & 14.8 & 25.4 & 25.6 & 60.7 & 58.6 \\
\quad $\Delta$ & +2.4 & +0.3 & +2.2 & $-$0.1 & +0.3 & $-$1.3 & $-$1.7 & +0.3 & $-$1.2 & $-$0.1 & +0.5 \\
+ pose-supervised training, pose labels only & 74.7 & 86.5 & 54.2 & 91.3 & 97.3 & 49.4 & 50.0 & 31.3 & 26.0 & 85.3 & 71.1 \\
\quad $\Delta$ & +24.5 & +7.5 & +7.2 & +3.8 & +1.8 & +21.2 & +35.3 & +5.9 & +0.4 & +24.6 & +12.4 \\
+ wearer loss (pose-supervised training) & 75.2 & 89.2 & 56.7 & 91.9 & 98.3 & 53.8 & 49.3 & 30.0 & 31.2 & 87.8 & 72.6 \\
\quad $\Delta$ & +0.5 & +2.6 & +2.5 & +0.5 & +0.9 & +4.4 & $-$0.7 & $-$1.2 & +5.1 & +2.5 & +1.6 \\
+ adaptation (\ours)$^\ddagger$ & 85.5 & 89.9 & 66.9 & 94.1 & 98.3 & 57.2 & 49.1 & 48.2 & 55.2 & 87.4 & 78.2 \\
\quad $\Delta$ & +10.3 & +0.7 & +10.1 & +2.2 & +0.1 & +3.5 & $-$0.2 & +18.1 & +24.0 & $-$0.5 & +5.5 \\
+ other cameras$^\ddagger$ & 85.2 & 92.9 & 48.6 & 95.6 & 99.4 & 54.8 & 58.1 & 57.5 & 63.1 & 90.3 & 79.3 \\
\quad $\Delta$ & $-$0.3 & +3.1 & $-$18.3 & +1.6 & +1.1 & $-$2.4 & +9.0 & +9.3 & +7.9 & +3.0 & +1.1 \\
\bottomrule
\end{tabular*}
\end{table*}

\subsection{Results by level and for real and simulated sets}
\label{sec:supp_levels}

Across generalization levels and across both real and simulated sets, \ours leads the compared baselines, whereas the end-to-end appearance model generalizes poorly to unseen wearer types (\cref{tab:levelsfull,tab:realsim,tab:levels}).

\begin{table}[!htb]
\centering
\caption{\textbf{\Who by level of generalization} (\ka, Full tier). DINOv3 is trained end to end on \dataset (\cref{sec:supp_train}), and the 3D models are the three strongest of \cref{tab:who}.}
\label{tab:levelsfull}
\footnotesize
\setlength{\tabcolsep}{3pt}
\begin{tabular}{lcccc}
\toprule
Method & L0 & L1 & L2 & L3 \\
\midrule
DINOv3~\cite{simeoni2025dinov3}, end to end & 93.5\ci{3.1} & 62.2\ci{1.9} & 30.6\ci{1.4} & 11.0\ci{0.7} \\
Pi3X~\cite{wang2025pi3} & 80.2\ci{4.1} & 50.4\ci{0.8} & 68.1\ci{1.8} & 42.1\ci{1.2} \\
MapAnything~\cite{keetha2025mapanything} & 83.5\ci{2.9} & 51.8\ci{1.4} & 67.4\ci{2.0} & 41.8\ci{1.1} \\
Depth Anything~3~\cite{lin2025da3} & 67.4\ci{8.8} & 53.3\ci{1.1} & 66.4\ci{2.7} & 40.0\ci{1.2} \\
\midrule
\rowcolor{tabourslight}
\ours-ZS & 89.1\ci{3.0} & 61.2\ci{1.0} & 76.5\ci{1.2} & 50.8\ci{1.2} \\
\rowcolor{tabours}
\ours$^\ddagger$ & \textbf{97.3}\ci{1.2} & \textbf{87.3}\ci{1.2} & \textbf{92.6}\ci{0.7} & \textbf{71.1}\ci{1.2} \\
\bottomrule
\end{tabular}
\end{table}

\begin{table}[t]
\centering
\caption{\textbf{Real and simulated sets evaluated separately} (\ka). Real seen consists of one domain. Improvement (adaptation): \ours minus pose-supervised training. As for the other 3D models, the exo-first \vggto row does not reflect ego cameras predicted behind the exo camera, unlike \cref{tab:components} (\cref{sec:supp_impl}).}
\label{tab:realsim}
\footnotesize
\setlength{\tabcolsep}{2.5pt}
\resizebox{\ifdim\width>\columnwidth\columnwidth\else\width\fi}{!}{%
\begin{tabular}{lcccc}
\toprule
 & \tabhead{2}{Real} & \tabhead{2}{Simulated} \\
\cmidrule(lr){2-3}\cmidrule(lr){4-5}
Method & Seen & Unseen & Seen & Unseen \\
\midrule
VGGT & 84.6\ci{2.8} & 66.3\ci{4.1} & 29.3\ci{1.3} & 28.0\ci{1.3} \\
Pi3X & 86.5\ci{3.1} & 73.6\ci{3.9} & 41.0\ci{1.2} & 46.5\ci{1.2} \\
MapAnything & 85.9\ci{3.0} & 78.3\ci{4.7} & 45.3\ci{1.8} & 45.4\ci{1.1} \\
Depth Anything~3 & 79.0\ci{5.8} & 59.6\ci{7.2} & 45.1\ci{1.5} & 45.8\ci{1.1} \\
\vggto, camera head & 93.4\ci{2.0} & 83.1\ci{2.3} & 54.5\ci{1.5} & 52.9\ci{1.1} \\
\midrule
\rowcolor{tabourslight}
\ours-ZS & 93.2\ci{2.0} & 83.8\ci{2.3} & 54.2\ci{1.4} & 53.6\ci{1.1} \\
\rowcolor{tabours}
\ours$^\ddagger$ & \textbf{98.6}\ci{1.0} & \textbf{94.5}\ci{1.7} & \textbf{85.1}\ci{1.6} & \textbf{76.4}\ci{1.1} \\
\midrule
Improvement (adaptation) & $-$0.3\ci{0.8} & +2.8\ci{1.6} & $-$0.5\ci{0.6} & +6.2\ci{0.7} \\
\bottomrule
\end{tabular}}
\end{table}

\begin{table}[!htb]
\centering
\caption{\textbf{Results by level of generalization.} ID: \who. Loc: \exo. Extension: \cref{sec:supp_joint}.}
\label{tab:levels}
\footnotesize
\setlength{\tabcolsep}{2pt}
\resizebox{\ifdim\width>\columnwidth\columnwidth\else\width\fi}{!}{%
\begin{tabular}{llcccc}
\toprule
Track & Method & L0 & L1 & L2 & L3 \\
\midrule
\multirow{4}{*}{ID (\ka)} & \ours-ZS & 91.3\ci{2.6} & 61.2\ci{1.2} & 77.2\ci{1.3} & 50.4\ci{1.4} \\
 & Pose-supervised training & \textbf{99.0}\ci{0.8} & 87.8\ci{1.3} & 92.2\ci{0.9} & 63.9\ci{1.4} \\
 & \ours & 98.1\ci{1.6} & 87.5\ci{1.3} & 93.0\ci{0.9} & 71.6\ci{1.3} \\
 & Extension & 98.1\ci{1.3} & \textbf{90.4}\ci{1.4} & \textbf{93.8}\ci{0.9} & \textbf{72.8}\ci{1.4} \\
\midrule
\multirow{4}{*}{Loc (MA@5)} & \ours-ZS & 51.1\ci{4.7} & 26.7\ci{2.2} & 30.0\ci{2.0} & 28.5\ci{0.9} \\
 & Pose-supervised training & \textbf{95.8}\ci{1.4} & \textbf{74.4}\ci{2.2} & 78.6\ci{1.2} & 41.4\ci{1.2} \\
 & \ours & 91.7\ci{2.6} & 72.9\ci{2.4} & 77.5\ci{1.4} & 45.1\ci{1.1} \\
 & Extension & 85.3\ci{4.0} & 74.1\ci{3.0} & \textbf{83.1}\ci{1.4} & \textbf{46.8}\ci{1.3} \\
\bottomrule
\end{tabular}}
\end{table}

\subsection{\Where}
\label{sec:supp_where}

In \where, choosing among proposals (including absence decisions) accounts for about half of the remaining mIoU-P error on present queries, and the proposals themselves for the rest: on the Gold tier, an oracle choice reaches 75.3 unseen mIoU-P versus 53.2 for \ours (\cref{tab:wherefull}). All unseen settings in the \where macro averages are simulated because real sets with automatic masks stay outside those averages. \ours adapts a separate LoRA per set on that set's unlabeled Gold queries (present and absent alike) using SAM~3 proposals as its only candidates; no ground-truth mask, wearer identity, or presence label enters adaptation or readout, and the presence threshold remains the one that the starting model chose on the development list. Adapting instead on each set's full unlabeled test list (478,615 queries regardless of label versus 41,534 Gold) changes no seen or unseen macro average by more than 0.3 points and keeps the margin over pose-supervised training (5.0 [4.6, 5.5] unseen mIoU-P versus 5.2 on the 43,283 Gold queries scored for both).

\Cref{tab:wherebb} evaluates the same proposals with other backbones, each with its own presence threshold chosen on the development list: without training, \vggto leads Pi3X, MapAnything, and Depth Anything~3 on every metric, by 2.8--4.9 unseen mIoU-P with paired intervals that exclude zero. After the same pose-supervised training and before adaptation, \vggto outperforms Pi3X and Depth Anything~3 by 8.1 and 3.3 unseen mIoU-P, while \ours, which also adapts, leads them by 13.4 [12.9, 14.1] and 8.6 [8.1, 9.1] unseen mIoU-P and 6.2 [5.8, 6.6] and 5.0 [4.6, 5.5] unseen \giou.

\begin{table}[!htb]
\centering
\caption{\textbf{\Where with pose-supervised training.} Columns and oracle as in \cref{tab:where}. Improvement: \ours minus pose-supervised training.}
\label{tab:wherefull}
\footnotesize
\setlength{\tabcolsep}{2.2pt}
\resizebox{\ifdim\width>\columnwidth\columnwidth\else\width\fi}{!}{%
\begin{tabular}{lcccccccc}
\toprule
 & \tabhead{4}{Seen} & \tabhead{4}{Unseen} \\
\cmidrule(lr){2-5}\cmidrule(lr){6-9}
Method & \giou & mIoU-P & Pr@0.5 & VA & \giou & mIoU-P & Pr@0.5 & VA \\
\midrule
Always absent & 20.0 & 0.0 & 0.0 & 50.0 & 20.0 & 0.0 & 0.0 & 50.0 \\
\rowcolor{tabourslight}
\ours-ZS & 56.6\ci{0.9} & 50.4\ci{1.2} & 58.0\ci{1.3} & 79.6\ci{1.0} & 46.3\ci{0.7} & 37.3\ci{0.9} & 41.7\ci{1.1} & 72.9\ci{0.8} \\
Pose-supervised training & \textbf{68.2}\ci{1.4} & 69.9\ci{1.5} & 79.9\ci{1.6} & \textbf{85.4}\ci{1.0} & 53.0\ci{0.7} & 47.9\ci{0.9} & 52.6\ci{1.0} & \textbf{76.5}\ci{0.9} \\
\rowcolor{tabours}
\ours & 67.8\ci{1.3} & \textbf{70.9}\ci{1.4} & \textbf{81.0}\ci{1.6} & 83.4\ci{1.0} & \textbf{55.1}\ci{0.7} & \textbf{53.2}\ci{1.0} & \textbf{58.9}\ci{1.2} & 75.9\ci{0.9} \\
\midrule
Improvement & $-$0.4\ci{0.3} & +1.0\ci{0.3} & +1.1\ci{0.5} & $-$2.0\ci{0.5} & +2.1\ci{0.3} & +5.3\ci{0.4} & +6.3\ci{0.5} & $-$0.6\ci{0.5} \\
\midrule
Oracle proposal & \no & 79.9 & \no & \no & \no & 75.3 & \no & \no \\
\bottomrule
\end{tabular}}
\end{table}

\begin{table}[!htb]
\centering
\caption{\textbf{\Where with other backbones} (seen / unseen).  \ours keeps the threshold of its starting model.}
\label{tab:wherebb}
\footnotesize
\setlength{\tabcolsep}{2.2pt}
\resizebox{\ifdim\width>\columnwidth\columnwidth\else\width\fi}{!}{%
\begin{tabular}{lccccc}
\toprule
Method & $\theta^\star$ & \giou & mIoU-P & Pr@0.5 & VA \\
\midrule
\multicolumn{6}{l}{\emph{Without training on our data}} \\
\vggto (\ours-ZS) & $12^\circ$ & 56.6 / 46.3 & 50.4 / 37.3 & 58.0 / 41.7 & 79.6 / 72.9 \\
Pi3X~\cite{wang2025pi3} & $30^\circ$ & 49.3 / 42.0 & 43.4 / 34.4 & 49.6 / 38.2 & 75.1 / 68.3 \\
MapAnything~\cite{keetha2025mapanything} & $20^\circ$ & 48.8 / 40.9 & 43.8 / 34.5 & 50.3 / 38.4 & 73.6 / 69.2 \\
Depth Anything~3~\cite{lin2025da3} & $12^\circ$ & 51.9 / 42.0 & 45.3 / 32.3 & 51.9 / 36.0 & 77.6 / 70.3 \\
\midrule
\multicolumn{6}{l}{\emph{Pose-supervised training with our recipe}} \\
Pi3X & $5^\circ$ & 66.1 / 48.9 & 66.0 / 39.8 & 76.1 / 44.9 & 85.6 / 74.1 \\
Depth Anything~3 & $12^\circ$ & 64.5 / 50.1 & 65.3 / 44.6 & 75.2 / 50.1 & 84.0 / 74.9 \\
\vggto & $8$--$20^\circ$ & 68.2 / 53.0 & 69.9 / 47.9 & 79.9 / 52.6 & 85.4 / 76.5 \\
\rowcolor{tabours}
\ours & $8$--$20^\circ$ & 67.8 / 55.1 & 70.9 / 53.2 & 81.0 / 58.9 & 83.4 / 75.9 \\
\bottomrule
\end{tabular}}
\end{table}

Although test-time adaptation improves mIoU-P, it lowers presence accuracy (VA) on both seen and unseen settings, with both intervals excluding zero (\cref{tab:wherefull}): on unseen settings, it finds more visible wearers (72.1 to 80.1\%) but calls more absent ones present (19.0 to 28.3\%).

\paragraph{Absent wearers with true candidates.}
To separate proposals from readout, we apply the presence rule of \where to the true candidate masks of \who (which scores only visible wearers), using the same model, threshold, and queries as on SAM~3 proposals (\cref{tab:presence}). On unseen settings, true masks increase the share of visible wearers that pose-supervised training finds by 8.3 [6.7, 9.6] points---about a third of those missed with proposals---while accuracy on absent queries changes little ($-$0.8 [$-$2.6, 0.0]), raising VA by 3.7 [2.5, 4.4] points; remaining misses occur because even the true mask's angular offset exceeds the threshold. Scoring absence on true masks would largely repeat what \where measures, so \who keeps prior work's closed candidate set.

\begin{table}[!htb]
\centering
\caption{\textbf{Presence verdicts on true candidate masks and on SAM~3 proposals} (40,267 Gold queries). Same model, threshold, and queries on both sides. T-acc and N-acc: accuracy on present and on absent queries.}
\label{tab:presence}
\footnotesize
\setlength{\tabcolsep}{2.2pt}
\resizebox{\ifdim\width>\columnwidth\columnwidth\else\width\fi}{!}{%
\begin{tabular}{ll*{3}{c}*{3}{c}}
\toprule
 & & \tabhead{3}{Seen} & \tabhead{3}{Unseen} \\
\cmidrule(lr){3-5}\cmidrule(lr){6-8}
Model & Candidates & T-acc & N-acc & VA & T-acc & N-acc & VA \\
\midrule
\rowcolor{tabourslight}
\ours-ZS & SAM~3 proposals & 70.9\ci{1.2} & 87.1\ci{2.0} & 79.0\ci{1.1} & 56.2\ci{1.3} & 88.8\ci{1.3} & 72.5\ci{0.9} \\
 & true masks & 72.6\ci{1.3} & 85.9\ci{1.9} & 79.2\ci{1.2} & 62.9\ci{1.3} & 87.4\ci{1.3} & 75.2\ci{0.9} \\
\rowcolor{tabourslight}
 & difference & +1.7\ci{0.4} & $-$1.2\ci{1.0} & +0.2\ci{0.5} & +6.7\ci{1.3} & $-$1.4\ci{0.6} & +2.7\ci{0.7} \\
\midrule
Pose-supervised training & SAM~3 proposals & 87.8\ci{1.3} & 79.5\ci{1.6} & 83.6\ci{1.1} & 72.2\ci{1.5} & 78.9\ci{1.3} & 75.6\ci{0.9} \\
 & true masks & 93.7\ci{0.9} & 77.2\ci{1.5} & 85.4\ci{0.9} & 80.5\ci{1.7} & 78.1\ci{1.8} & 79.3\ci{1.2} \\
 & difference & +5.9\ci{0.9} & $-$2.3\ci{1.5} & +1.8\ci{0.8} & +8.3\ci{1.4} & $-$0.8\ci{1.3} & +3.7\ci{0.9} \\
\bottomrule
\end{tabular}}
\end{table}

On real sets with automatic masks (\cref{tab:whereg3}), pose-supervised training improves segmentation everywhere except DROID (the three models lie within 0.5 mIoU-P), and test-time adaptation adds most on Ego4D-TF (80.0 to 87.1 mIoU-P). Yet adaptation lowers VA on CoMind (87.1 to 75.7) and CORE4D-Real (67.2 to 62.3) by pulling the predicted epipole of absent queries toward a proposal, cutting the median angular offset of the chosen proposal from 36.7$^\circ$ to 12.2$^\circ$ and from 7.9$^\circ$ to 2.8$^\circ$, so both sets would need a threshold near 1$^\circ$. Adapting on the full test list instead raises their VA by 5.4 and 3.4 points, because the share of absent queries it calls absent rises by 11 and 7 points, which leaves CORE4D-Real only 1.5 [$-$0.5, 3.7] points behind pose-supervised training. DROID segmentation stays well below the proposal ceiling because near-field wrist cameras are difficult for the geometry.

\begin{table*}[t]
\centering
\caption{\textbf{\Where on the sets with automatic masks.} ZS: \ours-ZS. PST: pose-supervised training. Ceiling: the best SAM~3 proposal. Ego-Exo4D and DROID have no absent queries.}
\label{tab:whereg3}
\scriptsize
\setlength{\tabcolsep}{2.2pt}
\resizebox{\ifdim\width>\textwidth\textwidth\else\width\fi}{!}{%
\begin{tabular}{lccccccccccccccc}
\toprule
 & & & & \tabhead{3}{mIoU-P} & \tabhead{3}{Pr@0.5} & \tabhead{3}{\giou} & \tabhead{3}{VA} \\
\cmidrule(lr){5-7}\cmidrule(lr){8-10}\cmidrule(lr){11-13}\cmidrule(lr){14-16}
Set & Level & Queries / present & Ceiling & ZS & PST & \ours & ZS & PST & \ours & ZS & PST & \ours & ZS & PST & \ours \\
\midrule
Ego-Exo4D, seen sites & L0 & 555 / 555 & 95.7 & 77.1 & \textbf{95.7} & 95.5 & 80.5 & \textbf{99.9} & 99.8 & \no & \no & \no & \no & \no & \no \\
Ego-Exo4D, held-out sites & L1 & 534 / 534 & 97.3 & 92.1 & 96.8 & \textbf{97.2} & 94.6 & 99.5 & \textbf{99.9} & \no & \no & \no & \no & \no & \no \\
CoMind & L0 & 709 / 609 & 97.1 & 85.5 & 94.9 & \textbf{95.7} & 88.0 & 97.7 & \textbf{98.6} & 85.0 & \textbf{91.1} & 87.0 & \textbf{87.2} & 87.1 & 75.7 \\
CORE4D-Real & L2 & 722 / 622 & 94.2 & 86.3 & 92.6 & \textbf{93.4} & 91.3 & 97.9 & \textbf{98.8} & 78.2 & \textbf{81.1} & 79.6 & \textbf{70.0} & 67.2 & 62.3 \\
EgoHumans & L2 & 701 / 601 & 96.3 & 75.8 & 95.6 & \textbf{96.0} & 78.7 & 99.3 & \textbf{99.7} & 80.6 & \textbf{95.6} & 92.4 & 92.4 & \textbf{97.6} & 89.0 \\
EgoHumans, undistorted & L2 & 701 / 601 & 91.8 & 87.6 & 90.8 & \textbf{91.0} & 92.8 & 96.3 & \textbf{96.5} & 89.9 & \textbf{91.6} & 89.6 & \textbf{97.1} & 96.4 & 91.4 \\
Ego4D-TF (peer) & L2 & 642 / 529 & 95.6 & 66.9 & 80.0 & \textbf{87.1} & 68.8 & 82.5 & \textbf{90.0} & 71.9 & 81.3 & \textbf{82.9} & 82.3 & \textbf{85.7} & 81.1 \\
DROID & L3 & 139 / 139 & 92.9 & \textbf{64.2} & 63.7 & \textbf{64.2} & \textbf{63.3} & 61.4 & 61.1 & \no & \no & \no & \no & \no & \no \\
\bottomrule
\end{tabular}}
\end{table*}

\subsection{\Exo}
\label{sec:supp_exo}

On seen settings, test-time adaptation costs 1.8 MA@5 points (\cref{sec:supp_tta}); the candidate constraint asks only that the epipole fall on a mask, so tall human masks may allow vertical drift. Among the other zero-shot baselines (\cref{tab:exoall}), none reaches \ours-ZS's MA@5 (28.6 and 28.9), even though the rows marked $^\dagger$ count only the pairs they solve. In the standard pose metrics of VGGT~\cite{wang2025vggt}, pooled over queries (\cref{tab:exopooled}), adaptation adds 3.5 [3.1, 3.9] AUC@30 points on unseen settings while costing 0.7 [0.5, 0.9] on seen ones.

\begin{table}[!htb]
\centering
\caption{\textbf{\Exo for the other zero-shot baselines} (exo image first). Columns as in \cref{tab:exo}. $^\ast$Pose from matches and an essential matrix. $^\dagger$MA@$\epsilon$ and $\aex$ count only pairs with a solution.}
\label{tab:exoall}
\footnotesize
\setlength{\tabcolsep}{2.5pt}
\resizebox{\ifdim\width>\columnwidth\columnwidth\else\width\fi}{!}{%
\begin{tabular}{lcccccc}
\toprule
 & \tabhead{3}{Seen} & \tabhead{3}{Unseen} \\
\cmidrule(lr){2-4}\cmidrule(lr){5-7}
Method & MA@5$\uparrow$ & MA@15$\uparrow$ & $\aex\downarrow$ & MA@5$\uparrow$ & MA@15$\uparrow$ & $\aex\downarrow$ \\
\midrule
DUSt3R~\cite{wang2024dust3r} & 9.5 & 26.1 & 56.5$^\circ$ & 8.5 & 24.9 & 54.7$^\circ$ \\
CUT3R~\cite{wang2025cut3r} & 12.2 & 30.1 & 39.0$^\circ$ & 8.8 & 27.7 & 38.3$^\circ$ \\
RoMa~\cite{edstedt2024roma} + E$^\ast$ & 12.3 & 27.6 & 26.6$^\circ$ & 9.8 & 21.8 & 37.8$^\circ$ \\
SIFT~\cite{lowe2004sift} + E$^{\ast\dagger}$ & 3.0 & 10.0 & 67.1$^\circ$ & 6.0 & 12.1 & 54.9$^\circ$ \\
SuperPoint + LightGlue~\cite{detone2018superpoint,lindenberger2023lightglue} + E$^{\ast\dagger}$ & 16.3 & 37.8 & 20.1$^\circ$ & 14.8 & 30.5 & 29.3$^\circ$ \\
MASt3R matches + E$^{\ast\dagger}$ & 24.1 & 49.6 & 20.2$^\circ$ & 22.5 & 44.3 & 25.2$^\circ$ \\
\bottomrule
\end{tabular}}
\end{table}

\begin{table}[!htb]
\centering
\caption{\textbf{\Exo, pooled pose metrics} (44,224 queries with complete true poses, seen / unseen). $\aeg$ and RE: median errors in degrees. Translation directions keep their sign (no folding).}
\label{tab:exopooled}
\footnotesize
\setlength{\tabcolsep}{2.2pt}
\resizebox{\ifdim\width>\columnwidth\columnwidth\else\width\fi}{!}{%
\begin{tabular}{lccccc}
\toprule
Method & AUC@30 & RRA@30 & RTA@30 & $\aeg$ & RE \\
\midrule
MASt3R & 33.7 / 29.3 & 55.2 / 54.8 & 48.1 / 46.2 & 40.0 / 43.0 & 19.7 / 20.9 \\
VGGT & 32.9 / 31.4 & 63.2 / 65.6 & 58.3 / 55.1 & 20.7 / 23.5 & 13.8 / 12.2 \\
Pi3X & 37.1 / 40.8 & 74.7 / 80.7 & 62.2 / 60.7 & 17.6 / 16.5 & 9.4 / 7.2 \\
MapAnything & 45.2 / 44.5 & 72.5 / 80.6 & 67.5 / 69.4 & 12.0 / 12.7 & 7.6 / 6.4 \\
Depth Anything 3 & 47.2 / 41.4 & 72.4 / 75.7 & 65.1 / 61.7 & 10.4 / 15.5 & 6.4 / 7.0 \\
Reloc3r & 36.4 / 28.9 & 64.8 / 64.9 & 58.7 / 51.8 & 18.8 / 27.8 & 11.6 / 12.8 \\
\rowcolor{tabourslight}
\ours-ZS & 52.6 / 51.6 & 78.8 / 86.6 & 73.0 / 72.5 & 8.2 / 8.8 & 6.1 / 5.4 \\
\midrule
Pose-supervised training & \textbf{78.1} / 68.6 & \textbf{92.0} / 92.1 & 89.5 / 85.1 & \textbf{1.8} / 3.6 & \textbf{1.7} / \textbf{2.7} \\
\rowcolor{tabours}
\ours & 77.5 / \textbf{72.1} & 91.9 / \textbf{92.8} & \textbf{89.6} / \textbf{89.1} & 2.1 / \textbf{3.4} & 1.9 / \textbf{2.7} \\
\bottomrule
\end{tabular}}
\end{table}

\subsection{Locked test sets}
\label{sec:supp_locked}

All queries from TUMTraf-V2X, ALOHA, and MILUV belong to the Gold tier but stay outside the macro averages because they were added after the design was fixed. Like every other set, they are scored with \cref{eq:mix}, which replaced the centroid of the mask's upper third after these sets were first evaluated (\cref{tab:design}); the change moved \ours from 90.8 to 81.1 Acc on TUMTraf-V2X (through the one sequence of overlapping boxes described below), from 93.5 to 93.6 on ALOHA, and from 89.4 to 89.5 on MILUV.

We obtained TUMTraf-V2X~\cite{zimmer2024tumtrafv2x} after pose-supervised training was frozen. Pairing its roadside camera south2 at intersection s110 (exo) with the test vehicle's front camera (ego) gives 294 L3 queries across four 10-second sequences from 3 recordings, with 13.3 candidate vehicles on average (chance 8.5\%); because the vehicle camera's extrinsics are not verified, we report only \who.

\ours reaches 81.1 [54.8, 100.0] Acc (intervals resample the four 10-second sequences); adaptation adds 6.4 [3.5, 8.6] points over pose-supervised training, and the whole method adds 19.5 [9.2, 33.8] over zero-shot. One 89-query sequence largely determines difficulty: unlike masks, bounding boxes overlap, and every wearer's box there overlaps another vehicle's (mean IoU 0.41), so once adaptation puts the epipole in that overlap, both boxes lie within a fraction of a degree of it and depth decides between them. As on other vehicle sets, EgoRate is 0.81--0.83 across the three models.

The static ALOHA recordings~\cite{zhao2023aloha} likewise informed no design decisions and, like TUMTraf-V2X, form an L3 set evaluated on \who only, here because the cameras are uncalibrated. Each query pairs an overhead exo view of two arms with one of the two wrist cameras as the ego view and asks which arm carries the camera (chance 50\%). We segment both arms with SAM~3 and label each left or right by the image edge it touches; all 24 audited queries were correct, and dropping one task where a coffee machine at the left edge was taken for the left arm leaves 540 queries in 13 tasks. The largest mask cannot distinguish a frame's two arms, and EgoRate is about 1 because replacing the wrist image with noise drops accuracy to chance. With intervals over tasks, adaptation changes Acc by $-$0.1 [$-$2.0, 1.4] points and the whole method adds 4.4 [$-$1.2, 9.3] over zero-shot, although with only 13 tasks this gain remains uncertain. Distinguishing two separated arms does not resolve the near-field wrist segmentation problem seen on DROID (\cref{sec:supp_where}).

Before any model saw MILUV~\cite{shalaby2026miluv}, we fixed its protocol, the readout of that time, and the rows to report, and ran every method once. Three identical quadcopters carrying color cameras fly in one indoor arena under motion capture; each query pairs one drone's frame (exo) with another visible drone's (ego), using the two other drones as candidates (chance 50\%). No annotator labels the set: because MILUV publishes no color-camera calibration, we fit one per recording on the arena's AprilTags (median held-out residuals 5--23 pixels at $1280\times720$), project each drone's body into the exo frame from its motion-capture pose, and segment it with SAM~3 inside the projected box; the projected outline replaces the 1.8\% of masks that disagree with it.

This yields 998 queries across 14 recordings. Because each exo frame generates two queries with the same two candidates, any deterministic method ignoring the ego image and candidate order scores exactly 50\% (including the largest mask), making EgoRate about 1 partly by construction. Resampling across recordings, \ours reaches 89.5 [87.0, 91.7] Acc; adaptation adds 7.0 [5.3, 8.5] points, the whole method adds 12.2 [8.7, 15.0] over zero-shot, and by the rule fixed in advance the strongest baseline is Depth Anything~3 at 81.2, which \ours leads by 8.3 [5.0, 11.3] points. With the angular offset alone (no mask depth), \ours scores 89.0 and adaptation still contributes 5.1 [3.6, 6.6] points, whereas the zero-shot model rises to 84.6, level with pose-supervised training (83.9). With one arena and two candidates per query, the set shows that the method transfers to real flying cameras, not that swarms are solved.

\Cref{tab:locked} compares the methods of \cref{tab:who}, together with SIFT and SuperPoint with LightGlue, on the three locked sets. On ALOHA, \ours leads the strongest baseline (\vggto's camera head) by 5.1 [$-$0.9, 11.1] Acc and Depth Anything~3 by 6.8 [3.1, 11.2], whereas SIFT and SuperPoint with LightGlue stay at chance because they solve only 10.6\% and 1.1\% of wrist-to-overhead pairs. ALOHA is also the only locked set that the LLMs answered, and there \ours leads GPT-6.1 Sol, GPT-6 Astra, and Claude Opus by 12.7, 9.4, and 13.5 points (intervals exclude zero).

\begin{table*}[t]
\centering
\caption{\textbf{Locked real sets.} Intervals resample tasks (ALOHA), sequences (TUMTraf-V2X), and recordings (MILUV). $^\ast$Pose only, read with the angular offset alone.}
\label{tab:locked}
\footnotesize
\setlength{\tabcolsep}{3pt}
\begin{tabular}{lcccccc}
\toprule
 & \tabhead{2}{ALOHA} & \tabhead{2}{TUMTraf-V2X} & \tabhead{2}{MILUV} \\
\cmidrule(lr){2-3}\cmidrule(lr){4-5}\cmidrule(lr){6-7}
Method & Acc & \ka & Acc & \ka & Acc & \ka \\
\midrule
Largest mask & 50.0\ci{0.0} & 0.0\ci{0.0} & 26.9\ci{31.9} & 20.1\ci{35.6} & 50.0\ci{0.0} & 0.0\ci{0.0} \\
\midrule
DUSt3R~\cite{wang2024dust3r} & 56.6\ci{6.0} & 13.2\ci{12.1} & 30.8\ci{15.0} & 24.4\ci{14.5} & 69.4\ci{2.9} & 38.8\ci{5.8} \\
MASt3R~\cite{leroy2024mast3r} & 72.9\ci{7.5} & 45.7\ci{14.9} & 22.8\ci{16.4} & 15.6\ci{15.5} & 70.3\ci{3.5} & 40.7\ci{7.0} \\
CUT3R~\cite{wang2025cut3r} & 76.3\ci{6.3} & 52.6\ci{12.6} & 9.9\ci{9.2} & 1.5\ci{6.9} & 76.2\ci{3.2} & 52.5\ci{6.5} \\
VGGT~\cite{wang2025vggt} & 80.8\ci{8.1} & 61.7\ci{16.1} & 21.4\ci{14.7} & 14.1\ci{15.6} & 76.3\ci{3.9} & 52.7\ci{7.8} \\
Pi3X~\cite{wang2025pi3} & 76.8\ci{6.2} & 53.5\ci{12.3} & 36.0\ci{31.4} & 30.1\ci{33.3} & 75.2\ci{4.2} & 50.5\ci{8.4} \\
MapAnything~\cite{keetha2025mapanything} & 75.5\ci{5.7} & 50.9\ci{11.4} & 40.8\ci{27.1} & 35.3\ci{28.0} & 77.9\ci{4.0} & 55.7\ci{8.0} \\
Depth Anything~3~\cite{lin2025da3} & 86.8\ci{6.6} & 73.7\ci{13.3} & 29.6\ci{25.4} & 23.1\ci{26.8} & 81.2\ci{3.8} & 62.3\ci{7.5} \\
Reloc3r~\cite{dong2025reloc3r}$^\ast$ & 73.7\ci{5.9} & 47.4\ci{11.8} & 10.2\ci{8.8} & 1.9\ci{7.8} & 77.0\ci{3.1} & 53.9\ci{6.3} \\
RoMa~\cite{edstedt2024roma} + E$^\ast$ & 54.6\ci{7.1} & 9.3\ci{14.2} & 29.2\ci{27.4} & 22.7\ci{29.1} & 69.5\ci{3.3} & 39.0\ci{6.7} \\
SIFT~\cite{lowe2004sift} + E$^\ast$ & 50.8\ci{1.1} & 1.7\ci{2.1} & 9.9\ci{7.2} & 1.5\ci{4.8} & 56.5\ci{1.8} & 13.0\ci{3.6} \\
SuperPoint + LightGlue~\cite{detone2018superpoint,lindenberger2023lightglue} + E$^\ast$ & 50.2\ci{0.4} & 0.4\ci{0.8} & 7.1\ci{4.0} & $-$1.5\ci{1.9} & 69.5\ci{3.2} & 39.1\ci{6.4} \\
MASt3R matches + E$^\ast$ & 75.9\ci{5.8} & 51.9\ci{11.5} & 27.1\ci{14.6} & 20.3\ci{15.5} & 76.5\ci{4.0} & 52.9\ci{8.1} \\
\vggto, dense tokens + PnP & 61.6\ci{4.4} & 23.1\ci{8.8} & 14.6\ci{11.8} & 6.7\ci{10.0} & 66.7\ci{3.1} & 33.4\ci{6.2} \\
\vggto, camera head & 88.5\ci{5.1} & 77.0\ci{10.3} & 57.1\ci{35.0} & 53.2\ci{37.7} & 77.6\ci{3.3} & 55.1\ci{6.5} \\
\midrule
GPT-6.1 Sol & 80.9\ci{6.1} & 61.9\ci{12.3} & \no & \no & \no & \no \\
GPT-6 Astra & 84.3\ci{8.7} & 68.5\ci{17.5} & \no & \no & \no & \no \\
Claude Opus & 80.2\ci{6.1} & 60.4\ci{12.2} & \no & \no & \no & \no \\
\midrule
\rowcolor{tabourslight}
\ours-ZS & 89.3\ci{4.6} & 78.5\ci{9.3} & 61.6\ci{33.8} & 58.0\ci{36.4} & 77.2\ci{3.7} & 54.5\ci{7.4} \\
Pose-supervised training & \textbf{93.7}\ci{5.3} & \textbf{87.4}\ci{10.5} & 74.7\ci{20.8} & 72.4\ci{22.4} & 82.4\ci{3.4} & 64.9\ci{6.7} \\
\rowcolor{tabours}
\ours & 93.6\ci{6.6} & 87.3\ci{13.2} & \textbf{81.1}\ci{22.6} & \textbf{79.3}\ci{24.5} & \textbf{89.5}\ci{2.4} & \textbf{78.9}\ci{4.7} \\
\bottomrule
\end{tabular}
\end{table*}

\subsection{Real sets one by one}
\label{sec:supp_realsets}

\Cref{tab:realsets} breaks down the real-human and real-vehicle averages by individual set.

\begin{table}[!htb]
\centering
\caption{\textbf{Real sets one by one} (Acc / \ka). Extension: other cameras of the same moment (\cref{sec:supp_joint}).}
\label{tab:realsets}
\footnotesize
\setlength{\tabcolsep}{2.5pt}
\resizebox{\ifdim\width>\columnwidth\columnwidth\else\width\fi}{!}{%
\begin{tabular}{lccc}
\toprule
Set & \ours-ZS & \ours & Extension \\
\midrule
Ego-Exo4D, val & 97.7 / 96.2 & 100.0 / 100.0 & 99.9 / 99.8 \\
TF2023, test & 95.0 / 93.3 & 99.0 / 98.7 & 99.2 / 98.9 \\
Harmony4D & 94.8 / 89.5 & 97.9 / 95.8 & 97.5 / 95.0 \\
EgoHumans & 94.5 / 92.2 & 99.8 / 99.8 & 99.8 / 99.8 \\
CoMind & 95.3 / 90.7 & 99.2 / 98.3 & 99.7 / 99.3 \\
CORE4D-Real & 97.1 / 94.3 & 99.8 / 99.7 & 99.5 / 99.0 \\
Ego4D-TF (peer) & 95.5 / 93.7 & 96.4 / 95.0 & 98.5 / 97.9 \\
V2X-Real, roadside & 98.2 / 97.8 & 92.0 / 90.0 & 97.0 / 96.3 \\
V2X-Real, vehicle to vehicle & 71.2 / 61.8 & 95.5 / 94.0 & 93.8 / 91.7 \\
V2X-Seq, roadside & 76.0 / 73.9 & 91.5 / 90.7 & 91.8 / 91.0 \\
UrbanIng, vehicle to vehicle & 64.5 / 58.4 & 87.5 / 85.3 & 87.2 / 85.0 \\
UrbanIng, thermal & 13.8 / 4.5 & 70.5 / 67.3 & 64.0 / 60.2 \\
\midrule
Real humans, mean & 95.7 / 92.8 & 98.9 / 98.2 & 99.2 / 98.5 \\
Real vehicles, mean & 64.8 / 59.3 & 87.4 / 85.5 & 86.8 / 84.8 \\
\bottomrule
\end{tabular}}
\end{table}

\subsection{Lens versions and thermal images}
\label{sec:supp_lens}

Real-device lenses (v2) reduce zero-shot accuracy---which is why v1 and v2 are evaluated as separate settings---whereas pose-supervised training closes this gap (\cref{tab:lens}). On EgoHumans, undistortion helps the zero-shot angular-offset readout, while pose-supervised training performs better on raw fisheye images. In the paired thermal test bed (\cref{tab:thermal}), pose-supervised training likewise raises RGB-to-thermal accuracy from 26.6 to 94.9.

\begin{table}[!htb]
\centering
\caption{\textbf{Lens versions} (Acc). Zero-shot: Full tier. Mean over 37 pairs of v1 and v2 settings.  PST: pose-supervised training.}
\label{tab:lens}
\footnotesize
\setlength{\tabcolsep}{3pt}
\begin{tabular}{lcc}
\toprule
Comparison & Zero-shot & PST \\
\midrule
v1 $-$ v2, mean & +6.8 & $-$0.5 \\
v1 $-$ v2, football fixed & +25.3 & $-$7.9 \\
EgoHumans, angular offset, raw / undistorted & 85.0 / 91.9 & 97.0 / 94.5 \\
EgoHumans, depth, raw / undistorted & 95.5 / 95.2 & 100.0 / 97.8 \\
\bottomrule
\end{tabular}
\end{table}

\begin{table}[!htb]
\centering
\caption{\textbf{Thermal test bed} (the same 12 test rooms in every row). Zero-shot: Full tier. PST: pose-supervised training.}
\label{tab:thermal}
\footnotesize
\setlength{\tabcolsep}{3pt}
\begin{tabular}{lcccc}
\toprule
 & \tabhead{2}{Acc} & \tabhead{2}{$\aex$} \\
\cmidrule(lr){2-3}\cmidrule(lr){4-5}
Ego $\rightarrow$ exo & Zero-shot & PST & Zero-shot & PST \\
\midrule
Thermal $\rightarrow$ thermal & 60.8 & 89.1 & 5.9$^\circ$ & 1.8$^\circ$ \\
RGB $\rightarrow$ thermal & 26.6 & 94.9 & 129.1$^\circ$ & 1.7$^\circ$ \\
RGB $\rightarrow$ RGB & 94.5 & 99.8 & 4.3$^\circ$ & 1.3$^\circ$ \\
\bottomrule
\end{tabular}
\end{table}

\section{Comparisons on Prior Benchmarks}
\label{sec:supp_baselines}

This section compares our method with prior work under each benchmark's protocol (or our closest approximation for MAF) and with multimodal LLMs on our wearer identification queries.

\subsection{TF2023 under PEN's protocol}
\label{sec:supp_pen}

Because PEN reports on original TF2023 labels, \cref{tab:pen} compares against PEN on those labels and uses corrected labels for its other columns (\cref{sec:supp_data}), while \cref{tab:penall} evaluates the methods of \cref{tab:who} other than the largest mask, plus SIFT and SuperPoint with LightGlue, on both label sets under that protocol. Paired over the 12 test scenes (0.9 pool, original labels), \ours-ZS differs from Depth Anything~3 by +1.2 [$-$1.2, 3.3] points, whereas \ours leads Depth Anything~3 by 5.0 [2.8, 7.8]. SIFT and SuperPoint with LightGlue recover poses for only 21.7 and 63.9\% of pairs, leaving the rest tied.

\begin{table}[t]
\centering
\caption{\textbf{All methods under PEN's protocol}. Pools as in \cref{tab:pen}. Acc on the original / corrected labels, mIoU on the original ones. $^\ast$Pose only, read with the angular offset alone.}
\label{tab:penall}
\footnotesize
\setlength{\tabcolsep}{2.5pt}
\resizebox{\ifdim\width>\columnwidth\columnwidth\else\width\fi}{!}{%
\begin{tabular}{lcccc}
\toprule
Method & Acc, 0.9 pool & mIoU & Acc, 0.5 pool & mIoU \\
\midrule
PEN~\cite{zhao2024pen}, reported & 52.9 / \no & 44 & \no & \no \\
\midrule
\multicolumn{5}{l}{\emph{Without training on our data, exo image first}} \\
DUSt3R~\cite{wang2024dust3r} & 63.9 / 66.7 & 54.8 & 67.7 / 70.6 & 57.9 \\
MASt3R~\cite{leroy2024mast3r} & 76.6 / 80.3 & 65.7 & 81.9 / 85.7 & 70.1 \\
CUT3R~\cite{wang2025cut3r} & 68.5 / 71.8 & 58.8 & 73.3 / 76.7 & 62.7 \\
VGGT~\cite{wang2025vggt} & 71.5 / 74.1 & 61.3 & 76.7 / 79.3 & 65.5 \\
Pi3X~\cite{wang2025pi3} & 76.5 / 80.2 & 65.5 & 82.1 / 85.9 & 70.1 \\
MapAnything~\cite{keetha2025mapanything} & 71.6 / 75.2 & 61.4 & 76.7 / 80.3 & 65.5 \\
Depth Anything~3~\cite{lin2025da3} & 76.9 / 80.6 & 66.0 & 82.2 / 86.0 & 70.4 \\
Reloc3r~\cite{dong2025reloc3r}$^\ast$ & 57.5 / 60.5 & 49.3 & 60.3 / 63.3 & 51.5 \\
RoMa~\cite{edstedt2024roma} + E$^\ast$ & 68.0 / 67.5 & 58.1 & 71.4 / 71.3 & 60.9 \\
SIFT~\cite{lowe2004sift} + E$^\ast$ & 27.4 / 27.3 & 23.6 & 27.0 / 26.9 & 23.3 \\
SuperPoint + LightGlue~\cite{detone2018superpoint,lindenberger2023lightglue} + E$^\ast$ & 49.9 / 49.7 & 43.0 & 51.8 / 51.7 & 44.6 \\
MASt3R matches + E$^\ast$ & 75.1 / 78.7 & 64.5 & 79.9 / 83.5 & 68.6 \\
\vggto, dense tokens + PnP & 59.3 / 60.6 & 51.0 & 62.7 / 64.1 & 53.8 \\
\vggto, camera head & 78.0 / 81.7 & 67.0 & 84.0 / 87.8 & 71.9 \\
\midrule
\rowcolor{tabourslight}
\ours-ZS & 78.0 / 81.8 & 67.0 & 84.0 / 87.8 & 71.9 \\
TF2023 wearer labels only & 81.7 / 85.4 & 70.0 & 87.8 / 91.6 & 74.9 \\
Pose-supervised training & 81.4 / 85.2 & 69.7 & 87.8 / 91.6 & 74.9 \\
\rowcolor{tabours}
\ours$^\ddagger$, adapted on the 0.9 pool & 81.8 / 85.6 & 70.1 & 88.0 / 91.8 & 75.2 \\
\ours$^\ddagger$, adapted on the 0.5 pool & 81.9 / 85.6 & 70.1 & 87.8 / 91.6 & 75.1 \\
\quad adapted on the dataset's masks & 82.1 / 85.8 & 70.3 & 88.3 / 92.1 & 75.4 \\
\midrule
Ceiling & 86.0 & 73.6 & 92.8 & 79.1 \\
\bottomrule
\end{tabular}}
\end{table}

\subsection{Sequences and video benchmarks}
\label{sec:supp_video}

\paragraph{MAF's protocol.}
MAF~\cite{zhao2026sequence} identifies the wearer from 30-frame sequences, built from PEN's official TF2023 test split and scored with one pick per sequence. To approximate this without temporal training, for each official test frame we take the same camera pair's official frames within the 30 video frames ending at it, score each frame pair on its own with \cref{eq:mix}, divide each candidate's score in a frame by the median over candidates, and sum these normalized scores over the sequence (\cref{tab:video}), with no tuning. MAF and PEN train on TF2023, whereas our trained models learn from our training sets, which include TF2023's; \ours-ZS is never trained on TF2023, and \ours also adapts to the test frames without labels.

\begin{table}[!htb]
\centering
\caption{\textbf{An approximation of MAF's sequence protocol on TF2023} (Acc). Candidates: TF2023's ground-truth masks, as in MAF's protocol.}
\label{tab:video}
\footnotesize
\setlength{\tabcolsep}{4pt}
\resizebox{\ifdim\width>\columnwidth\columnwidth\else\width\fi}{!}{%
\begin{tabular}{llcc}
\toprule
Method & Frames & Original & Corrected \\
\midrule
PEN~\cite{zhao2024pen}, scores summed (reported by MAF~\cite{zhao2026sequence}) & 30 & 67.2 & \no \\
MAF~\cite{zhao2026sequence} (reported) & 30 & 90.3 & \no \\
\midrule
\rowcolor{tabourslight}
\ours-ZS & 1 & 91.0 & 94.8 \\
 & $\leq$30 & 95.0 & 98.9 \\
\rowcolor{tabours}
\ours & 1 & 95.6 & 99.5 \\
 & $\leq$30 & \textbf{96.0} & \textbf{99.9} \\
\midrule
Ceiling & & 96.1 & 100.0 \\
\bottomrule
\end{tabular}}
\end{table}

Temporal fusion adds 4.0 [1.1, 8.2] points zero-shot and 0.4 [0.2, 0.6] for \ours, placing it within 0.1 points of the ceiling of the original labels. That ceiling exists because the original labels either name the wrong human or leave the wearer unannotated in 842 queries (3.9\%), mostly through the identity swap of scene 27 (\cref{sec:supp_data}), whereas the corrected labels omit the 45 sequences whose wearer is unannotated. Comparing with MAF is inexact: we have only official frames (averaging 22.2 per 30-frame window), keep TF2023's timing where MAF re-synchronized videos by hand, and do not know how MAF built sequences. Across four window constructions and two candidate sets, \ours reaches 95.9--96.1\% on the original labels and 99.7--100.0\% on the corrected ones; MAF's TF2025 results cannot be reproduced because TF2025 is unreleased.

\paragraph{CVMHAT with DMHA's metric.}
DMHA~\cite{han2022dmha} ranks the humans in a top view by how likely each is to wear the side-view camera and reports the AUC of the resulting top-$k$ accuracy curve, reaching 79.26 on real CVMHAT videos~\cite{han2020cvmhat}. CVMHAT is an unseen domain (L2) for our models and is not named in the training data of any of the other models. Across all 15 pairs at every tenth frame (1,170 frames), \ours-ZS, pose-supervised training, and \ours rank the wearer first in 99.0\%, 100.0\%, and 100.0\% of frames. All three achieve an AUC at or within 0.3 points of the perfect-ranking AUC, and both trained models reach it on any seven-pair test split. Our candidates, however, are ground-truth boxes, in effect a perfect detector, whereas DMHA ranks its own detections; the benchmark is thus saturated for \ours and does not measure the gap between the methods.

\paragraph{Our real video sets.}
We fuse each query with the test queries of the same camera pair within $\pm$16 sampled frames, matching candidates by ID; we chose per-frame fusion over joint input without multi-frame training after comparing them on the six real human test sets other than Ego4D-TF. After pose-supervised training, this raises macro Acc on those six sets from 99.4 to 99.9 (+0.6 [0.3, 0.8]), while \ours reaches 100.0.

On Ego4D-TF, where sampled frames are 20\,s apart, fusion adds +0.4 [0.0, 1.1] after pose-supervised training and +1.7 [0.7, 2.6] for \ours (raising it from 96.4 to 98.1). Across all seven sets, \ours improves from 98.9 to 99.7 (+0.8 [0.5, 1.2]).
We could not obtain the other video benchmarks~\cite{fan2017identifying,xu2018joint,yonetani2015egosurfing,wen2021seeing} from their documented links.

\subsection{BEVSee on CSRD-II}
\label{sec:supp_bevsee}
We also compare with BEVSee~\cite{bevsee2024} (cited in Related Work as closest to our approach), which locates cameras from the layout of humans visible to several cameras. Under its protocol---the first 1,000 frames of hor1 and hor3 in virtual CSRD-II, which none of our training sets contains---pose-supervised training reaches 1.39$^\circ$ mean orientation error (median 1.07$^\circ$) and 99.5\% of cameras within $5^\circ$, versus 5.78$^\circ$ and 59.5\% for BEVSee and 6.79$^\circ$ and 86.6\% for zero-shot \vggto. Because camera positions need a metric scale, which one image pair does not provide, combining our predicted direction with the true distance places 97.8\% of cameras within 0.5\,m after pose-supervised training, versus 42.2\% for BEVSee; this comparison favors us, since BEVSee estimates the distance as well.

\subsection{Multimodal LLMs}
\label{sec:supp_llm}

In the matched-item columns of \cref{tab:llm}, \ours outperforms the proprietary LLMs on the Gold items each model answered; runtimes are median end-to-end seconds per item (the GPT models via the OpenAI API, Claude Opus via its command-line client, open-weight models via vLLM). The 32B and 38B open-weight models score at chance overall except on Ego-Exo4D---where the largest candidate is usually the wearer (\cref{sec:supp_protocol})---reaching 79.2 and 81.7\% Acc versus 100.0\% for \ours. Per domain on each model's items, only on robot arms does any of the three proprietary LLMs lead \ours (likely because a wrist camera sees the gripper and the object); on real humans they reach 90.3--92.5 Acc versus 96.7--98.2 for \ours, whereas on real ALOHA arms \ours leads all three by 9.4--13.5 points (\cref{tab:locked}). CoMind is excluded because multimodal-LLM votes labeled its wearers.

Mirroring the exo image preserves both images' content while breaking their geometry. Across the 48 sets with masks, this lowers \ours's seen \ka by 65.6 points, to 21.9, whereas on the 410 items that all three proprietary LLMs answered, mirroring hurts the LLMs far less than \ours (\cref{tab:flip}).

\begin{table}[t]
\centering
\caption{\textbf{Mirrored exo images} (\ka, the 410 Gold items with masks that all three proprietary LLMs answered).}
\label{tab:flip}
\footnotesize
\setlength{\tabcolsep}{4pt}
\begin{tabular}{lccc}
\toprule
Model & Original & Mirrored & Difference \\
\midrule
Claude Opus & 67.7 & 32.2 & $-$35.5 [$-$42.2, $-$28.3] \\
GPT-6.1 Sol & 49.7 & 41.3 & \textbf{$-$8.5} [$-$13.8, $-$3.0] \\
GPT-6 Astra & 56.0 & \textbf{45.7} & $-$10.3 [$-$15.5, $-$5.2] \\
\rowcolor{tabours}
\ours & \textbf{83.2} & 12.3 & $-$70.8 [$-$76.0, $-$65.7] \\
\bottomrule
\end{tabular}
\end{table}

For the LLM evaluation, items are drawn from Gold by a hash of the query key, up to 300 per setting. Each item shows the exo image, with every candidate boxed and numbered in random order, and the ego image, both resized to a long side of 1280\,px, and requests one number in a JSON object: ``Both photos were taken at the same moment in the same place. Image 1 is a third-person view. Image 2 was taken by a camera carried by one of the [wearers] in Image 1. In Image 1 every candidate is marked with a coloured box and a number. Which numbered candidate carries the camera that took Image 2? Answer from the images only.'' We queried the models in November 2026 in a single turn at high reasoning effort where offered; unparseable answers count as wrong. Open-weight models ran in vLLM~\cite{kwon2023vllm} with their own sampling parameters: temperature 0.7, top-p 0.8, and top-k 20 for Qwen3-VL~\cite{bai2025qwen3vl}, while InternVL3.5~\cite{wang2025internvl35} sets none and samples at temperature 1.

\section{Training and Alternative Models}
\label{sec:supp_ablation}
\label{sec:supp_more}

This section examines which training data matter, whether our recipe transfers to other backbones and multiple ego cameras, and whether appearance models can replace geometry.

\subsection{Training data}
\label{sec:supp_ftdata}

\Cref{tab:train} ablates training data and supervision for pose-supervised training. Under PEN's protocol on corrected labels (\cref{tab:penall}), a model trained on all our domains matches one trained on TF2023 wearer labels alone, yet the latter---whose training data come from fixed exo cameras that often see occluded wearers---generalizes poorly to unseen settings (\cref{tab:train}).

Under the main recipe, simulated-only training shifts macro averages by at most 0.8 \ka (\cref{tab:train}); on real sets (paired scene-level confidence intervals), it changes Acc by $-$2.0 [$-$2.6, $-$1.3] on seen humans and $-$0.3 [$-$0.8, 0.0] on humans from unseen sources while gaining +4.0 [1.6, 6.5] on vehicles. Scenes from rendering engines never used in training---Unreal, Isaac Sim, and CARLA~\cite{dosovitskiy2017carla}---gain 18.2 \ka on L3, versus 12.9 for Blender scenes, although engine and wearer type are confounded because no L3 wearer type appears in two engines. In the one matched comparison (humans at L2), Unreal sets gain 17.8 points versus 23.0 on the Blender sports sets, with Unreal starting higher (70.4 versus 59.7) and ending higher (88.2 versus 82.7).

\begin{table}[t]
\centering
\caption{\textbf{Pose-supervised training.} Each indented row changes the recipe above it as named.}
\label{tab:train}
\footnotesize
\setlength{\tabcolsep}{2.5pt}
\begin{tabular}{@{}>{\raggedright\arraybackslash\hangindent=1em\hangafter=1}p{0.42\columnwidth}cccc@{}}
\toprule
 & \tabhead{2}{\ka} & \tabhead{2}{MA@5} \\
\cmidrule(lr){2-3}\cmidrule(lr){4-5}
Training data & Seen & Unseen & Seen & Unseen \\
\midrule
\rowcolor{tabourslight}
\ours-ZS, no training & 60.7\ci{1.2} & 58.6\ci{1.1} & 28.6\ci{1.6} & 28.9\ci{0.9} \\
\midrule
\multicolumn{5}{@{}l}{\emph{Main recipe: rank 64, color jitter}} \\
Main recipe & 87.8\ci{1.3} & 72.6\ci{1.0} & 74.5\ci{2.2} & 50.7\ci{0.9} \\
\quad pose labels only, no wearer label & 85.3\ci{1.2} & 71.1\ci{1.0} & \textbf{78.4}\ci{1.8} & \textbf{55.3}\ci{0.9} \\
\quad simulated data only & \textbf{87.9}\ci{1.2} & \textbf{73.4}\ci{1.1} & 69.6\ci{2.4} & 52.7\ci{0.9} \\
\quad TF2023 wearer labels only, a shorter schedule & 64.9\ci{1.6} & 57.4\ci{1.0} & 5.6\ci{0.6} & 5.9\ci{0.4} \\
\midrule
\multicolumn{5}{@{}l}{\emph{Rank-16 recipe with a shorter schedule}} \\
Real and simulated data & 81.4\ci{1.9} & 72.1\ci{1.0} & 68.6\ci{2.2} & 50.7\ci{0.9} \\
\quad real data only & 59.4\ci{1.4} & 56.2\ci{1.2} & 39.2\ci{1.0} & 27.5\ci{0.8} \\
\quad simulated data only & 83.3\ci{1.4} & 72.1\ci{1.1} & 65.8\ci{2.4} & 51.7\ci{1.0} \\
\bottomrule
\end{tabular}
\end{table}

\subsection{Other backbones}
\label{sec:supp_bb}

\Cref{tab:bbft} applies our pose-supervised training recipe to VGGT, Pi3X, and Depth Anything~3; evaluated with both input orders, VGGT trained the same way reaches 83.4 seen and 64.7 unseen \ka.

\begin{table}[t]
\centering
\caption{\textbf{Pose-supervised training on other backbones} (exo image first). As for the other backbones, the \vggto rows do not reflect ego cameras predicted behind the exo camera, unlike \cref{tab:components} (\cref{sec:supp_impl}).}
\label{tab:bbft}
\footnotesize
\setlength{\tabcolsep}{2.5pt}
\resizebox{\ifdim\width>\columnwidth\columnwidth\else\width\fi}{!}{%
\begin{tabular}{lcccc}
\toprule
 & \tabhead{2}{\ka} & \tabhead{2}{MA@5} \\
\cmidrule(lr){2-3}\cmidrule(lr){4-5}
Backbone & Seen & Unseen & Seen & Unseen \\
\midrule
VGGT~\cite{wang2025vggt} & 38.5\ci{1.2} & 35.7\ci{1.2} & 11.9\ci{0.9} & 13.5\ci{0.8} \\
\quad pose-supervised training & 82.2\ci{1.7} & 63.2\ci{1.1} & 66.2\ci{2.4} & 41.1\ci{0.8} \\
\quad + adaptation$^\ddagger$ & 82.4\ci{1.6} & 69.1\ci{1.1} & 63.5\ci{2.6} & 42.8\ci{0.8} \\
Pi3X~\cite{wang2025pi3} & 48.6\ci{1.2} & 50.0\ci{1.1} & 14.2\ci{0.8} & 22.6\ci{0.7} \\
\quad pose-supervised training & 84.7\ci{1.5} & 66.3\ci{1.0} & 68.3\ci{2.5} & 41.5\ci{0.8} \\
\quad + adaptation$^\ddagger$ & 84.1\ci{1.6} & 72.3\ci{1.1} & 65.3\ci{2.5} & 44.9\ci{0.9} \\
Depth Anything~3~\cite{lin2025da3} & 50.7\ci{1.6} & 48.3\ci{1.3} & 29.9\ci{1.5} & 24.0\ci{0.8} \\
\quad pose-supervised training & 79.3\ci{1.5} & 65.0\ci{1.1} & 64.2\ci{2.4} & 46.6\ci{0.8} \\
\quad + adaptation$^\ddagger$ & 79.6\ci{1.5} & 70.0\ci{1.0} & 61.7\ci{2.5} & 49.1\ci{0.9} \\
\vggto & 61.0\ci{1.2} & 58.2\ci{1.1} & 28.9\ci{1.8} & 28.8\ci{0.9} \\
\quad pose-supervised training & 87.1\ci{1.4} & 71.6\ci{1.0} & 73.9\ci{2.2} & 50.2\ci{0.9} \\
\quad + adaptation$^\ddagger$ & 86.9\ci{1.3} & 77.4\ci{0.9} & 72.1\ci{2.2} & 52.6\ci{0.9} \\
\midrule
\multicolumn{5}{l}{\emph{Paired: \vggto minus backbone, both trained the same way}} \\
VGGT & +4.9\ci{1.1} & +8.5\ci{0.9} & +7.7\ci{0.8} & +9.2\ci{0.7} \\
Pi3X & +2.3\ci{0.8} & +5.3\ci{0.8} & +5.7\ci{0.9} & +8.8\ci{0.7} \\
Depth Anything~3 & +7.8\ci{0.8} & +6.7\ci{0.9} & +9.7\ci{1.0} & +3.6\ci{0.7} \\
\midrule
\multicolumn{5}{l}{\emph{Paired: gain from adaptation}} \\
VGGT & +0.2\ci{0.6} & +5.9\ci{0.8} & $-$2.7\ci{0.6} & +1.7\ci{0.5} \\
Pi3X & $-$0.7\ci{0.5} & +5.9\ci{0.8} & $-$3.0\ci{1.0} & +3.4\ci{0.5} \\
Depth Anything~3 & +0.3\ci{0.5} & +5.1\ci{0.6} & $-$2.5\ci{1.2} & +2.6\ci{0.4} \\
\vggto & $-$0.2\ci{0.5} & +5.7\ci{0.5} & $-$1.9\ci{0.5} & +2.4\ci{0.4} \\
\bottomrule
\end{tabular}}
\end{table}

\subsection{Other cameras at the same moment}
\label{sec:supp_joint}

When the exo and ego images share little of the scene---as in many underwater pairs (\cref{tab:env})---the pose has few direct cues, yet other ego images from the same frame often overlap with both. To exploit this, our extension feeds the exo image together with the ego images of the other wearers in its frame as one input (once with the exo image first and once in reverse order, with it last) and scores each (exo, ego) pair as before. Because a model trained on pairs produces poor mask depth under joint input, we continue training on one exo image and several ego images from its frame, with a loss per pair (multi-view training), and adapt on joint input. Continuing pair training on as many pairs as multi-view training sees (108K) leaves pair results unchanged (88.1 seen and 72.5 unseen \ka) and does not repair mask depth under joint input (80.3 versus 90.2 Acc on the development list), confirming that seeing several ego images together---not more pairs---makes joint input work.

We chose the number of ego images per pass on eight bird, drone, and underwater settings and used 800 adaptation steps from \vggto (chosen on all Gold sets). \Cref{tab:fullci,tab:joint,tab:views} report the extension results and the sweep over ego images per pass: only after multi-view training does joint input help on seen settings, both with and without adaptation (+2.2 [1.5, 2.9] seen \ka with adaptation, \cref{tab:joint}), whereas robot arms lose 18.3 points with the extension.

\begin{table}[!htb]
\centering
\caption{\textbf{The extension with other cameras} (\ka, the 91 settings with joint input, an earlier adaptation configuration). Bottom: Shapley values~\cite{shapley1953value}.}
\label{tab:joint}
\footnotesize
\setlength{\tabcolsep}{2pt}
\begin{tabular}{llcc}
\toprule
Start & Input & Seen & Unseen \\
\midrule
\multirow{4}{*}{\vggto, no training} & pair & 60.9 & 57.0 \\
 & pair + adaptation & 73.5 & 67.6 \\
 & joint & 61.9 & 56.8 \\
 & joint + adaptation & 71.5 & 66.6 \\
\midrule
\multirow{4}{*}{Pair training} & pair & 87.8 & 70.0 \\
 & pair + adaptation & 87.6 & 75.6 \\
 & joint & 82.7 & 66.1 \\
 & joint + adaptation & 88.6 & 75.6 \\
\midrule
\multirow{4}{*}{Multi-view training} & pair & 87.6 & 70.4 \\
 & pair + adaptation & 88.1 & 76.2 \\
 & joint & 88.7 & 69.8 \\
 & joint + adaptation & \textbf{90.3} & \textbf{76.9} \\
\midrule
\multicolumn{2}{l}{\emph{Shapley value, multi-view model}} & & \\
 & Pose-supervised training & +22.1 & +11.5 \\
 & Joint input & +0.9 & $-$0.1 \\
 & Adaptation & +6.4 & +8.5 \\
\bottomrule
\end{tabular}
\end{table}

\begin{table}[!htb]
\centering
\caption{\textbf{Ego images per pass} (Acc, eight bird, drone, and underwater settings).}
\label{tab:views}
\footnotesize
\setlength{\tabcolsep}{5pt}
\begin{tabular}{lccccc}
\toprule
Ego images in the pass & 1 & 2 & 3 & 4 & 8 \\
\midrule
Acc & 62.8 & 64.2 & 66.2 & 66.6 & 68.9 \\
\bottomrule
\end{tabular}
\end{table}

\subsection{Appearance models}
\label{sec:supp_train}

Our first non-geometric baseline, the appearance head (\cref{tab:nogeo}), pools frozen DINOv3 ViT-L/16 exo features inside each candidate mask and feeds them with the ego class token and patch mean to an MLP over the candidate feature, ego feature, and their elementwise product, trained with cross-entropy over candidates on 148K training queries. It stays near chance on vehicles, robot arms, birds, drones, and top views.

To test whether a stronger 2D architecture trained on more of our data makes geometry unnecessary, the end-to-end model encodes both views with DINOv3 ViT-L/16 (rank-16 LoRA), applies four cross-view Transformer blocks attending within and between views, and scores each candidate with an MLP on mask-pooled exo tokens and pooled patch position, seeing no camera parameters, epipole, or pose. We train it with the loss of the appearance head on 640K samples drawn from 409K Track A training queries with the domain weights of pose-supervised training; its \ka on held-out training scenes saturates (70.6 after 480K samples and 71.2 after 640K). On the Gold tier, it does well where wearers resemble those of training---2.0 [0.5, 3.8] points above \ours-ZS on seen settings---but stays near chance on unseen wearer types, trailing \ours by 24.6 [23.0, 26.1] seen and 61.0 [59.9, 62.1] unseen points. On the Full tier (\cref{sec:exp_more}), it reaches 62.7 seen and 17.0 unseen \ka.

The learned head tests whether a classifier trained directly on the same backbone can replace our geometric readout: we remove \vggto's camera and dense heads and jointly train a new LoRA (configured as in pose-supervised training) with an MLP head that scores each candidate from the exo tokens in its mask, the ego camera token, and the ego patch mean, averaged over both input orders. Using the pose-supervised training recipe and compute budget with cross-entropy alone, the learned head sees 1.21 times as many wearer labels and exceeds pose-supervised training by 1.2 [0.5, 2.0] unseen \ka (73.9), showing that before adaptation such a classifier performs as well as the readout. After adaptation, however, \ours leads on birds, drones, robot arms, and vehicles, while the head stays competitive or ahead elsewhere.

To adapt the head without labels, we minimize the entropy of its softmax over the candidates~\cite{wang2021tent}, training the head and a per-set LoRA for 200 steps on the same queries and at the same learning rate as \ours (no other configuration was tried); the picks become confident (mean entropy drops from 0.84 to 0.15 on birds and from 0.60 to 0.14 on drones) but hardly change. Across all sets, entropy minimization gains the head only 0.5 [0.1, 1.0] unseen points, against 5.5 [4.9, 6.2] for \ours, leaving \ours ahead of the adapted head by 3.8 [2.9, 4.5] unseen points. Whereas our candidate constraint pulls \ours's continuous epipole onto a mask---letting a wrong direction move to the right wearer---a candidate softmax already satisfies that constraint by construction, so entropy minimization mostly sharpens existing picks.

\paragraph{Look-alike candidates.}
\label{sec:supp_lookalike}
To measure sensitivity to visually similar candidates, we define appearance similarity $a$ per Gold query as the cosine similarity between frozen DINOv3 features pooled in the wearer's mask and its most similar other candidate's mask. We regress correctness on $a$ and the logarithms of wearer area, angular gap to the nearest other candidate, and candidate count, with domain fixed effects (36,127 queries with an angular gap, excluding Ego4D-TF and TF2023). Per 0.1 increase in $a$, the appearance head (\cref{tab:nogeo}) loses 4.7 [4.2, 5.1] Acc points, the end-to-end model 3.8 [3.4, 4.3], the three proprietary LLMs 2.8--3.2 on their answered items, and \ours $-$0.2 [$-$0.5, 0.1]. When the angular gap is left out of the regression (all 38,700 queries), \ours loses 1.5 points---consistent with look-alike candidates also standing close together---while the two appearance models barely change (4.3 and 3.8).

\begin{table}[t]
\centering
\caption{\textbf{Without the geometric readout} (\ka).}
\label{tab:nogeo}
\footnotesize
\setlength{\tabcolsep}{3pt}
\begin{tabular}{lcc}
\toprule
Method & Seen & Unseen \\
\midrule
Largest mask & 6.5\ci{1.1} & 1.2\ci{0.8} \\
Appearance head, exo image only & 9.3\ci{0.8} & 0.9\ci{0.8} \\
Appearance head & 39.3\ci{2.0} & 7.4\ci{0.8} \\
\quad + box and size & 39.6\ci{2.2} & 8.0\ci{1.0} \\
End to end, LoRA and cross-view & 62.8\ci{1.9} & 17.1\ci{0.9} \\
\midrule
\rowcolor{tabourslight}
\ours-ZS & 60.7\ci{1.2} & 58.6\ci{1.1} \\
Learned head on \vggto & 88.8\ci{1.1} & 73.9\ci{0.9} \\
\quad + entropy minimization$^\ddagger$ & 88.3\ci{1.4} & 74.4\ci{1.0} \\
Pose-supervised training & 87.8\ci{1.3} & 72.6\ci{1.0} \\
\rowcolor{tabours}
\ours$^\ddagger$ & 87.4\ci{1.4} & 78.2\ci{0.9} \\
\bottomrule
\end{tabular}
\end{table}

\section{Qualitative Results}
\label{sec:supp_qual}
\begin{figure}[!htb]
\centering
\includegraphics[width=\columnwidth]{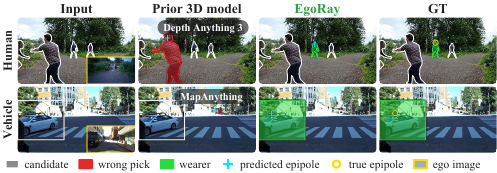}
\caption{\textbf{\ours versus the strongest competing zero-shot 3D model} of each example's set. Examples follow a fixed rule described in the text.}
\label{fig:compare}
\end{figure}

\Cref{fig:compare} compares \ours, on Gold queries with three to twelve candidates, with the most accurate of Pi3X, MapAnything, and Depth Anything~3 on each set (all three read with our score, exo image first); each example is the first query in hash order that this model gets wrong and \ours gets right. The human example comes from EgoHumans, where Depth Anything~3 identifies the wearer in 96.1\% of these queries and \ours in all of them. The vehicle example comes from the three real vehicle sets where \ours outperforms all three models (V2X-Real and UrbanIng between vehicles, V2X-Seq roadside), restricted to queries whose wearer box spans at least a tenth of the image height and whose candidate boxes do not overlap, so that the boxes stay legible; on its set (V2X-Real between vehicles), MapAnything identifies the wearer in 66.4\% of qualifying queries versus 95.4\% for \ours.

\begin{figure*}[p]
\centering
\qtile{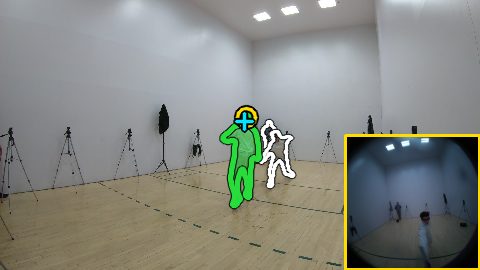}{Real humans}\hfill\qtile{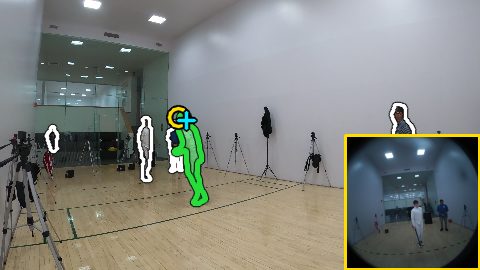}{}\hfill\qtile{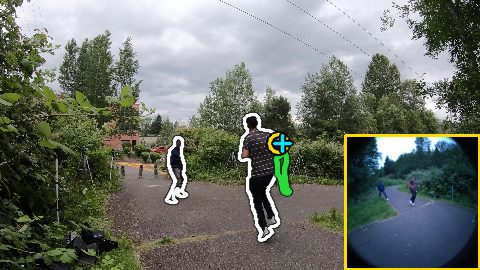}{}\\[2pt]
\qtile{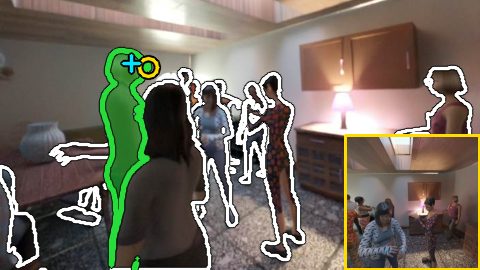}{Indoor}\hfill\qtile{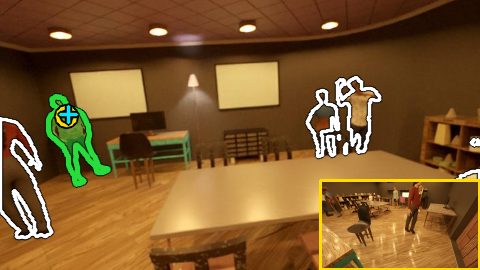}{}\hfill\qtile{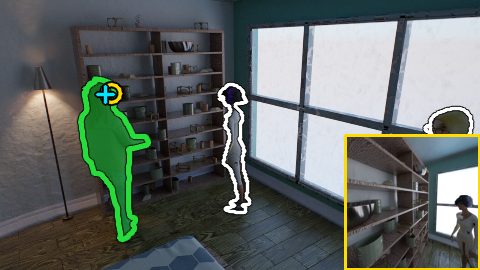}{}\\[2pt]
\qtile{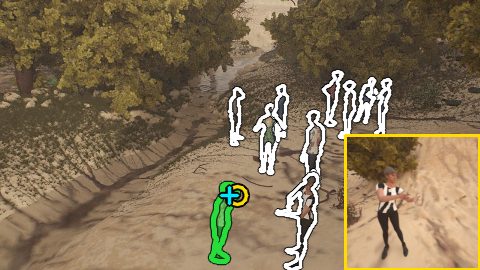}{Nature}\hfill\qtile{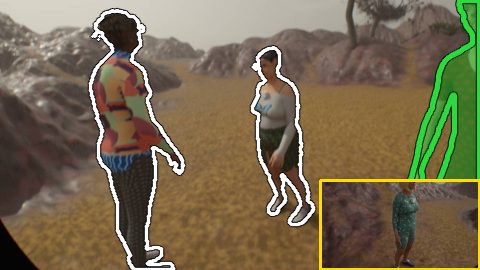}{}\hfill\qtile{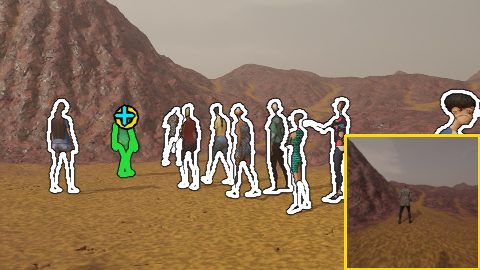}{}\\[2pt]
\qtile{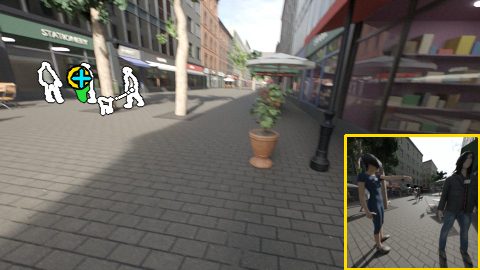}{City}\hfill\qtile{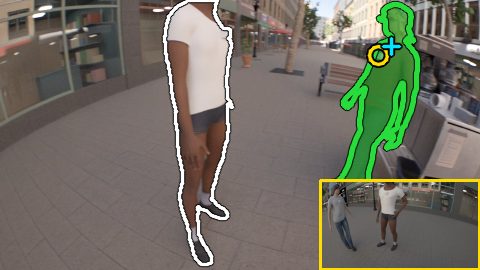}{}\hfill\qtile{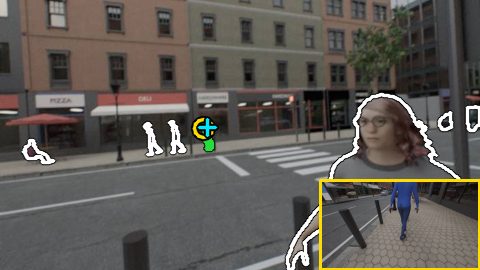}{}\\[2pt]
\qtile{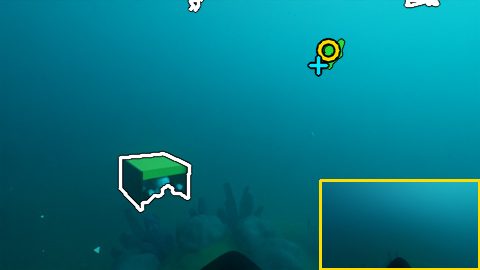}{Underwater}\hfill\qtile{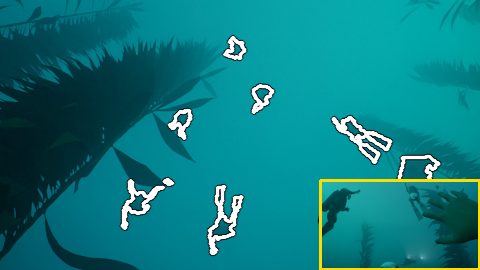}{}\hfill\qtile{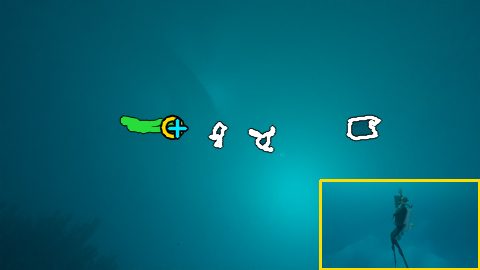}{}\\[2pt]
\caption{\textbf{Correct answers in every domain} (\ours). Real: the humans~\cite{khirodkar2023egohumans}, top views 1 and 3~\cite{han2020cvmhat}, the vehicles~\cite{xiang2024v2xreal,yu2023v2xseq}, and the third robot arm~\cite{khazatsky2024droid}. The rest are simulated. Green: the wearer. White: other candidates. Cyan cross: the epipole that \ours predicts. Yellow ring: the true epipole. Inset: the ego image. Continued in \cref{fig:qual_right2,fig:qual_right3}.}
\label{fig:qual_right}
\end{figure*}

\begin{figure*}[p]
\centering
\qtile{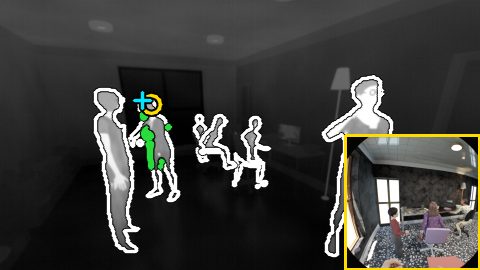}{Thermal}\hfill\qtile{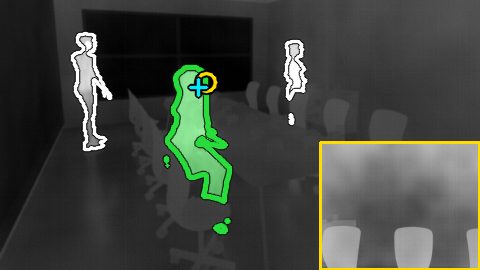}{}\hfill\qtile{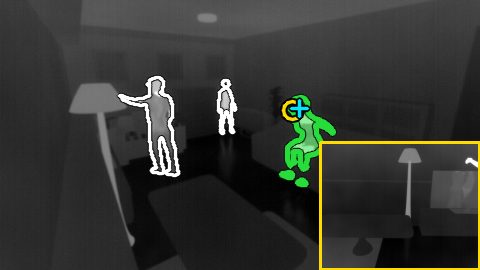}{}\\[2pt]
\qtile{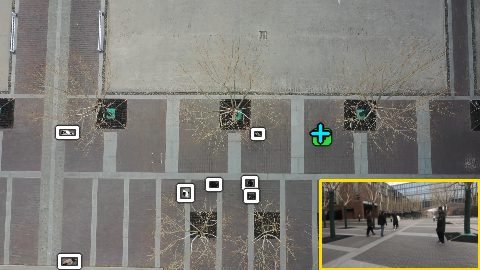}{Top view}\hfill\qtile{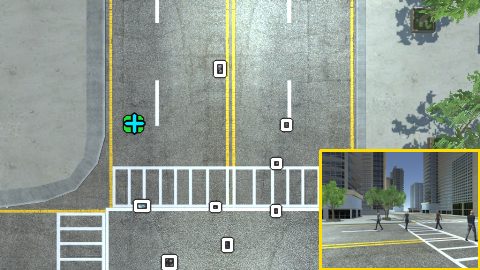}{}\hfill\qtile{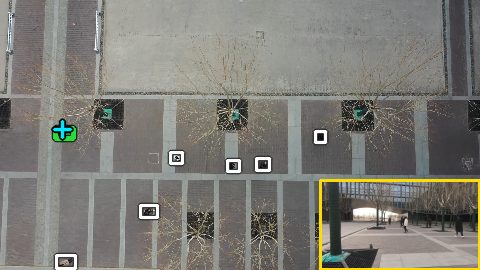}{}\\[2pt]
\qtile{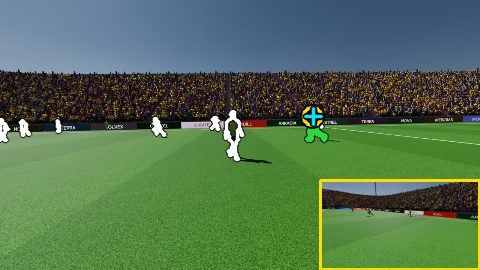}{Sports}\hfill\qtile{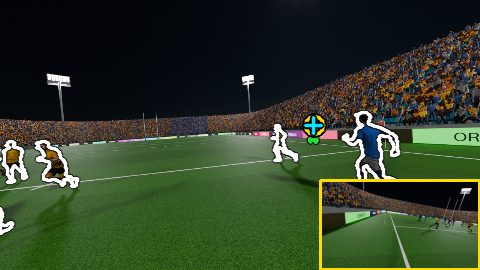}{}\hfill\qtile{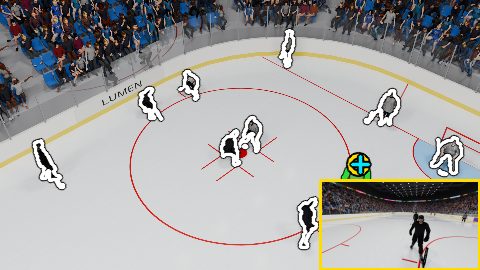}{}\\[2pt]
\qtile{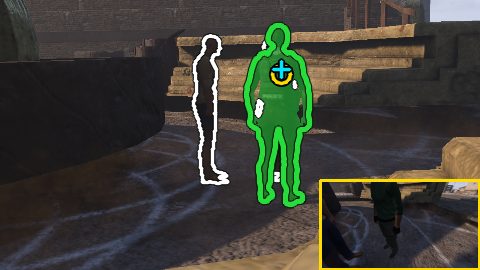}{Unreal}\hfill\qtile{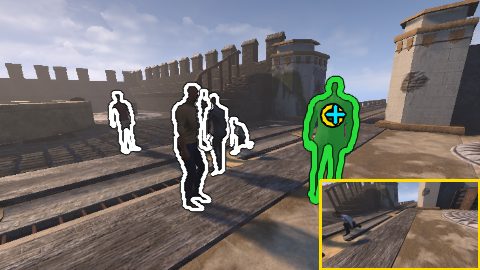}{}\hfill\qtile{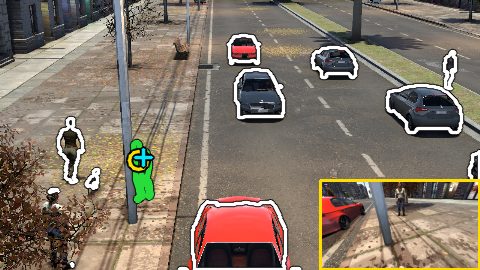}{}\\[2pt]
\qtile{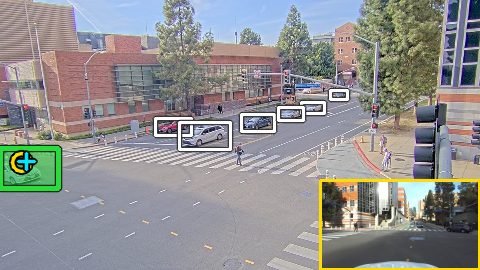}{Vehicles}\hfill\qtile{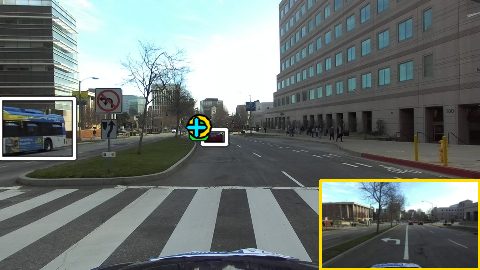}{}\hfill\qtile{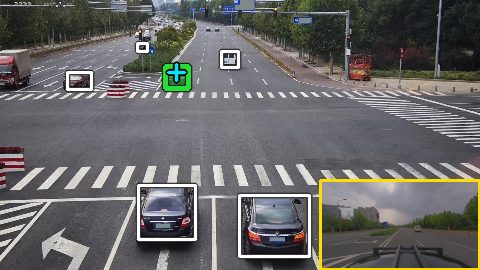}{}\\[2pt]
\qtile{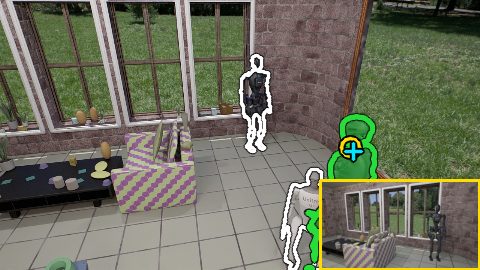}{Legged robots}\hfill\qtile{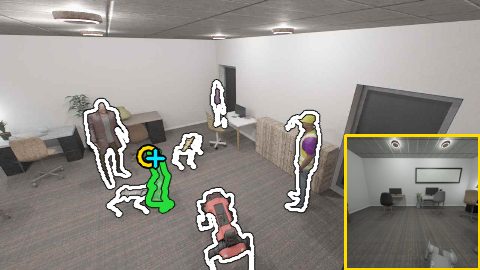}{}\hfill\qtile{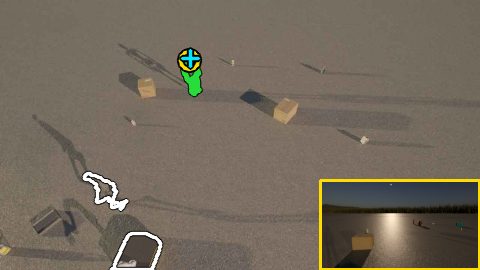}{}\\[2pt]
\caption{\textbf{Correct answers in every domain}, continued from \cref{fig:qual_right}, with its markers.}
\label{fig:qual_right2}
\end{figure*}

\begin{figure*}[p]
\centering
\qtile{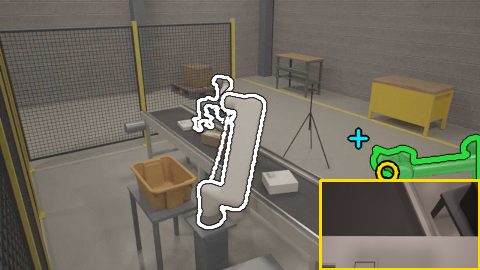}{Robot arms}\hfill\qtile{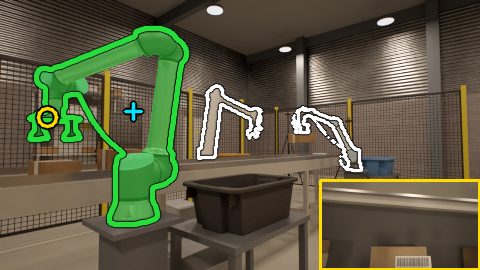}{}\hfill\qtile{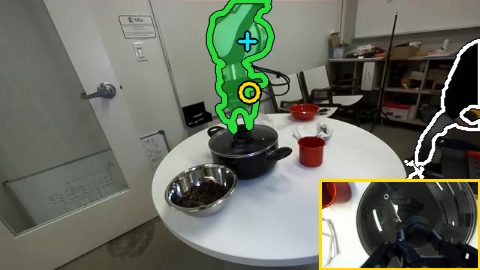}{}\\[2pt]
\qtile{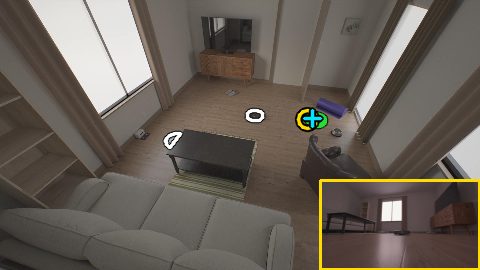}{Vacuums}\hfill\qtile{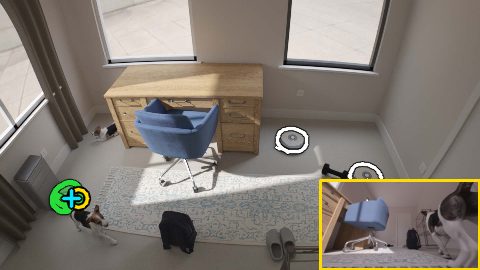}{}\hfill\qtile{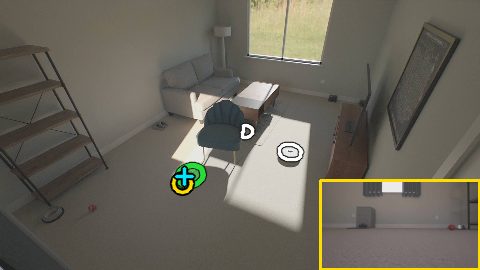}{}\\[2pt]
\qtile{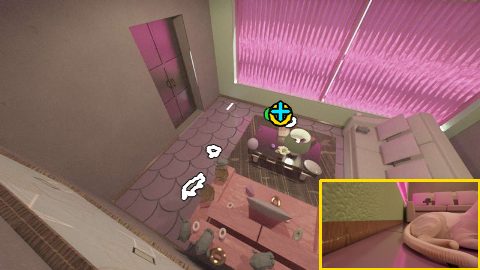}{Pets}\hfill\qtile{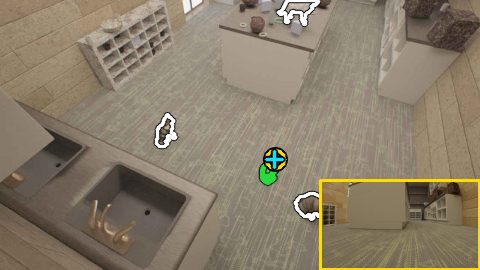}{}\hfill\qtile{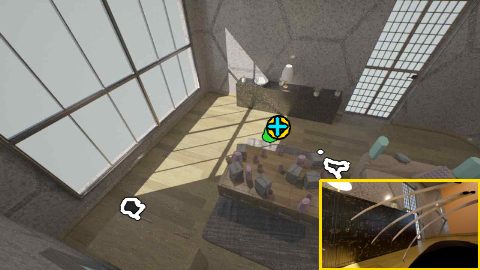}{}\\[2pt]
\qtile{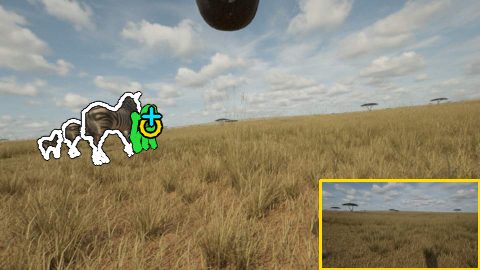}{Wild animals}\hfill\qtile{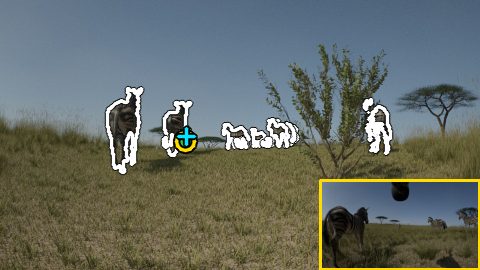}{}\hfill\qtile{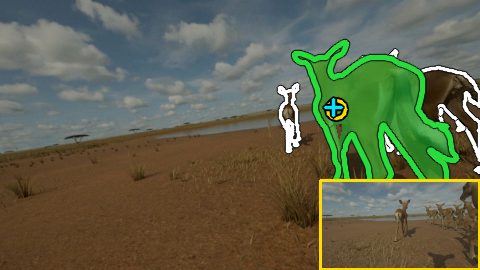}{}\\[2pt]
\qtile{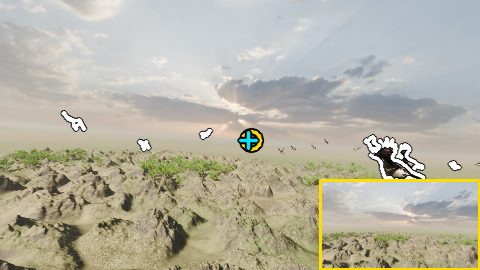}{Birds}\hfill\qtile{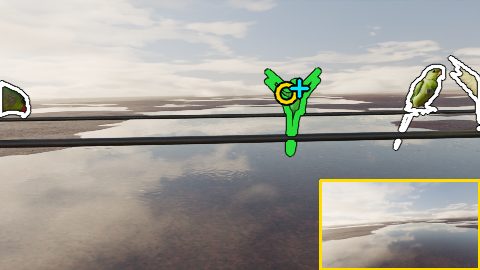}{}\hfill\qtile{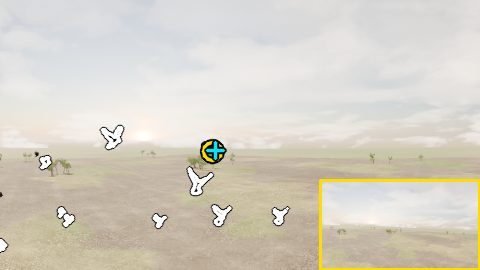}{}\\[2pt]
\qtile{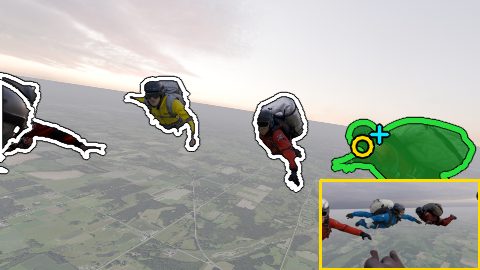}{Drones}\hfill\qtile{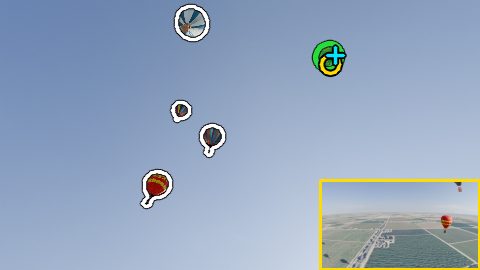}{}\hfill\qtile{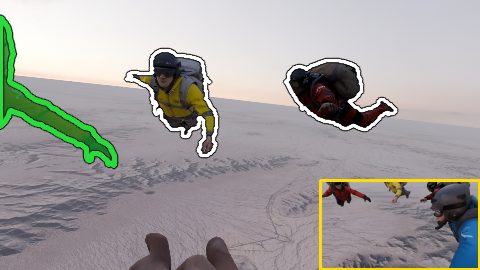}{}\\[2pt]
\caption{\textbf{Correct answers in every domain}, continued from \cref{fig:qual_right}, with its markers.}
\label{fig:qual_right3}
\end{figure*}

\begin{figure*}[p]
\centering
\qtile{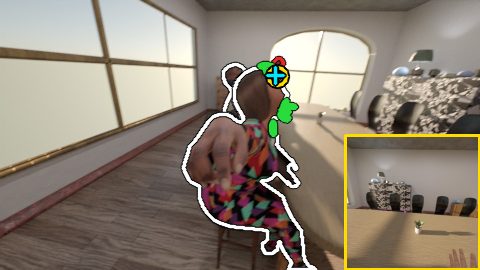}{Indoor}\hfill\qtile{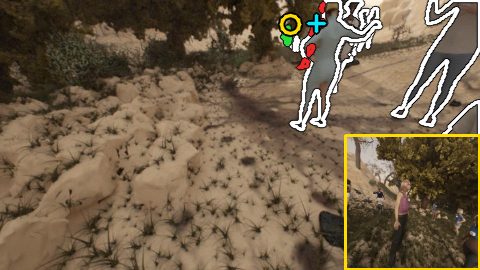}{Nature}\hfill\qtile{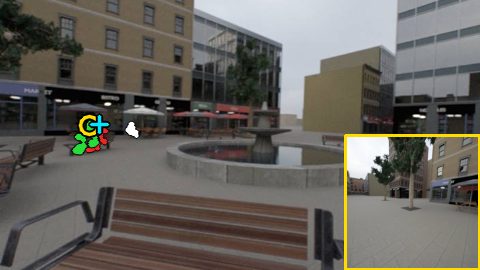}{City}\\[2pt]
\qtile{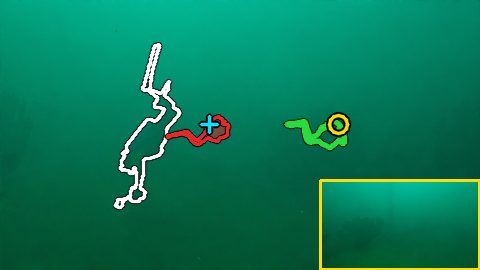}{Underwater}\hfill\qtile{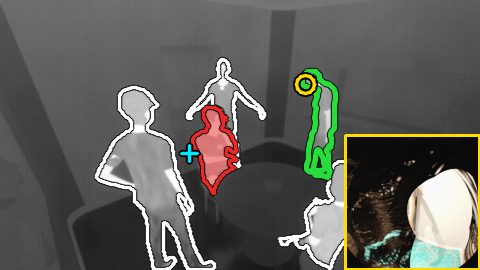}{Thermal}\hfill\qtile{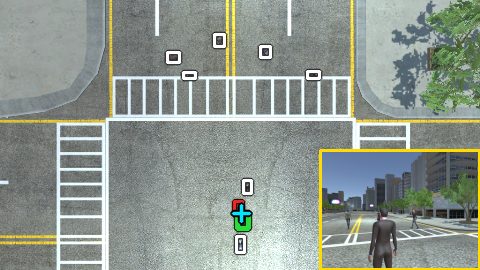}{Top view}\\[2pt]
\qtile{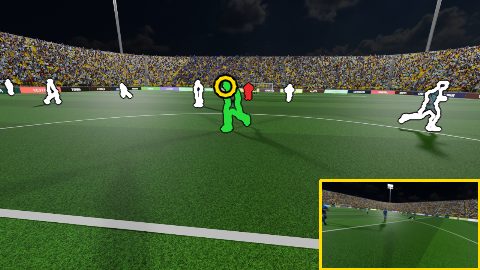}{Sports}\hfill\qtile{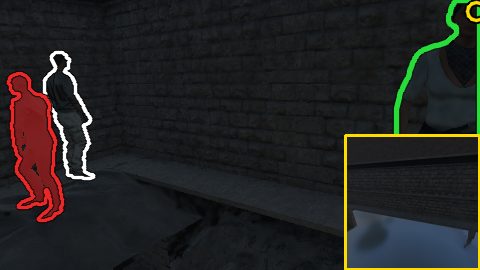}{Unreal}\hfill\qtile{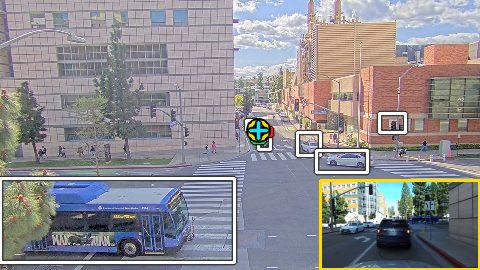}{Vehicles}\\[2pt]
\qtile{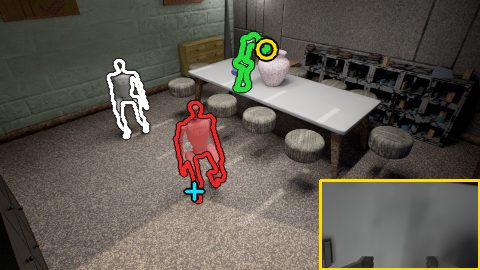}{Legged robots}\hfill\qtile{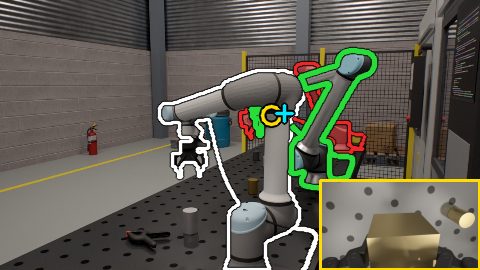}{Robot arms}\hfill\qtile{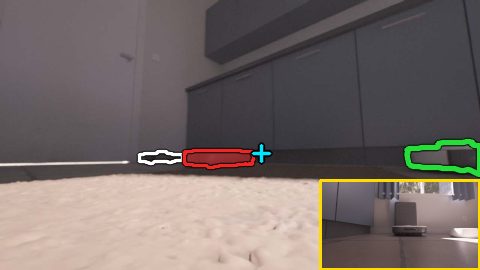}{Vacuums}\\[2pt]
\qtile{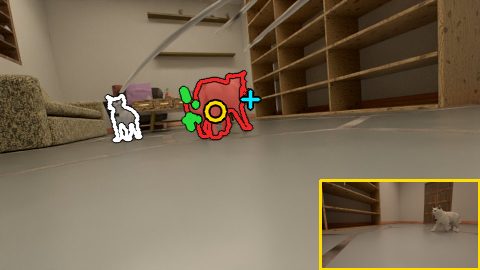}{Pets}\hfill\qtile{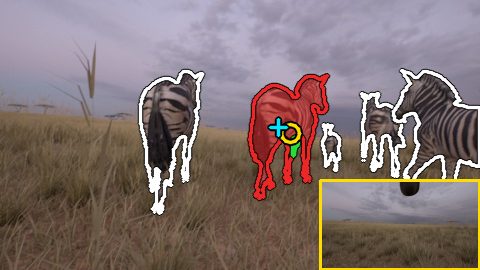}{Wild animals}\hfill\qtile{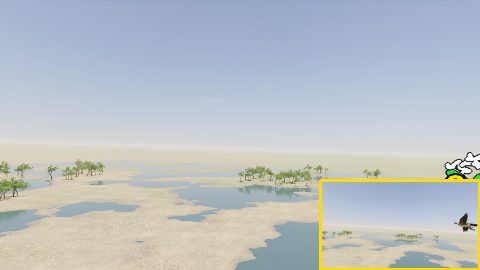}{Birds}\\[2pt]
\qtile{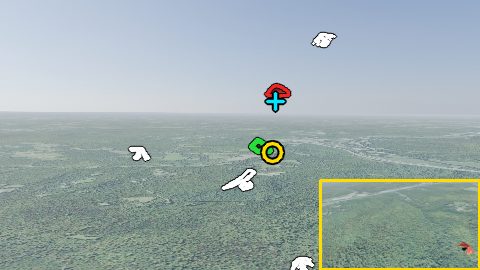}{Drones}\hfill\mbox{}\\[2pt]
\caption{\textbf{Failures} (\ours). All simulated except the vehicles~\cite{xiang2024v2xreal}. Red: the wrong pick; other markers as in \cref{fig:qual_right}.}
\label{fig:qual_wrong}
\end{figure*}

\Cref{fig:qual_right,fig:qual_right2,fig:qual_right3} show three correct answers per domain and \cref{fig:qual_wrong} one failure per domain, all taken as each domain's first qualifying Gold queries in hash order (with three to twelve candidates); among those drawn from real-human sets whose images may be shown, all from EgoHumans (\cref{fig:qual_right}), \ours has no failure, so \cref{fig:qual_wrong} covers sixteen domains. In \cref{fig:qual_right,fig:qual_right2,fig:qual_right3,fig:qual_wrong}, outlines mark the wearer (green), a wrong pick (red), and other candidates (white); the cyan cross is \ours's predicted epipole (averaged over both input orders), the yellow ring the true epipole, and the inset the ego image, with each tile cropped to the largest 16:9 window around the wearer. Among the sixteen failures, a neighbor at most $2^\circ$ farther from the true epipole than the wearer wins in seven domains, while pose error exceeds that gap in seven others.

\section{Limitations}
\label{sec:supp_limits}

We note three limitations. First, challenging scene configurations remain difficult. In dense herds, neighboring candidates can fall within the angular uncertainty of estimated camera poses; underwater, limited shared visual context makes cross-view correspondence unreliable. These challenges affect all evaluated methods: on wild animals, divers, and fish, the baselines in \cref{tab:who} achieve at most 23.0 \ka, compared with 46.4--55.9 for \ours.

Second, test-time adaptation requires unlabeled deployment queries. While queries from other scenes in the same deployment suffice (\cref{tab:carry}), and 25 queries recover two thirds of the adaptation gain (\cref{tab:stream}), this requirement may limit adaptation in data-scarce deployments. Without adaptation, \ours-ZS still leads all baselines on both macro averages in \cref{tab:who}, while pose-supervised training improves unseen \ka from 58.6 to 72.6 on the Gold tier (\cref{tab:components}).

Third, real-world coverage of non-human wearers remains limited. Apart from road vehicles, the robot arms in DROID and ALOHA, and indoor quadcopters, non-human wearers are represented by simulation. Simulated humans and vehicles preserve the method rankings observed on their real counterparts, and the appearance-level kernel inception distance is at most 1.15 times the real-to-real reference for the compared wearer types (\cref{sec:supp_realism}). Moreover, \ours achieves 81.1, 93.6, and 89.5 Acc on the three locked real-world sets (TUMTraf-V2X, ALOHA, and MILUV), versus 57.1, 88.5, and 81.2 for their strongest respective baselines (\cref{sec:supp_locked}). This reliance on simulation primarily reflects the scarcity of real-world paired ego–exo datasets for these wearer types, leaving broader real-world generalization to be validated in future work.

\section{Broader Impact}
\label{sec:supp_impact}
Identifying a camera wearer in a third-person view can serve safety, coordination, and accountability (for example, in robot fleets or body-camera review), and \exo also has a privacy use, telling those being filmed where the camera is. Conversely, this capability can single out individuals who film (such as journalists, protesters, or witnesses) in surveillance or broadcast footage and locate individuals wishing to stay unidentified. Our data contain no new recordings of humans, access to all data is gated, and any data or models we release will carry terms prohibiting their use to recognize or track specific individuals.

\makeatletter\setlength{\@dblfptop}{0pt}\setlength{\@dblfpbot}{0pt plus 1fil}\makeatother

\end{document}